\documentclass[runningheads]{llncs}

\usepackage{tikz}
\usepackage{comment}
\usepackage{amsmath,amssymb} 
\usepackage{color}

\usepackage{graphicx}
\usepackage{booktabs}
\usepackage{enumitem}
\usepackage{makecell} 
\usepackage[caption=false]{subfig}
\usepackage{multirow} 
\usepackage{adjustbox} 
\usepackage{colortbl} 
\usepackage{url} 
\usepackage{hyperref}
\usepackage{footnote}

\usepackage[capitalize]{cleveref}
\crefname{section}{Sec.}{Secs.}
\Crefname{section}{Section}{Sections}
\Crefname{table}{Table}{Tables}
\crefname{table}{Table}{Tables}

\newcommand{\norm}[1]{\left\lVert#1\right\rVert}

\newcommand{\dname}{\textsc{GT-RAIN}}

\usepackage[accsupp]{axessibility}  

\begin{document}

\pagestyle{headings}
\mainmatter
\def\ECCVSubNumber{XXXX}  

\title{Not Just Streaks: Towards Ground Truth for Single Image Deraining} 

\titlerunning{Not Just Streaks: Towards Ground Truth for Single Image Deraining}

\author{Yunhao Ba\inst{1}$^{\star}$ \and
Howard Zhang\inst{1}\thanks{Equal contribution.} \and
Ethan Yang\inst{1} \and
Akira Suzuki\inst{1} \and
Arnold Pfahnl\inst{1} \and
Chethan Chinder Chandrappa\inst{1}\index{Chandrappa, Chethan Chinder} \and
Celso M. de Melo\inst{2}\index{de Melo, Celso M.} \and
Suya You\inst{2} \and
Stefano Soatto\inst{1} \and
Alex Wong\inst{3} \and
Achuta Kadambi\inst{1}}

\authorrunning{Y. Ba et al.}
%
\institute{University of California, Los Angeles
\email{\{yhba,hwdz15508,eyang657,asuzuki100,ajpfahnl,chinderc\}@ucla.edu}\\
\email{soatto@cs.ucla.edu}, \email{achuta@ee.ucla.edu}\\ \and 
DEVCOM Army Research Laboratory
\email{\{celso.m.demelo.civ,suya.you.civ\}@army.mil} \\ \and
Yale University \\
\email{alex.wong@yale.edu}}


\maketitle
\vspace{-20pt}
\begin{abstract}
We propose a large-scale dataset of real-world rainy and clean image pairs and a method to remove degradations, induced by rain streaks and rain accumulation, from the image. As there exists no real-world dataset for deraining, current state-of-the-art methods rely on synthetic data and thus are limited by the sim2real domain gap; moreover, rigorous evaluation remains a challenge due to the absence of a real paired dataset. We fill this gap by collecting a real paired deraining dataset through meticulous control of non-rain variations. Our dataset enables paired training and quantitative evaluation for diverse real-world rain phenomena (e.g. rain streaks and rain accumulation). To learn a representation robust to rain phenomena, we propose a deep neural network that reconstructs the underlying scene by minimizing a rain-robust loss between rainy and clean images. Extensive experiments demonstrate that our model outperforms the state-of-the-art deraining methods on real rainy images under various conditions. Project website: \url{https://visual.ee.ucla.edu/gt_rain.htm/}.
\renewcommand{\thefootnote}{\fnsymbol{footnote}}%
\footnotetext[0]{\emph{Approved for public release: distribution is unlimited.}}

\end{abstract}
\vspace{-10pt}
\section{Introduction}

Single-image deraining aims to remove degradations induced by rain from images. Restoring rainy images not only improves their aesthetic properties, but also supports reuse of abundant publicly available pretrained models across computer vision tasks. Top performing methods use deep networks, but suffer from a common issue: it is not possible to obtain ideal real ground-truth pairs of rain and clean images. The same scene, in the same space and time, cannot be observed both with and without rain. To overcome this, deep learning based rain removal relies on synthetic data. 

The use of synthetic data in deraining is prevalent~\cite{fu2017removing,hu2019depth,li2019heavy,li2016rain,yang2017deep,zhang2018density,zhang2019image}. However, current rain simulators cannot model all the complex effects of rain, which leads to unwanted artifacts when applying models trained on them to real-world rainy scenes. For instance, a number of synthetic methods add \emph{rain streaks} to clean images to generate the pair~\cite{fu2017removing,li2016rain,yang2017deep,zhang2018density,zhang2019image}, but rain does not only manifest as streaks: If raindrops are further away, the streaks meld together, creating \emph{rain accumulation}, or \emph{veiling} effects, which are exceedingly difficult to simulate. A further challenge with synthetic data is that results on real test data can only be evaluated qualitatively, for no real paired ground truth exists.

\begin{figure}[t]
    \centering
    \includegraphics[width=\textwidth]{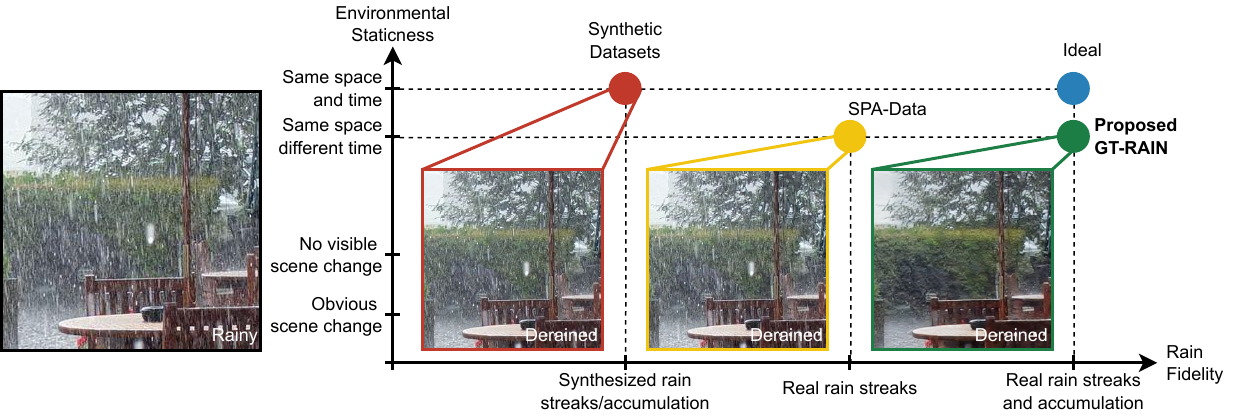}
    \caption{\textbf{The points above depict datasets and their corresponding  outputs from models trained on them.} These outputs come from a real rain image from the Internet. Our opinion* is that \dname\ can be the right dataset for the deraining community to use because it has a smaller domain gap to the ideal ground truth.\, * Why an asterisk? The asterisk emphasizes that this is an ``opinion". It is impossible to quantify the domain gap because collecting true real data is infeasible. To date, deraining is largely a viewer's imagination of what the derained scene should look like. Therefore, we present the derained images above and leave it to the viewer to judge the gap. Additionally, \dname\ can be used in complement with the litany of synthetic datasets~\cite{fu2017removing,hu2019depth,li2019heavy,li2016rain,yang2017deep,zhang2018density,zhang2019image}, as illustrated in~\cref{tab:finetune_results}.}
    \label{fig:teaser}
\end{figure}

Realizing these limitations of synthetic data, we tackle the problem from another angle by relaxing the concept of ideal ground truth to a sufficiently short time window (see~\cref{fig:teaser}). We decide to conduct the experiment of obtaining short time interval paired data, particularly in light of the timely growth and diversity of landscape YouTube live streams. We strictly filter such videos with objective criteria on illumination shifts, camera motions, and motion artifacts. Further correction algorithms are applied for subtle variations, such as slight movements of foliage. We call this dataset \dname, as it is a first attempt to provide real paired data for deraining. Although our dataset relies on streamers, YouTube's fair use policy allows its release to the academic community. \\ 

\noindent\textbf{Defining ``real, paired ground truth'':} Clearly, obtaining real, paired ground truth data by capturing a rain and rain-free image pair at the exact same space and time is not feasible. However, the dehazing community has accepted several test sets~\cite{ancuti2019dense,ancuti2020nh,ancuti2018Ohaze,ancuti2018Ihaze} following these guidelines as a satisfactory replacement for evaluation purposes: 
\begin{itemize}[itemsep=0.5pt] 
    \item A pair of degraded and clean images is captured as real photos at two different timestamps;
    \item Illumination shifts are limited by capturing data on cloudy days;
    \item The camera configuration remains identical while capturing the degraded and clean images.
\end{itemize}
We produce the static pairs in \dname\ by following the above criterion set forth by the dehazing community while enforcing a stricter set of rules on sky and local motion. More importantly, as a step closer towards obtaining real ground truth pairs, we capture natural weather effects instead, which address problems of scale and variability that inherently come with simulating weather through man-made methods. In the results of the proposed method, we not only see quantitative and qualitative improvements, but also showcase a unique ability to handle diverse rain physics that was not previously handled by synthetic data. \\

\noindent\textbf{Contributions:} In summary, we make the following contributions:
\begin{itemize}[itemsep=1pt]
    \item We propose a real-world paired dataset: \dname. The dataset captures \textit{real} rain phenomena, from rain streaks to accumulation under various rain fall conditions, to bridge the domain gap that is too complex to be modeled by synthetic~\cite{fu2017removing,hu2019depth,li2019heavy,li2016rain,yang2017deep,zhang2018density,zhang2019image} and semi-real~\cite{wang2019spatial} datasets. 
    
    \item We introduce an avenue for the deraining community to now have standardized quantitative and qualitative evaluations. Previous evaluations were quantifiable only wrt. simulations.
    
    \item We propose a framework to reconstruct the underlying scene by learning representations robust to the rain phenomena via a rain-robust loss function. Our approach outperforms the state of the art~\cite{zamir2021multi} by 12.1\% PSNR on average for deraining real images.
    
\end{itemize}

\section{Related Work}

\noindent\textbf{Rain physics:} Raindrops exhibit diverse physical properties while falling, and 
many experimental studies have been conducted to investigate them, i.e. equilibrium shape~\cite{beard1987new}, size~\cite{marshall1948distribution}, terminal velocity~\cite{foote1969terminal,gunn1949terminal}, spatial distribution~\cite{manning1993stochastic}, and temporal distribution~\cite{zhang2006rain}. A mixture of these distinct properties transforms the photometry of a raindrop into a complex mapping of the environmental radiance which considers refraction, specular reflection, and internal reflection~\cite{garg2007vision}:
\begin{equation}
    L(\hat{n}) = L_r(\hat{n}) + L_s(\hat{n}) + L_p(\hat{n}),
\end{equation}
where $L(\hat{n})$ is the radiance at a point on the raindrop surface with normal $\hat{n}$, $L_r(\cdot)$ is the radiance of the refracted ray, $L_s(\cdot)$ is the radiance of the specularly reflected ray, and $L_p(\cdot)$ is the radiance of the internally reflected ray. In real images, the appearance of rain streaks is also affected by motion blur and background intensities. Moreover, the dense rain accumulation results in sophisticated veiling effects. Interactions of these complex phenomena make it challenging to simulate realistic rain effects. Until \dname, previous works~\cite{guo2021efficientderain,hu2021single,jiang2020multi,li2019heavy,wang2020a,wang2019spatial,zamir2021multi} have relied heavily on simulated rain and are limited by the sim2real gap. \\

\begin{table}[t]
\caption{\textbf{Our proposed large-scale dataset enables paired training and quantitative evaluation for real-world deraining.} We consider SPA-Data~\cite{wang2019spatial} as a semi-real dataset since it only contains real rainy images, where the pseudo ground-truth images are synthesized from a rain streak removal algorithm.} 

\scriptsize
\centering
\begin{tabular}{cccc}

\cellcolor[HTML]{656565}\rule{0pt}{3ex}\textcolor{white}{\textbf{Dataset}} & 
\cellcolor[HTML]{656565}\rule{0pt}{3ex}\textcolor{white}{\textbf{Type}} & 
\cellcolor[HTML]{656565}\rule{0pt}{3ex}\textcolor{white}{\textbf{Rain Effects}} &
\cellcolor[HTML]{656565}\rule{0pt}{3ex}\textcolor{white}{\textbf{Size}}
\\[1ex]

\cellcolor[HTML]{EFEFEF}\begin{tabular}[c]{@{}c@{}}\rule{0pt}{3ex} Rain12~\cite{li2016rain}\end{tabular} & 
\cellcolor[HTML]{FFCCC9}\begin{tabular}[c]{@{}c@{}}\rule{0pt}{3ex} Simulated\end{tabular} & 
\cellcolor[HTML]{FFCCC9}\begin{tabular}[c]{@{}c@{}}\rule{0pt}{3ex} Synth. streaks only\end{tabular} &
\cellcolor[HTML]{FFCCC9}\begin{tabular}[c]{@{}c@{}}\rule{0pt}{3ex} 12\end{tabular} 
\\[1ex]

\cellcolor[HTML]{EFEFEF}\begin{tabular}[c]{@{}c@{}}\rule{0pt}{3ex} Rain100L~\cite{yang2017deep}\end{tabular} & 
\cellcolor[HTML]{FFCCC9}\begin{tabular}[c]{@{}c@{}}\rule{0pt}{3ex} Simulated\end{tabular} & 
\cellcolor[HTML]{FFCCC9}\begin{tabular}[c]{@{}c@{}}\rule{0pt}{3ex} Synth. streaks only\end{tabular} &
\cellcolor[HTML]{FFCCC9}\begin{tabular}[c]{@{}c@{}}\rule{0pt}{3ex} 300\end{tabular} 
\\[1ex]

\cellcolor[HTML]{EFEFEF}\begin{tabular}[c]{@{}c@{}}\rule{0pt}{3ex} Rain800~\cite{zhang2019image}\end{tabular} & 
\cellcolor[HTML]{FFCCC9}\begin{tabular}[c]{@{}c@{}}\rule{0pt}{3ex} Simulated\end{tabular} & 
\cellcolor[HTML]{FFCCC9}\begin{tabular}[c]{@{}c@{}}\rule{0pt}{3ex} Synth. streaks only\end{tabular} &
\cellcolor[HTML]{FFCCC9}\begin{tabular}[c]{@{}c@{}}\rule{0pt}{3ex} 800\end{tabular} 
\\[1ex]

\cellcolor[HTML]{EFEFEF}\begin{tabular}[c]{@{}c@{}}\rule{0pt}{3ex} Rain100H~\cite{yang2017deep}\end{tabular} & 
\cellcolor[HTML]{FFCCC9}\begin{tabular}[c]{@{}c@{}}\rule{0pt}{3ex} Simulated\end{tabular} & 
\cellcolor[HTML]{FFCCC9}\begin{tabular}[c]{@{}c@{}}\rule{0pt}{3ex} Synth. streaks only\end{tabular} &
\cellcolor[HTML]{FFCCC9}\begin{tabular}[c]{@{}c@{}}\rule{0pt}{3ex} 1.9K\end{tabular} 
\\[1ex]

\cellcolor[HTML]{EFEFEF}\begin{tabular}[c]{@{}c@{}}\rule{0pt}{3ex} Outdoor-Rain~\cite{li2019heavy}\end{tabular} & 
\cellcolor[HTML]{FFCCC9}\begin{tabular}[c]{@{}c@{}}\rule{0pt}{3ex} Simulated\end{tabular} & 
\cellcolor[HTML]{FFFFC7}\begin{tabular}[c]{@{}c@{}}\rule{0pt}{3ex} Synth. streaks \& Synth. accumulation\end{tabular} &
\cellcolor[HTML]{FFFFC7}\begin{tabular}[c]{@{}c@{}}\rule{0pt}{3ex} 10.5K\end{tabular} 
\\[1ex]

\cellcolor[HTML]{EFEFEF}\begin{tabular}[c]{@{}c@{}}\rule{0pt}{3ex} RainCityscapes~\cite{hu2019depth}\end{tabular} & 
\cellcolor[HTML]{FFCCC9}\begin{tabular}[c]{@{}c@{}}\rule{0pt}{3ex} Simulated\end{tabular} & 
\cellcolor[HTML]{FFFFC7}\begin{tabular}[c]{@{}c@{}}\rule{0pt}{3ex} Synth. streaks \& Synth. accumulation\end{tabular} &
\cellcolor[HTML]{FFFFC7}\begin{tabular}[c]{@{}c@{}}\rule{0pt}{3ex} 10.62K\end{tabular} 
\\[1ex]

\cellcolor[HTML]{EFEFEF}\begin{tabular}[c]{@{}c@{}}\rule{0pt}{3ex} Rain12000~\cite{zhang2018density}\end{tabular} & 
\cellcolor[HTML]{FFCCC9}\begin{tabular}[c]{@{}c@{}}\rule{0pt}{3ex} Simulated\end{tabular} & 
\cellcolor[HTML]{FFCCC9}\begin{tabular}[c]{@{}c@{}}\rule{0pt}{3ex} Synth. streaks only\end{tabular} &
\cellcolor[HTML]{FFFFC7}\begin{tabular}[c]{@{}c@{}}\rule{0pt}{3ex} 13.2K\end{tabular} 
\\[1ex]

\cellcolor[HTML]{EFEFEF}\begin{tabular}[c]{@{}c@{}}\rule{0pt}{3ex} Rain14000~\cite{fu2017removing}\end{tabular} & 
\cellcolor[HTML]{FFCCC9}\begin{tabular}[c]{@{}c@{}}\rule{0pt}{3ex} Simulated\end{tabular} & 
\cellcolor[HTML]{FFCCC9}\begin{tabular}[c]{@{}c@{}}\rule{0pt}{3ex} Synth. streaks only\end{tabular} &
\cellcolor[HTML]{FFFFC7}\begin{tabular}[c]{@{}c@{}}\rule{0pt}{3ex} 14K\end{tabular} 
\\[1ex]

\cellcolor[HTML]{EFEFEF}\begin{tabular}[c]{@{}c@{}}\rule{0pt}{3ex} NYU-Rain~\cite{li2019heavy}\end{tabular} & 
\cellcolor[HTML]{FFCCC9}\begin{tabular}[c]{@{}c@{}}\rule{0pt}{3ex} Simulated\end{tabular} & 
\cellcolor[HTML]{FFFFC7}\begin{tabular}[c]{@{}c@{}}\rule{0pt}{3ex} Synth. streaks \& Synth. accumulation\end{tabular} &
\cellcolor[HTML]{FFFFC7}\begin{tabular}[c]{@{}c@{}}\rule{0pt}{3ex} 16.2K\end{tabular} 
\\[1ex]

\cellcolor[HTML]{EFEFEF}\begin{tabular}[c]{@{}c@{}}\rule{0pt}{3ex} SPA-Data~\cite{wang2019spatial}\end{tabular} & 
\cellcolor[HTML]{FFFFC7}\begin{tabular}[c]{@{}c@{}}\rule{0pt}{3ex} Semi-real\end{tabular}& 
\cellcolor[HTML]{FFCCC9}\begin{tabular}[c]{@{}c@{}}\rule{0pt}{3ex} Real streaks only\end{tabular} &
\cellcolor[HTML]{9AFF99}\begin{tabular}[c]{@{}c@{}}\rule{0pt}{3ex} 29.5K\end{tabular} 
\\[1ex]

\cellcolor[HTML]{EFEFEF}\rule{0pt}{3ex} \textbf{Proposed} & 
\cellcolor[HTML]{9AFF99}\rule{0pt}{3ex} Real & 
\cellcolor[HTML]{9AFF99}\rule{0pt}{3ex} Real streaks \& Real accumulation &
\cellcolor[HTML]{9AFF99}\rule{0pt}{3ex} 31.5K
\\[1ex]
\end{tabular}
\label{tab:related_work}
\end{table}

\noindent\textbf{Deraining datasets:} Most data-driven deraining models require paired rainy and clean, rain-free ground-truth images for training. Due to the difficulty of collecting real paired samples, previous works focus on synthetic datasets, such as Rain12 \cite{li2016rain}, Rain100L \cite{yang2017deep}, Rain100H \cite{yang2017deep}, Rain800 \cite{zhang2019image}, Rain12000 \cite{zhang2018density}, Rain14000 \cite{fu2017removing}, NYU-Rain \cite{li2019heavy}, Outdoor-Rain \cite{li2019heavy}, and RainCityscapes \cite{hu2019depth}. Even though synthetic images from these datasets incorporate some physical characteristics of real rain, significant gaps still exist between synthetic and real data~\cite{yang2020single}. More recently, a ``paired" dataset with real rainy images (SPA-Data) was proposed in~\cite{wang2019spatial}. However, their ``ground-truth'' images are in fact a product of a video-based deraining method -- synthesized based on the temporal motions of raindrops which may introduce artifacts and blurriness; moreover, the associated rain accumulation and veiling effects are not considered. In contrast, we collect pairs of real-world rainy and clean ground-truth images by enforcing rigorous selection criteria to minimize the environmental variations. To the best of our knowledge, our dataset is the first large-scale dataset with real paired data. Please refer to~\cref{tab:related_work} for a detailed comparison of the deraining datasets. \linebreak

\noindent\textbf{Single-image deraining:} Previous methods used model-based solutions to derain~\cite{chen2013generalized,jiang2018fastderain,li2016rain,luo2015removing}. More recently, deep-learning based methods have seen increasing popularity and progress~\cite{fu2017clearing,guo2021efficientderain,hu2021single,jiang2020multi,li2019heavy,pan2018learning,ren2020single,wang2020a,wang2019spatial,yang2017deep,zamir2021multi,zhang2018density}. The multi-scale progressive fusion network (MSPFN)~\cite{jiang2020multi} characterizes and reconstructs rain streaks at multiple scales. The rain convolutional dictionary network (RCDNet)~\cite{wang2020a} encodes the rain shape using the intrinsic convolutional dictionary learning mechanism. The multi-stage progressive image restoration network (MPRNet)~\cite{zamir2021multi} splits the image into different sections in various stages to learn contextualized features at different scales. The spatial attentive network (SPANet)~\cite{wang2019spatial} learns physical properties of rain streaks in a local neighborhood and reconstructs the clean background using non-local information. EfficientDeRain (EDR)~\cite{guo2021efficientderain} aims to derain efficiently in real time by using pixel-wise dilation filtering. Other than rain streak removal, the heavy rain restorer (HRR)~\cite{li2019heavy} and the depth-guided non-local network (DGNL-Net)~\cite{hu2021single} have also attempted to address rain accumulation effects. All of these prior methods use synthetic or semi-real datasets, and show limited generalizability to real images. In contrast, we propose a derainer that learns a rain-robust representation directly.

\begin{figure*}[t]
  \centering
  \includegraphics[width=\textwidth]{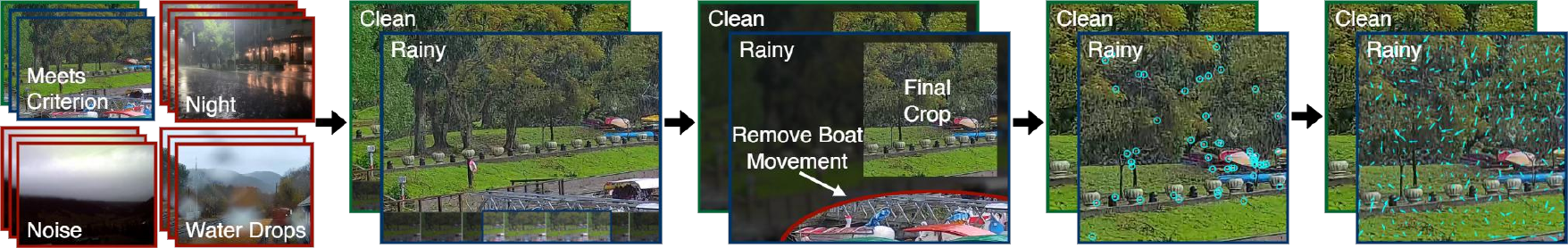}
  \includegraphics[width=\textwidth]{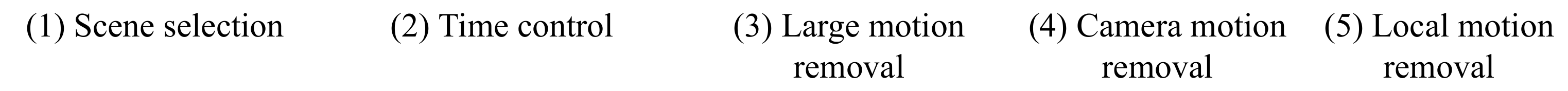}
  \caption{\textbf{We collect the a real paired deraining dataset by rigorously controlling the environmental variations.} First, we remove heavily degraded videos such as scenes without proper exposure, noise, or water droplets on the lens. Next, we carefully choose the rainy and clean frames as close as possible in time to mitigate illumination shifts before cropping to remove large movement. Lastly, we correct for small camera motion (due to strong wind) using SIFT~\cite{lowe2004sift} and RANSAC~\cite{fischler1981random} and perform elastic image registration~\cite{thirion1998image,vercauteren2009diffeomorphic} by estimating the displacement field when necessary.}
  \label{fig:collection_pipeline}
\end{figure*}

\section{Dataset} \label{sec:dataset}

We now describe our method to control variations in a real dataset of paired images taken at two different timestamps, as illustrated in~\cref{fig:collection_pipeline}. \\

\noindent\textbf{Data collection:} 
We collect rain and clean ground-truth videos using a Python program based on FFmpeg to download videos from YouTube live streams across the world. For each live stream, we record the location in order to determine whether there is rain according to the OpenWeatherMap API~\cite{openWeather}. We also determine the time of day to filter out nighttime videos. After the rain stops, we continue downloading in order to collect clean ground-truth frames. Note: while our dataset is formatted for single-image deraining, it can be re-purposed for video deraining as well by considering the timestamps of the frames collected. \\

\begin{figure*}[t]
  \centering
  \includegraphics[width=\linewidth]{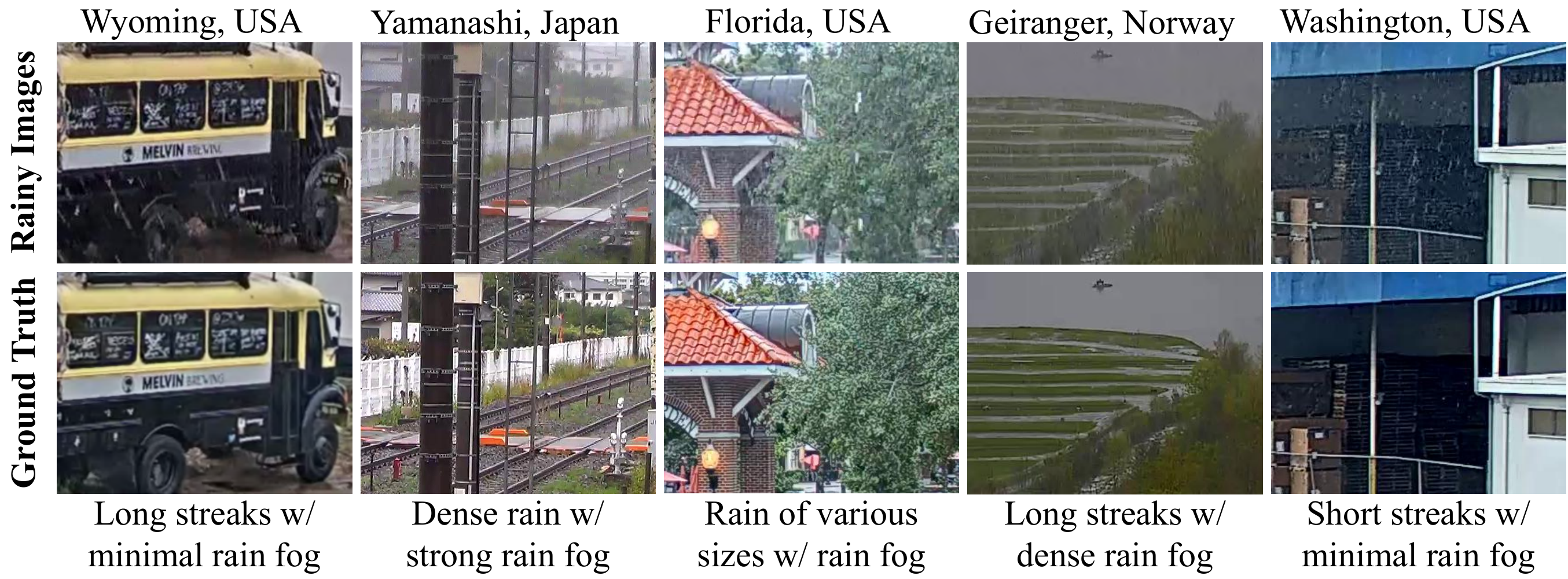}
  \caption{\textbf{Our proposed dataset contains diverse rainy images collected across the world.} We illustrate several representative image pairs with various rain streak appearances and rain accumulation strengths at different geographic locations.} 
  \label{fig:dataset_samples}
\end{figure*}

\noindent\textbf{Collection criteria:} To minimize variations between rainy and clean frames, videos are filtered based on a strict set of collection criteria. Note that we perform realignment for camera and local motion only when necessary -- with manual oversight to filter out cases where motion still exists after realignment. Please see examples of motion correction and alignment in the supplement.
\begin{itemize}[itemsep=1pt]
    \item \textbf{Heavily degraded scenes} that contain excessive noise, webcam artifacts, poor resolution, or poor camera exposure are filtered out as the underlying scene cannot be inferred from the images.

    \item \textbf{Water droplets} on the surface of the lens occlude large portions of the scene and also distort the image. Images containing this type of degradation are filtered out as it is out of the scope of this work -- we focus on rain streak and rain accumulation phenomena.
    
    \item \textbf{Illumination shifts} are mitigated by minimizing the time difference between rainy and clean frames. Our dataset has an average time difference of 25 minutes, which drastically limits large changes in global illumination due to sun position, clouds, etc. 

    \item \textbf{Background changes} containing large discrepancies (e.g cars, people, swaying foliage, water surfaces) are cropped from the frame to ensure that clean and rainy images are aligned. By limiting the average time difference between scenes, we also minimize these discrepancies before filtering. All sky regions are cropped out as well to ensure proper background texture.
    
    \item \textbf{Camera motion.} Adverse weather conditions, i.e. heavy wind, can cause camera movements between the rainy and clean frames. To address this, we use the Scale Invariant Feature Transform (SIFT)~\cite{lowe2004sift} and Random Sample Consensus (RANSAC)~\cite{fischler1981random} to compute the homography to realign the frames.
    
    \item \textbf{Local motion.} Despite controlling for motion whenever possible, certain scenes still contain small local movements that are unavoidable, especially in areas of foliage. To correct for this, we perform elastic image registration when necessary by estimating the displacement field~\cite{thirion1998image,vercauteren2009diffeomorphic}. 
\end{itemize}

\noindent\textbf{Dataset statistics:} Our large-scale dataset includes a total of 31,524 rainy and clean frame pairs, which is split into 26,124 training frames, 3,300 validation frames, and 2,100 testing frames. These frames are taken from 101 videos, covering a large variety of background scenes from urban locations (e.g. buildings, streets, cityscapes) to natural scenery (e.g. forests, plains, hills). We span a wide range of geographic locations (e.g. North America, Europe, Oceania, Asia) to ensure that we capture diverse scenes and rain fall conditions. The scenes also include varying degrees of illumination from different times of day and rain of varying densities, streak lengths, shapes, and sizes. The webcams cover a wide array of resolutions, noise levels, intrinsic parameters (focal length, distortion), etc. As a result, our dataset captures diverse rain effects that cannot be accurately reproduced by SPA-Data~\cite{wang2019spatial} or synthetic datasets~\cite{fu2017removing,hu2019depth,li2019heavy,li2016rain,yang2017deep,zhang2018density,zhang2019image}. See~\cref{fig:dataset_samples} for representative image pairs in \dname. 

\section{Learning to Derain Real Images} \label{sec:proposed_method}
\begin{figure*}[t]
    \centering
    \includegraphics[width=\textwidth]{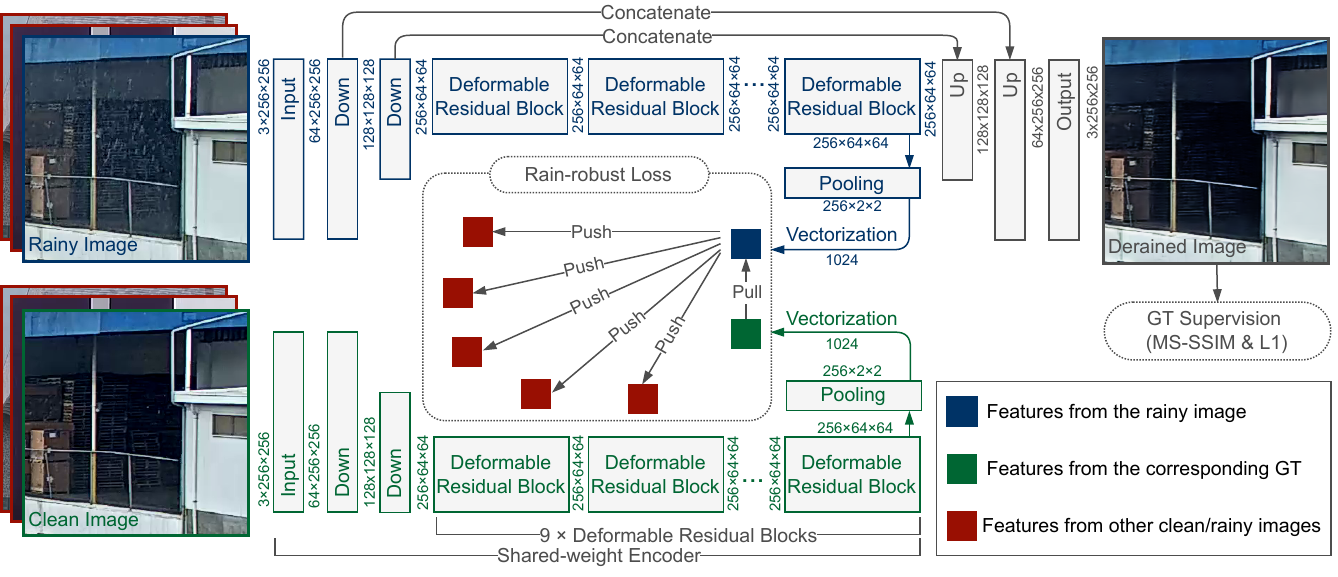}
    \caption{\textbf{By minimizing a rain-robust objective, our model learns robust features for reconstruction.} When training, a shared-weight encoder is used to extract features from rainy and ground-truth images. These features are then evaluated with the rain-robust loss, where features from a rainy image and its ground-truth are encouraged to be similar. Learned features from the rainy images are also fed into a decoder to reconstruct the ground-truth images with MS-SSIM and $\ell1$ loss functions.} 
    \label{fig:pipeline}
\end{figure*}

To handle greater diversity of rain streak appearance, we propose to learn a representation (illustrated in~\cref{fig:pipeline}) that is robust to rain for real image deraining. \\

\noindent\textbf{Problem formulation:} 
Most prior works emphasize on the rain streak removal and rely on the following equation to model rain~\cite{deng2018directional,fu2017removing,li2018non,li2016rain,wang2020a,wang2019spatial,yasarla2019uncertainty,zhang2018density,zhu2017joint}:
\begin{equation} \label{eq:rain_model}
    \mathbf{I} = \mathbf{J} + \sum_i^n \mathbf{S}_i,
\end{equation}
where $\mathbf{I} \in \mathbb{R}^{3 \times H \times W}$ is the observed rainy image, $\mathbf{J} \in \mathbb{R}^{3 \times H \times W}$ is the rain-free or ``clean'' image, and $\mathbf{S}_i$ is the $i$-th rain layer. However, real-world rain can be more complicated due to the dense rain accumulation and the rain veiling effect~\cite{li2019heavy,li2020all,yang2019joint}. These additional effects, which are visually similar to fog and mist, may cause severe degradation, and thus their removal should also be considered for single-image deraining. With \dname, it now becomes possible to study and conduct optically challenging, real-world rainy image restoration. 

Given an image $\mathbf{I}$ of a scene captured during rain, we propose to learn a function $\mathcal{F}(\cdot, \theta)$ parameterized by $\theta$ to remove degradation induced by the rain phenonmena. This function is realized as a neural network (see~\cref{fig:pipeline}) that takes as input a rainy image $\mathbf{I}$ and outputs a ``clean'' image $\hat{\mathbf{J}} = \mathcal{F}(\mathbf{I}, \theta) \in \mathbb{R}^{3 \times H \times W}$, where undesirable characteristics, i.e. rain streaks and rain accumulation, are removed from the image to reconstruct the underlying scene $\mathbf{J}$. \\

\noindent\textbf{Rain-robust loss:}
To derain an image $\mathbf{I}$, one may directly learn a map from $\mathbf{I}$ to $\hat{\mathbf{J}}$ simply by minimizing the discrepancies between $\hat{\mathbf{J}}$ and the ground truth $\mathbf{J}$, i.e. an image reconstruction loss -- such is the case for existing methods. Under this formulation, the model must explore a large hypothesis space, e.g. any region obfuscated by rain streaks is inherently ambiguous, making learning difficult.

Unlike previous works, we constrain the learned representation such that it is robust to rain phenomena. To ``learn away'' the rain, we propose to map both the rainy and clean images of the same scene to an embedding space where they are close to each other by optimizing a similarity metric. Additionally, we minimize a reconstruction objective to ensure that the learned representation is sufficient to recover the underlying scene. Our approach is inspired by the recent advances in contrastive learning~\cite{chen2020simple}, and we aim to distill rain-robust representations of real-world scenes by directly comparing the rainy and clean images in the feature space. But unlike \cite{chen2020simple}, we do not define a positive pair as augmentation to the same image, but rather any rainy image and its corresponding clean image from the same scene.

When training, we first randomly sample a mini-batch of $N$ rainy images with the associated clean images to form an augmented batch $\{(\mathbf{I}_i, \mathbf{J}_i)\}_{i=1}^N$, where $\mathbf{I}_i$ is the $i$-th rainy image, and $\mathbf{J}_i$ is its corresponding ground-truth image. This augmented batch is fed into a shared-weight feature extractor $\mathcal{F}_E(\cdot, \theta_E)$ with weights $\theta_E$ to obtain a feature set $\{(\mathbf{z}_{\mathbf{I}_i}, \mathbf{z}_{\mathbf{J}_i})\}_{i=1}^N$, where $\mathbf{z}_{\mathbf{I}_i} = \mathcal{F}_E(\mathbf{I}_i, \theta_E)$ and $\mathbf{z}_{\mathbf{J}_i} = \mathcal{F}_E(\mathbf{J}_i, \theta_E)$. We consider every $(\mathbf{z}_{\mathbf{I}_i}, \mathbf{z}_{\mathbf{J}_i})$ as the positive pairs. This is so that the learned features from the same scene should be close to each other regardless of the rainy conditions. We treat the other $2(N - 1)$ samples from the same batch as negative samples. Based on the noise-contrastive estimation (NCE)~\cite{gutmann2010noise}, we adopt the following InfoNCE~\cite{oord2018representation} criterion to measure the rain-robust loss for a positive pair $(\mathbf{z}_{\mathbf{J}_i}, \mathbf{z}_{\mathbf{I}_i})$:
\begin{equation} \label{eq:rain_robust_loss}
    \ell_{\mathbf{z}_{\mathbf{J}_i}, \mathbf{z}_{\mathbf{I}_i}} = -\log \frac{\exp \Big(\text{sim}_{\text{cos}}(\mathbf{z}_{\mathbf{I}_i}, \mathbf{z}_{\mathbf{J}_i}) / \tau \Big)}{\sum_{\mathbf{k} \in \mathcal{K}} \exp \Big(\text{sim}_{\text{cos}}(\mathbf{z}_{\mathbf{J}_i}, \mathbf{k}) / \tau \Big)},
\end{equation}
where $\mathcal{K} = \{\mathbf{z}_{\mathbf{I}_j}, \mathbf{z}_{\mathbf{J}_j}\}_{j=1, j \neq i}^{N}$ is a set that contains the features extracted from other rainy and ground-truth images in the selected mini-batch, $\text{sim}_{\text{cos}}(\mathbf{u}, \mathbf{v}) = \mathbf{u}^\intercal \mathbf{v} / \norm{\mathbf{u}} \norm{\mathbf{v}}$ is the cosine similarity between two feature vectors $\mathbf{u}$ and $\mathbf{v}$, and $\tau$ is the temperature parameter~\cite{wu2018unsupervised}. We set $\tau$ as 0.25, and this loss is calculated across all positive pairs within the mini-batch for both $(\mathbf{z}_{\mathbf{I}_i}, \mathbf{z}_{\mathbf{J}_i})$ and $(\mathbf{z}_{\mathbf{J}_i}, \mathbf{z}_{\mathbf{I}_i})$. \\

\noindent\textbf{Full objective:}
While minimizing~\cref{eq:rain_robust_loss} maps features of clean and rainy images to the same subspace, we also need to ensure that the representation is sufficient to reconstruct the scene. Hence, we additionally minimize a Multi-Scale Structural Similarity Index (MS-SSIM)~\cite{wang2003multiscale} loss and a $\ell1$ image reconstruction loss to prevent the model from discarding useful information for the reconstruction task. Our full objective $\mathcal{L}_{\text{full}}$ is as follows:
\begin{equation} \label{eq:full_objective}
    \mathcal{L}_{\text{full}}(\hat{\mathbf{J}}, \mathbf{J}) = \mathcal{L}_{\text{MS-SSIM}}(\hat{\mathbf{J}}, \mathbf{J}) + \lambda_{\ell1} \mathcal{L}_{\ell1}(\hat{\mathbf{J}}, \mathbf{J}) + \lambda_{\text{robust}} \mathcal{L}_{\text{robust}}(\mathbf{z}_{\mathbf{J}}, \mathbf{z}_{\mathbf{I}}),
\end{equation}
where $\mathcal{L}_{\text{MS-SSIM}}(\cdot)$ is the MS-SSIM loss that is commonly used for image restoration~\cite{zhao2016loss}, $\mathcal{L}_{\ell1}(\cdot)$ is the $\ell1$ distance between the estimated clean images $\hat{\mathbf{J}}$ and the ground-truth images $\mathbf{J}$, $\mathcal{L}_{\text{robust}}(\cdot)$ is the rain-robust loss in~\cref{eq:rain_robust_loss}, and $\lambda_{\ell1}$ and $\lambda_{\text{robust}}$ are two hyperparameters to control the relative importance of different loss terms. In our experiments, we set both $\lambda_{\ell1}$ and $\lambda_{\text{robust}}$ as 0.1. \\

\noindent\textbf{Network architecture \& implementation details:}
We design our model based on the architecture introduced in~\cite{johnson2016perceptual,zhu2017unpaired}. As illustrated in~\cref{fig:pipeline}, our network includes an encoder of one input convolutional block, two downsampling blocks, and nine residual blocks~\cite{he2016deep} to yield latent features $\mathbf{z}$. This is followed by a decoder of two upsampling blocks and one output layer to map the features to $\mathbf{J}$. We fuse skip connections into the decoder using $3 \times 3$ up-convolution blocks to retain information lost in the bottleneck.
Note: normal convolution layers are replaced by deformable convolution~\cite{zhu2019deformable} in our residual blocks -- in doing so, we enable our model to propagate non-local spatial information to reconstruct local degradations caused by rain effects. Latent features $\mathbf{z}$ are used for the rain-robust loss described in~\cref{eq:rain_robust_loss}. Since these features are high dimensional ($256 \times 64 \times 64$), we use an average pooling layer to condense the feature map of each channel to $2 \times 2$. The condensed features are flattened into a vector of length $1024$ for the rain-robust loss. It is worth noting that our rain-robust loss does not require additional modifications on the model architectures. 

Our deraining model is trained on $256 \times 256$ patches and a mini-batch size $N=8$ for 20 epochs. We use the Adam optimizer~\cite{kingma2014adam} with $\beta_1 = 0.9$ and $\beta_2 = 0.999$. The initial learning rate is $2 \times 10^{-4}$, and it is steadily modified to $1 \times 10^{-6}$ based on a cosine annealing schedule~\cite{loshchilov2017sgdr}. We also use a linear warm-up policy for the first 4 epochs. For data augmentation, we use random cropping, random rotation, random horizontal and vertical flips, and RainMix augmentation~\cite{guo2021efficientderain}. More details can be found in the supplementary material.

\section{Experiments} \label{sec:experiments}

We compare to state-of-the-art methods both quantitatively and qualitatively on \dname, and qualitatively Internet rainy images~\cite{wei2019semi}. To quantify the difference between the derained results and ground-truth, we adopt peak signal-to-noise ratio (PSNR)~\cite{huynh2008scope} and structure similarity (SSIM)~\cite{wang2004image}. \\ 

\begin{table*}[t]
\caption{\textbf{Quantitative comparison on \dname.} Our method outperforms the existing state-of-the-art derainers. The preferred results are marked in \textbf{bold}.}
  \label{tab:our_real_results} 
  \centering
  \scriptsize
  \resizebox{\columnwidth}{!}{
  \begin{tabular}{ccccccccccc}
    \toprule
    \makecell{Data Split} & Metrics & \makecell{Rainy \\ Images} & \makecell{SPANet~\cite{wang2019spatial} \\ (CVPR'19)} & \makecell{HRR~\cite{li2019heavy} \\ (CVPR'19)} & \makecell{MSPFN~\cite{jiang2020multi} \\ (CVPR'20)} & \makecell{RCDNet~\cite{wang2020a} \\ (CVPR'20)} & \makecell{DGNL-Net~\cite{hu2021single} \\ (IEEE TIP'21)} & \makecell{EDR~\cite{guo2021efficientderain} \\ (AAAI'21)} & \makecell{MPRNet~\cite{zamir2021multi} \\ (CVPR'21)} & \makecell{Ours} \\
    \midrule
    \makecell{Dense Rain \\ Streaks} & \makecell{PSNR$\uparrow$ \\ SSIM$\uparrow$} & \makecell{18.46 \\ 0.6284} & \makecell{18.87 \\ 0.6314} & \makecell{17.86 \\ 0.5872} & \makecell{19.58 \\ 0.6342} & \makecell{19.50 \\ 0.6218} & \makecell{17.33 \\ 0.5947} & \makecell{18.86 \\ 0.6296} & \makecell{19.12 \\ 0.6375} & \makecell{\textbf{20.84} \\ \textbf{0.6573}} \\
    \midrule
    \makecell{Dense Rain \\ Accumulation} & \makecell{PSNR$\uparrow$ \\ SSIM$\uparrow$} & \makecell{20.87 \\ 0.7706} & \makecell{21.42 \\ 0.7696} & \makecell{14.82 \\ 0.4675} & \makecell{21.13 \\ 0.7735} & \makecell{21.27 \\ 0.7765} & \makecell{20.75 \\ 0.7429} & \makecell{21.07 \\ 0.7766} & \makecell{21.38 \\ 0.7808} & \makecell{\textbf{24.78} \\ \textbf{0.8279}} \\
    \midrule
    \makecell{Overall} & \makecell{PSNR$\uparrow$ \\ SSIM$\uparrow$} & \makecell{19.49 \\ 0.6893} & \makecell{19.96 \\ 0.6906} & \makecell{16.55 \\ 0.5359} & \makecell{20.24 \\ 0.6939} & \makecell{20.26 \\ 0.6881} & \makecell{18.80 \\ 0.6582} & \makecell{19.81 \\ 0.6926} & \makecell{20.09 \\ 0.6989} & \makecell{\textbf{22.53} \\ \textbf{0.7304}} \\
    \bottomrule
  \end{tabular}}
\end{table*}

\noindent\textbf{Quantitative evaluation on \dname:}
To quantify the sim2real gap of the existing datasets, we test seven representative existing state-of-the-art methods~\cite{guo2021efficientderain,hu2021single,jiang2020multi,li2019heavy,wang2020a,wang2019spatial,zamir2021multi} on our \dname\ test set.\footnote{We use the original code and network weights from the authors for comparison. Code links for all comparison methods are provided in the supplementary material.} Since there exist numerous synthetic datasets proposed by previous works~\cite{fu2017removing,hu2019depth,li2019heavy,li2016rain,yang2017deep,zhang2018density,zhang2019image}, we found it intractable to train our method on each one; whereas, it is more feasible to take the best derainers for each respective dataset and test on our proposed dataset as a proxy (\cref{tab:our_real_results}). This follows the conventions of previous deraining dataset papers~\cite{fu2017clearing,hu2021single,li2016rain,wang2019spatial,yang2020single,zhang2018density,zhang2019image} to compare with top performing methods from each existing dataset. 

SPANet~\cite{wang2019spatial} is trained on SPA-Data~\cite{wang2019spatial}. HRR~\cite{li2019heavy} utilizes both NYU-Rain~\cite{li2019heavy} and Outdoor-Rain~\cite{li2019heavy}. MSPFN~\cite{jiang2020multi} and MPRNet~\cite{zamir2021multi} are trained on a combination of multiple synthetic datasets~\cite{fu2017removing,li2016rain,yang2017deep,zhang2019image}. DGNL-Net~\cite{hu2021single} is trained on RainCityscapes~\cite{hu2019depth}. For RCDNet~\cite{wang2020a} and EDR~\cite{guo2021efficientderain}, multiple weights from different training sets are provided. We choose RCDNet trained on SPA-Data and EDR V4 trained on Rain14000~\cite{fu2017removing} due to superior performance. 

Compared to training on \dname\ (ours), methods trained on other data perform worse, with the largest domain gap being in NYU-Rain and Outdoor-Rain (HRR) and RainCityscapes (DGNL). Two trends do hold: training on (1) more synthetic data gives better results (MSPFN, MPRNet) and (2) semi-real data also helps (SPANet). However, even when multiple synthetic~\cite{fu2017removing,li2016rain,yang2017deep,zhang2019image} or semi-real~\cite{wang2019spatial} datasets are used, their performance on real data is still around 2dB lower than training on \dname\ (ours). 

\cref{fig:our_real_results} illustrates some representative derained images across scenarios with various rain appearance and rain accumulation densities. Training on \dname\ enables the network to remove most rain streaks and rain accumulation; whereas, training on synthetic/semi-real data tends to leave visible rain streaks. We note that HRR~\cite{li2019heavy} and DGNL~\cite{hu2021single} may seem like they remove rain accumulation, but they in fact introduce undesirable artifacts, e.g. dark spots on the back of the traffic sign, tree, and sky. The strength of having ground-truth paired data is demonstrated by our 2.44 dB gain compared to the state of the art~\cite{zamir2021multi}. On test images with dense rain accumulation, the boost improves to 3.40 dB. 

\newcommand{\figwidth}{.19}
\newcommand{\Qwidth}{\figwidth\textwidth}
\begin{figure*}[!ht]
    \captionsetup[subfloat]{font=scriptsize,farskip=3pt,captionskip=2pt}
    \centering
        

    
    
    \subfloat[\centering Rain (23.64/0.8561)]{\includegraphics[width=\Qwidth]{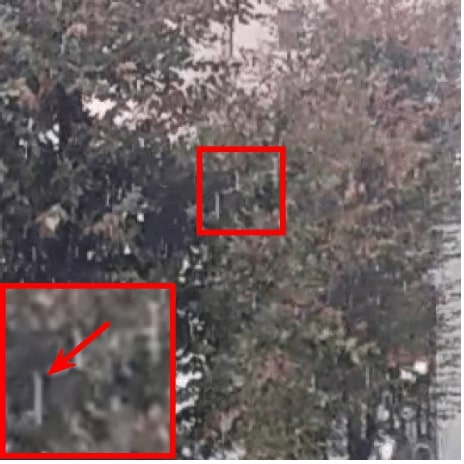}}
    \hfill
    \subfloat[\centering SPANet~\cite{wang2019spatial} (23.56/0.8474)]{\includegraphics[width=\Qwidth]{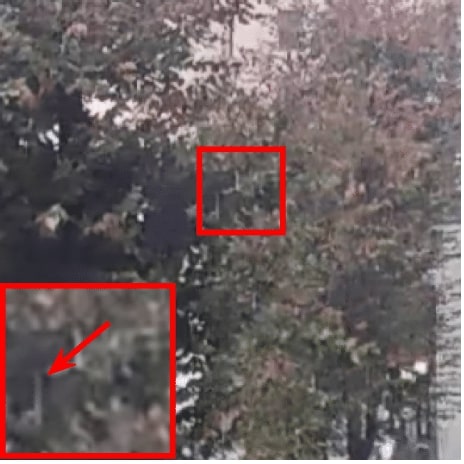}}
    \hfill
    \subfloat[\centering HRR~\cite{li2019heavy} (19.78/0.7508)]{\includegraphics[width=\Qwidth]{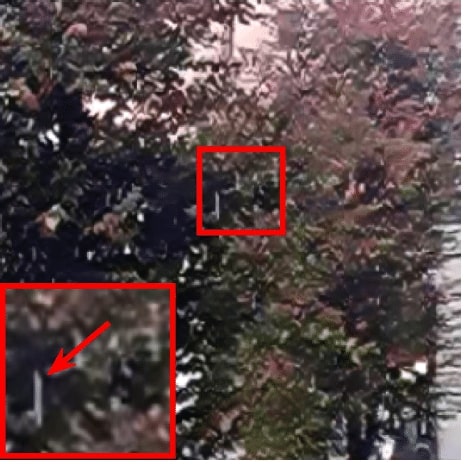}}
    \hfill
    \subfloat[\centering MSPFN~\cite{jiang2020multi} (25.57/0.8659)]{\includegraphics[width=\Qwidth]{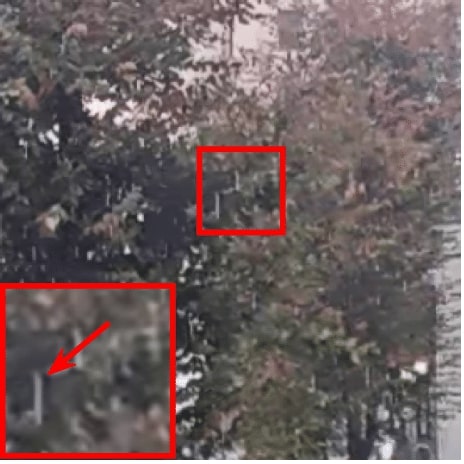}}
    \hfill
    \subfloat[\centering RCDNet~\cite{wang2020a} (24.71/0.8654)]{\includegraphics[width=\Qwidth]{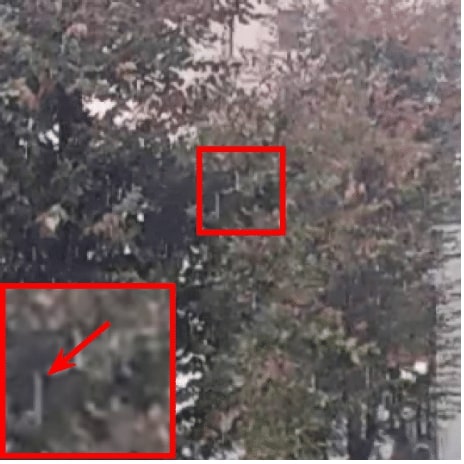}}
        
    \par

    \subfloat[\centering DGNL~\cite{hu2021single} (17.26/0.7516)]{\includegraphics[width=\Qwidth]{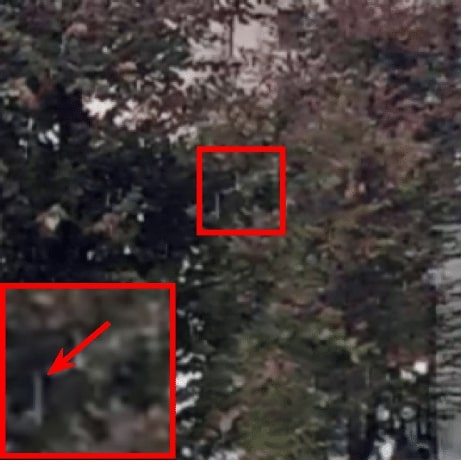}}
    \hfill
    \subfloat[\centering EDR V4~\cite{guo2021efficientderain} (23.93/0.8539)]{\includegraphics[width=\Qwidth]{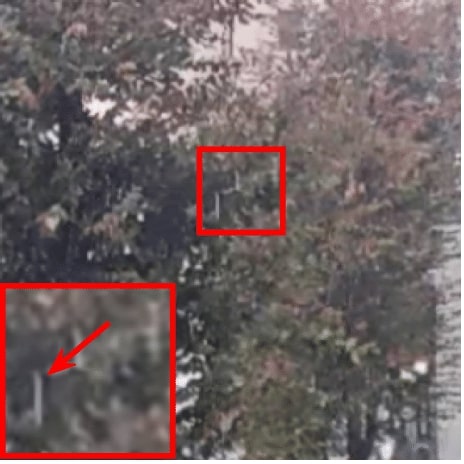}}
    \hfill
    \subfloat[\centering MPRNet~\cite{zamir2021multi} (24.33/0.8657)]{\includegraphics[width=\Qwidth]{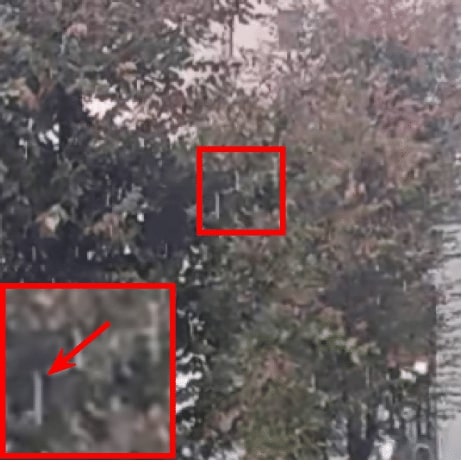}}
    \hfill
    \subfloat[\centering \textbf{Ours (26.31/0.8763)}]{\includegraphics[width=\Qwidth]{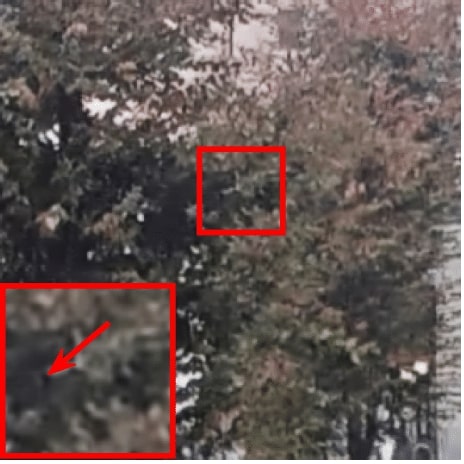}}
    \hfill
    \subfloat[\centering Ground Truth  (PSNR/SSIM)]{\includegraphics[width=\Qwidth]{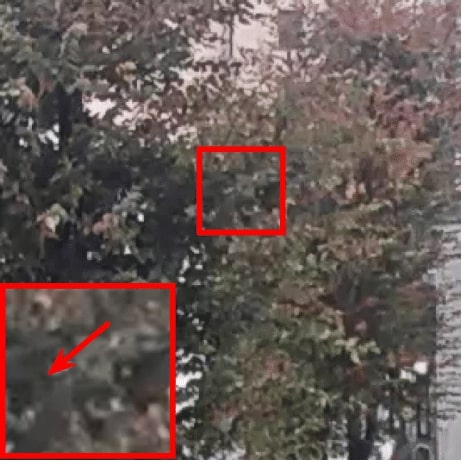}}
    
    \par
    
    \subfloat[\centering Rain (19.81/0.7541)]{\includegraphics[width=\Qwidth]{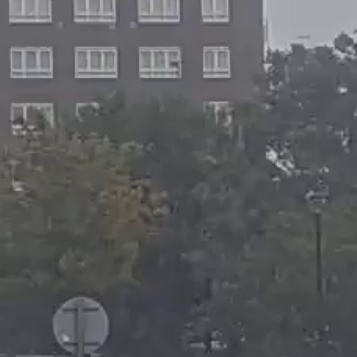}}
    \hfill
    \subfloat[\centering SPANet~\cite{wang2019spatial} (20.03/0.7244)]{\includegraphics[width=\Qwidth]{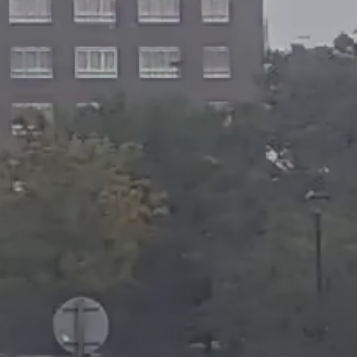}}
    \hfill
    \subfloat[\centering HRR~\cite{li2019heavy} (15.03/0.4944)]{\includegraphics[width=\Qwidth]{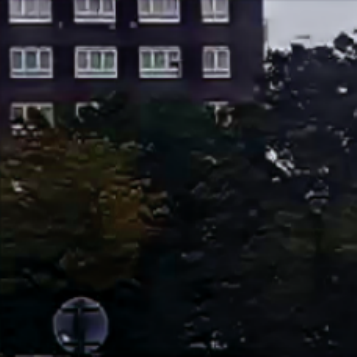}}
    \hfill
    \subfloat[\centering MSPFN~\cite{jiang2020multi} (19.64/0.7491)]{\includegraphics[width=\Qwidth]{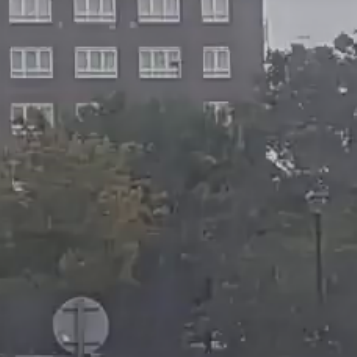}}
    \hfill
    \subfloat[\centering RCDNet~\cite{wang2020a} (20.58/0.7164)]{\includegraphics[width=\Qwidth]{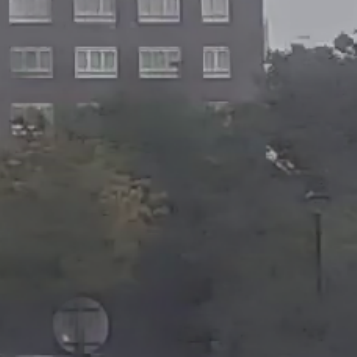}}
    
    \par

    \subfloat[\centering DGNL~\cite{hu2021single} (15.51/0.6508)]{\includegraphics[width=\Qwidth]{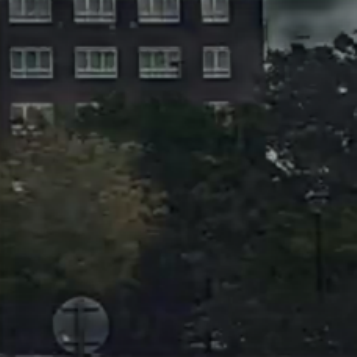}}
    \hfill
    \subfloat[\centering EDR V4~\cite{guo2021efficientderain}
    (19.96/0.7461) ]{\includegraphics[width=\Qwidth]{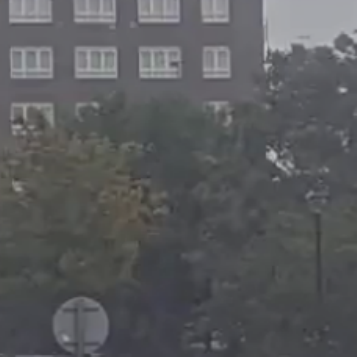}}
    \hfill
    \subfloat[\centering MPRNet~\cite{zamir2021multi}
    (19.88/0.7551) ]{\includegraphics[width=\Qwidth]{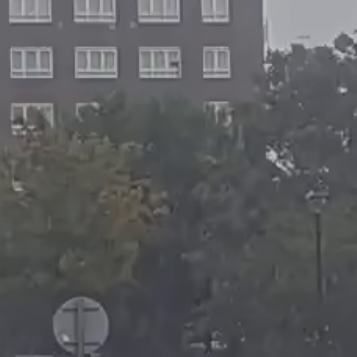}}
    \hfill
    \subfloat[\centering \textbf{Ours
    (23.89/0.7906)} ]{\includegraphics[width=\Qwidth]{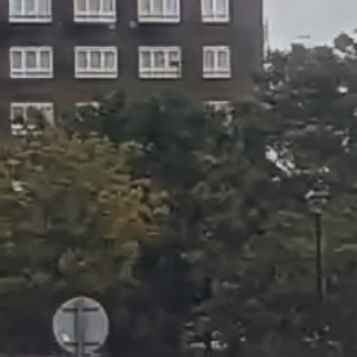}}
    \hfill
    \subfloat[\centering Ground Truth
    (PSNR/SSIM) ]{\includegraphics[width=\Qwidth]{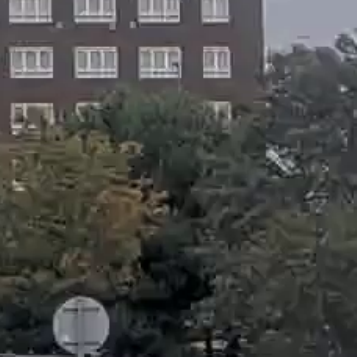}}
    
  \caption{\textbf{Our model simultaneously removes rain streaks and rain accumulation, while the existing models fail to generalize to real-world data.} The {\color{red}red arrows} highlight the difference between the proposed and existing methods on the \dname\ test set (zoom for details, PSNR and SSIM scores are listed below the images). }
\label{fig:our_real_results}
\end{figure*}

\begin{figure*}[t]
    \captionsetup[subfloat]{font=scriptsize,farskip=3pt,captionskip=2pt}
    \centering
    
    \subfloat[\centering Rainy Image]{\includegraphics[width=\Qwidth]{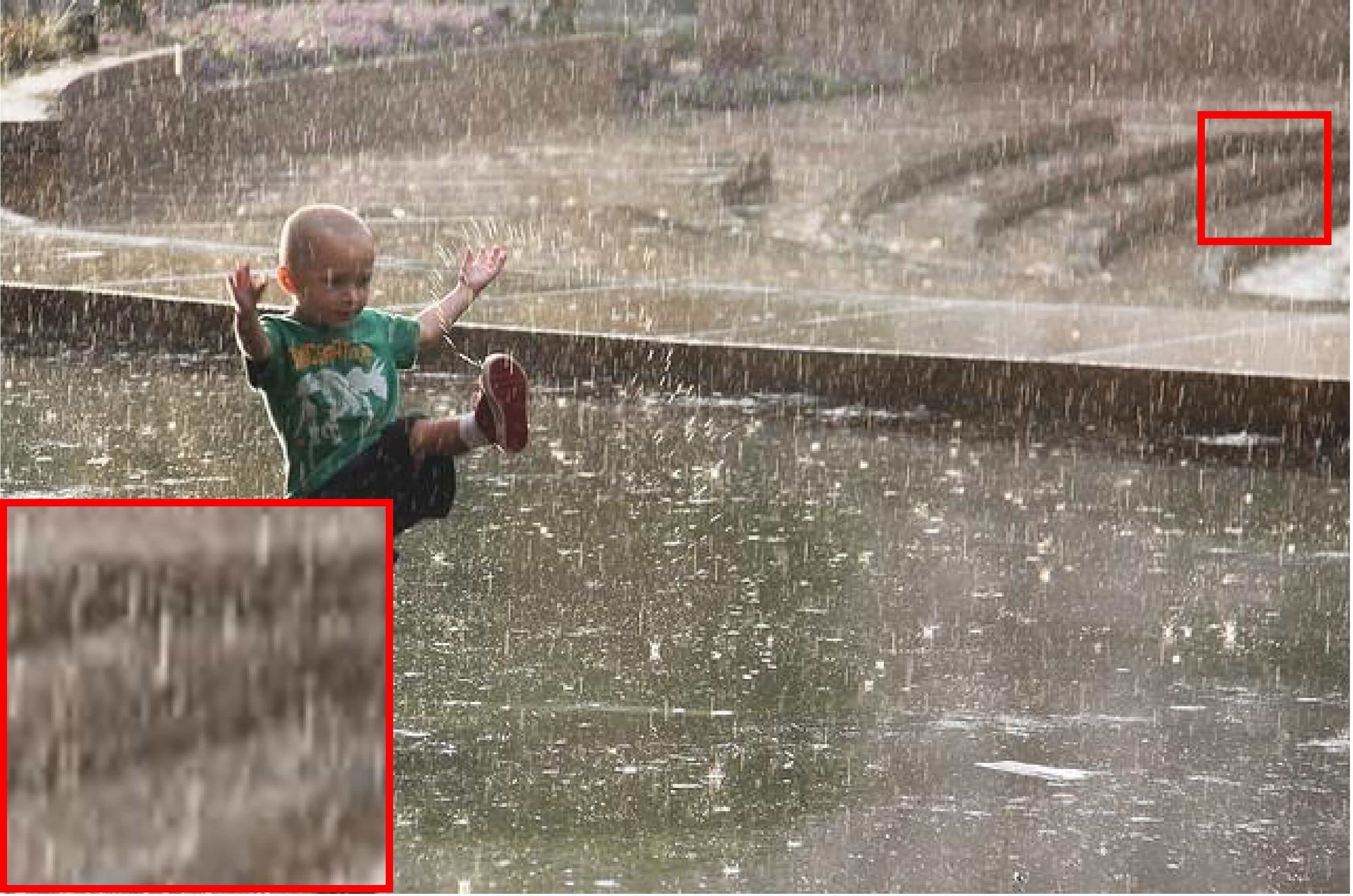}}
    \hfill
    \subfloat[\centering SPANet~\cite{wang2019spatial}]{\includegraphics[width=\Qwidth]{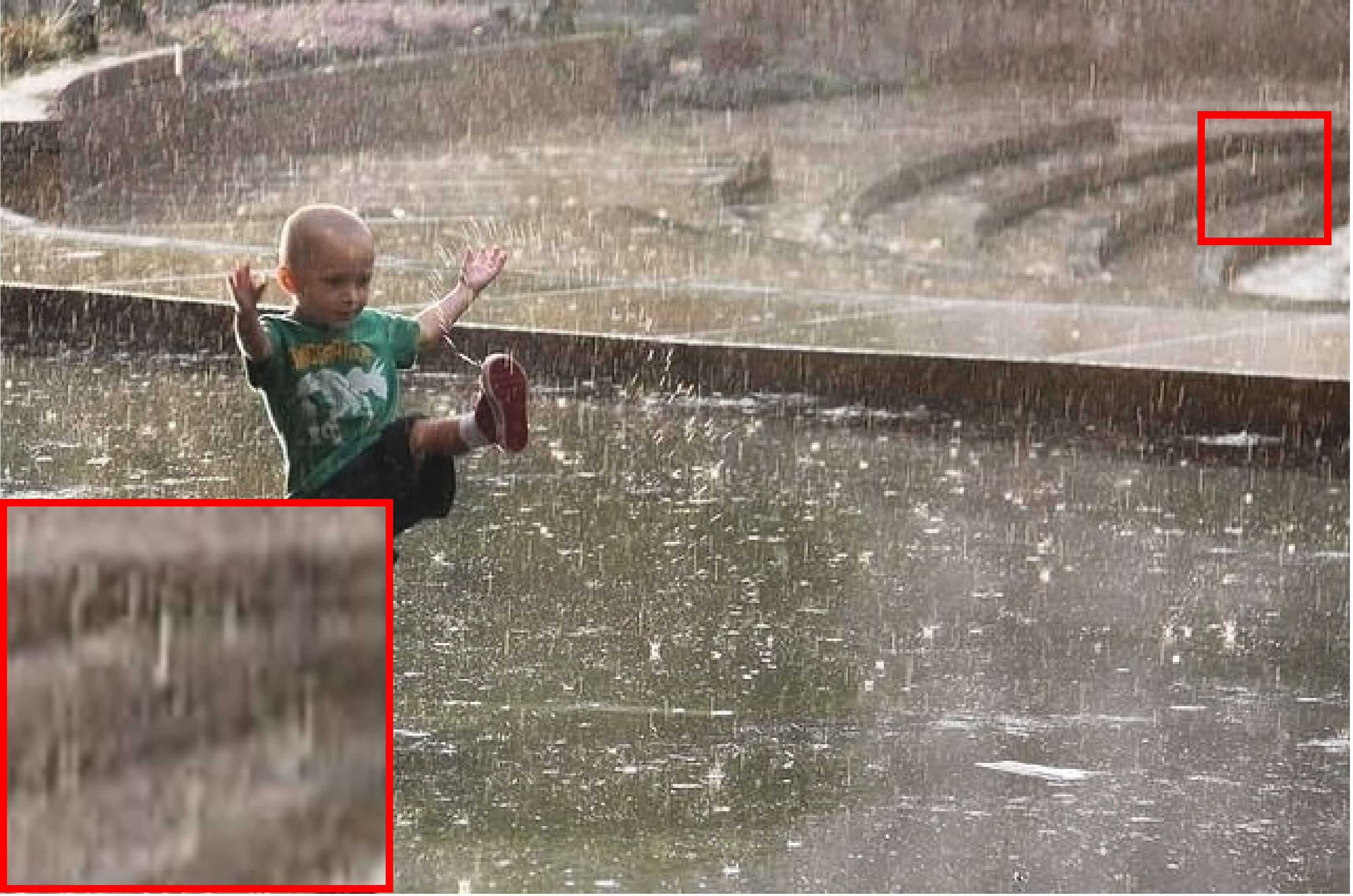}}
    \hfill
    \subfloat[\centering HRR~\cite{li2019heavy}]{\includegraphics[width=\Qwidth]{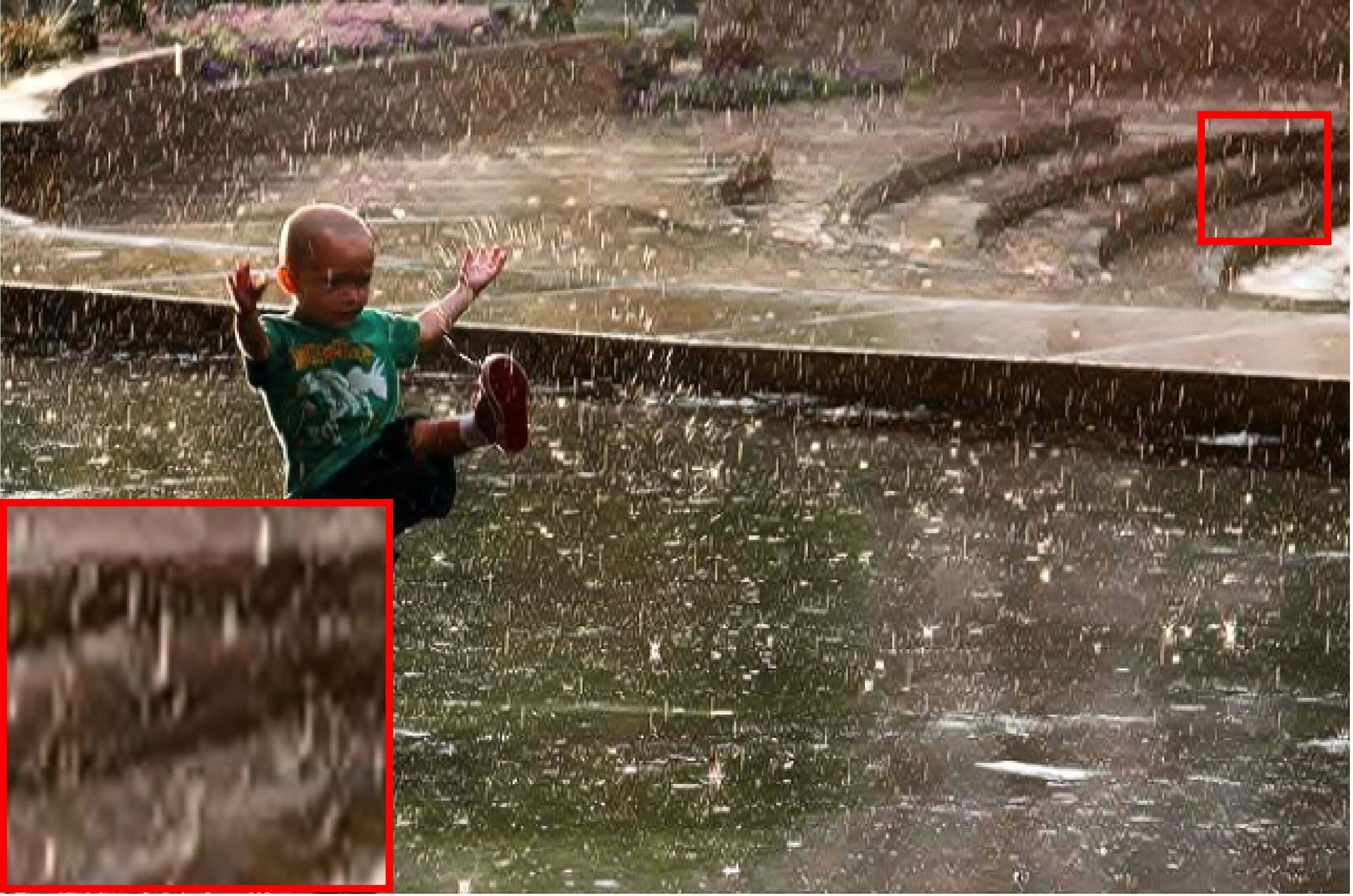}}
    \hfill
    \subfloat[\centering MSPFN~\cite{jiang2020multi}]{\includegraphics[width=\Qwidth]{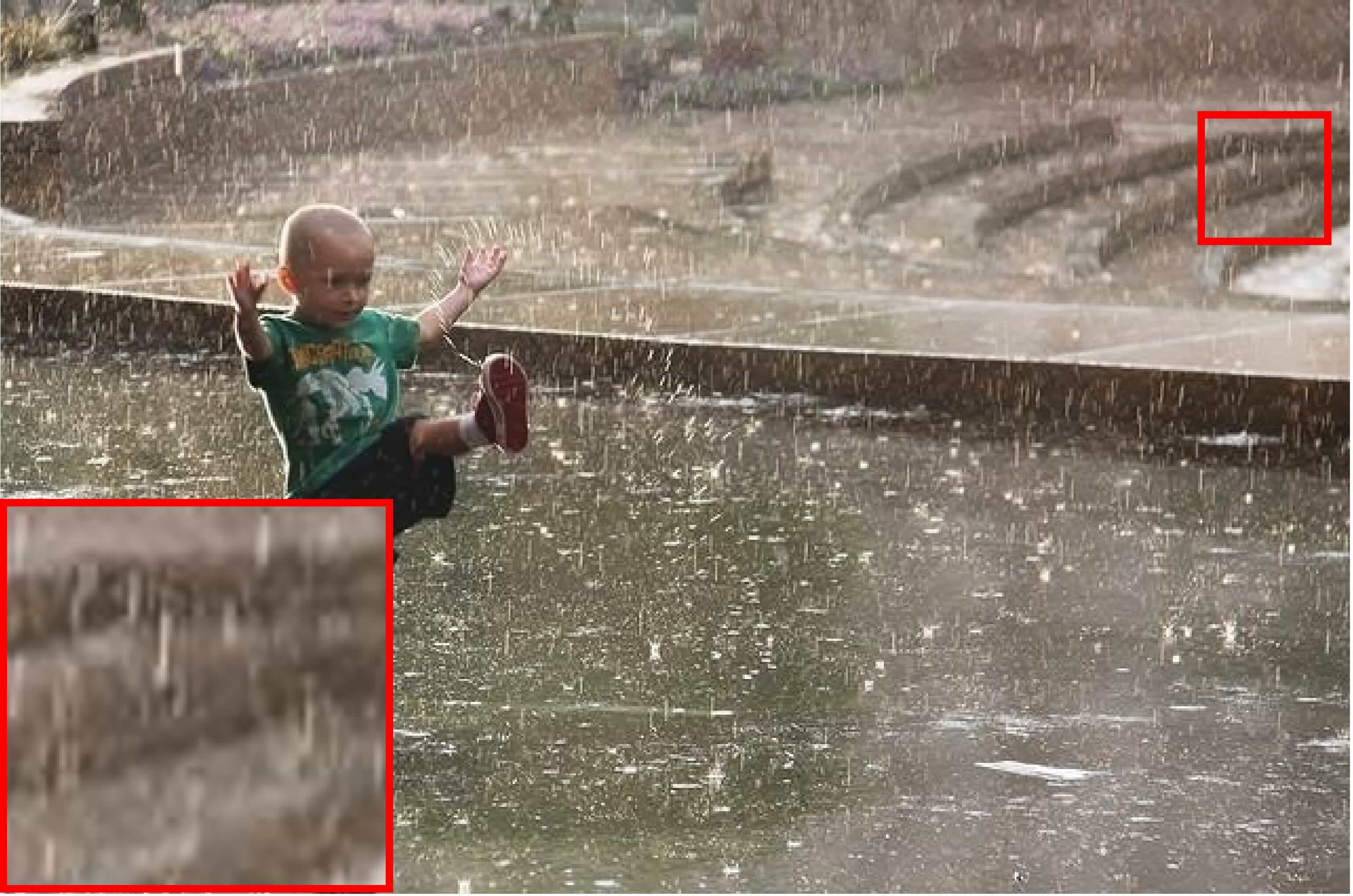}}
    \hfill
    \subfloat[\centering RCDNet~\cite{wang2020a}]{\includegraphics[width=\Qwidth]{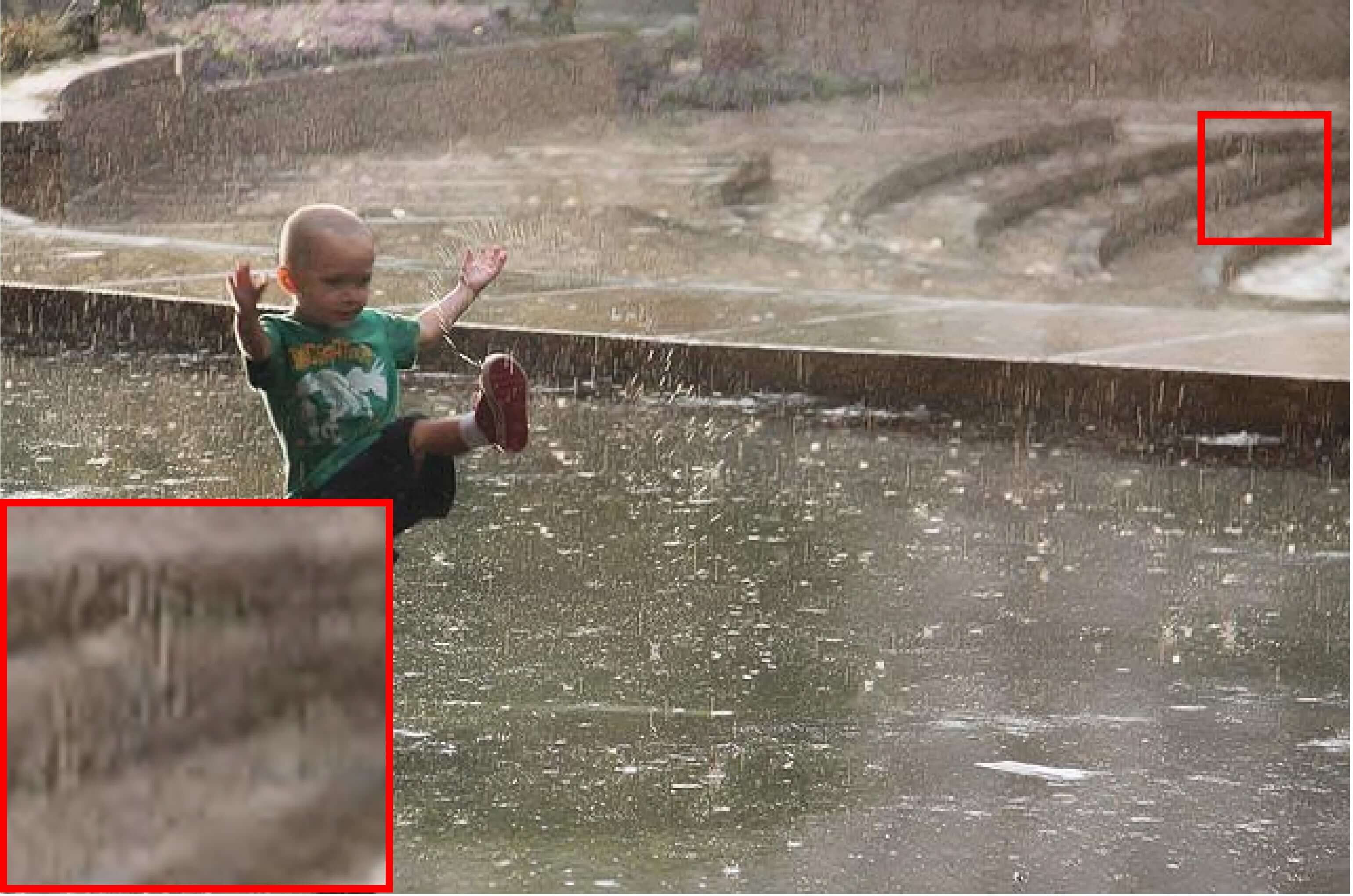}}
    
    \par
    
    \subfloat[\centering DGNL-Net~\cite{hu2021single}]{\includegraphics[width=\Qwidth]{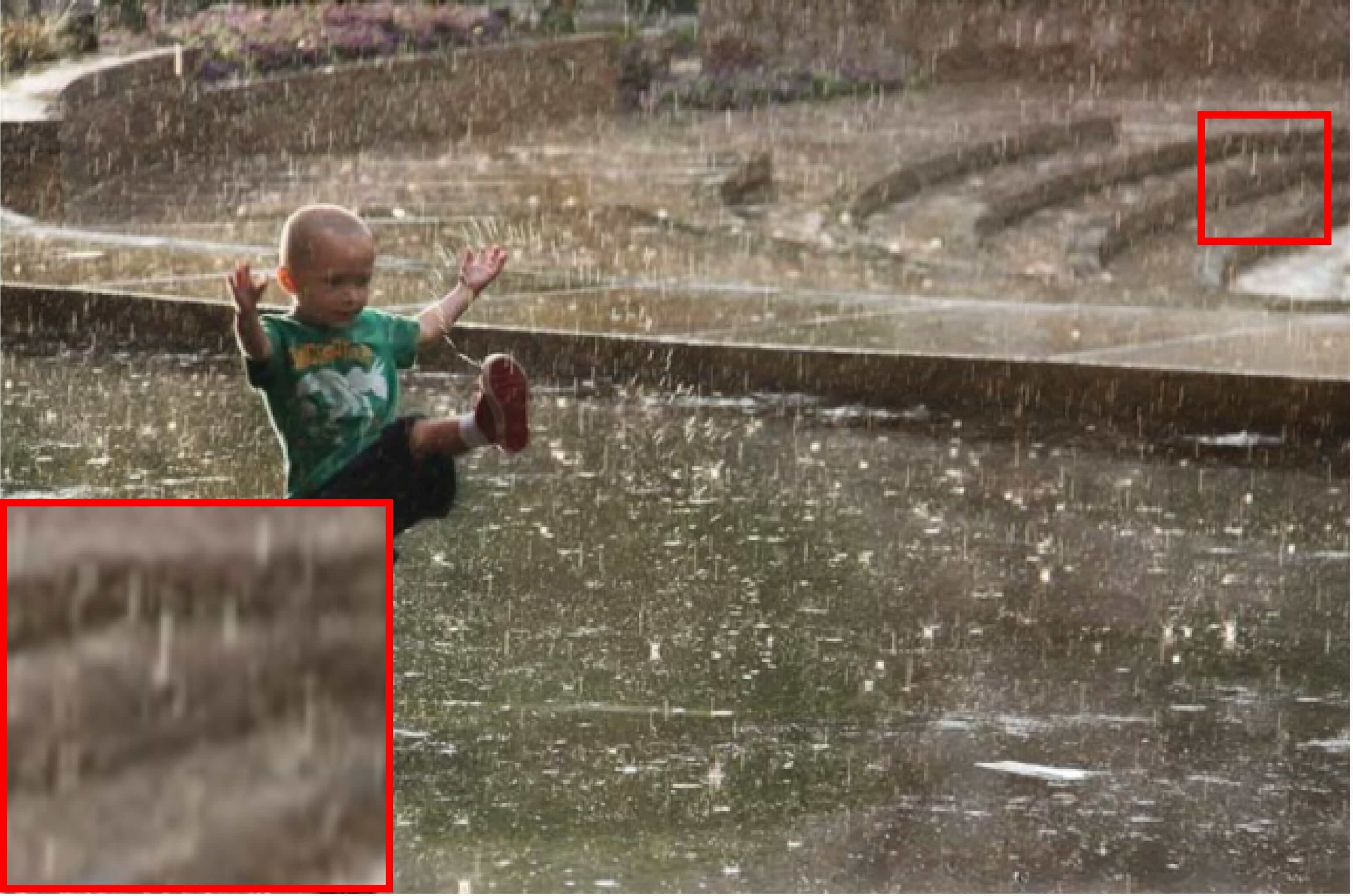}}
    \hfill
    \subfloat[\centering EDR V4 (S)~\cite{guo2021efficientderain}]{\includegraphics[width=\Qwidth]{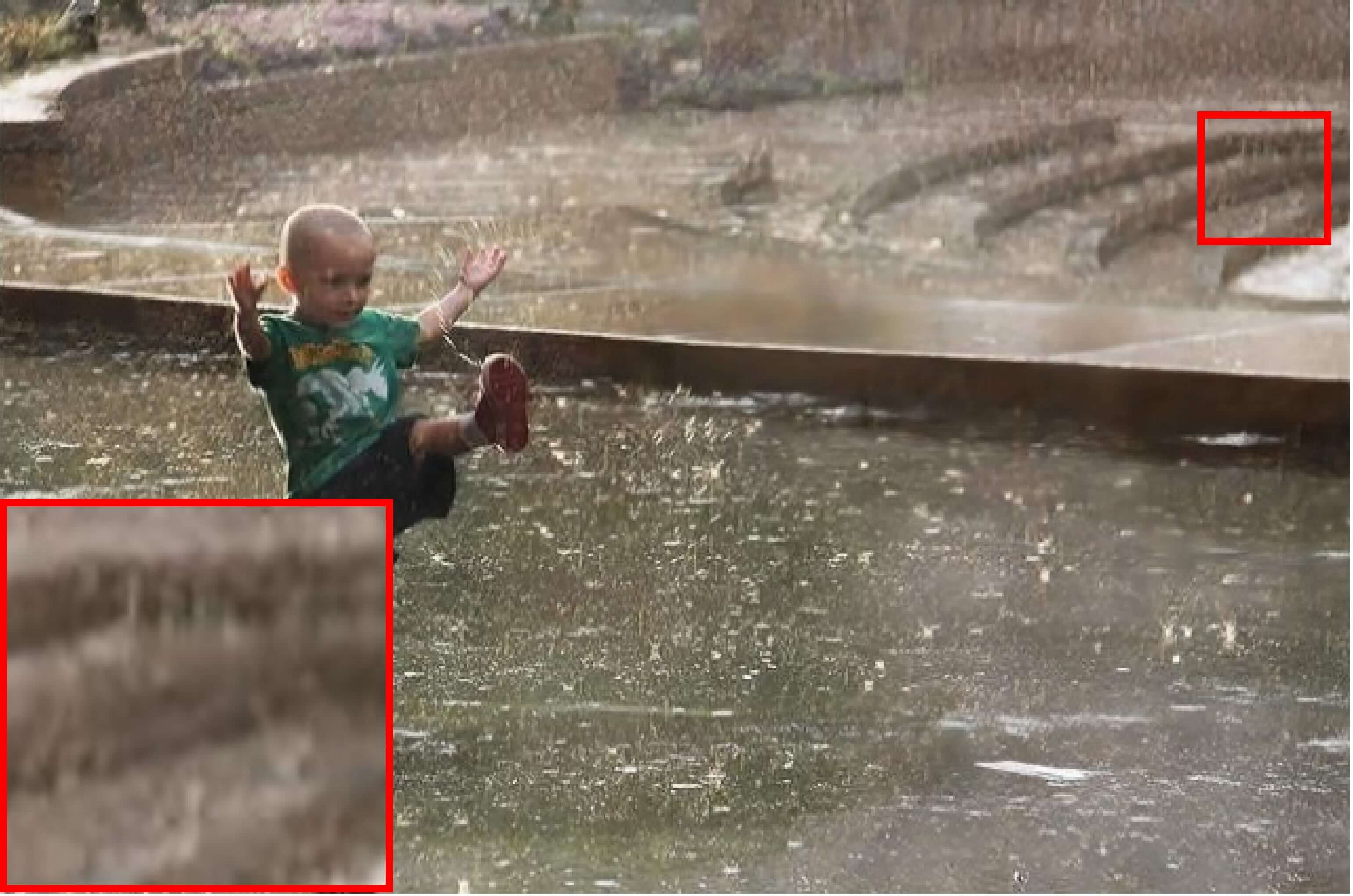}}
    \hfill
    \subfloat[\centering EDR V4 (R)~\cite{guo2021efficientderain}]{\includegraphics[width=\Qwidth]{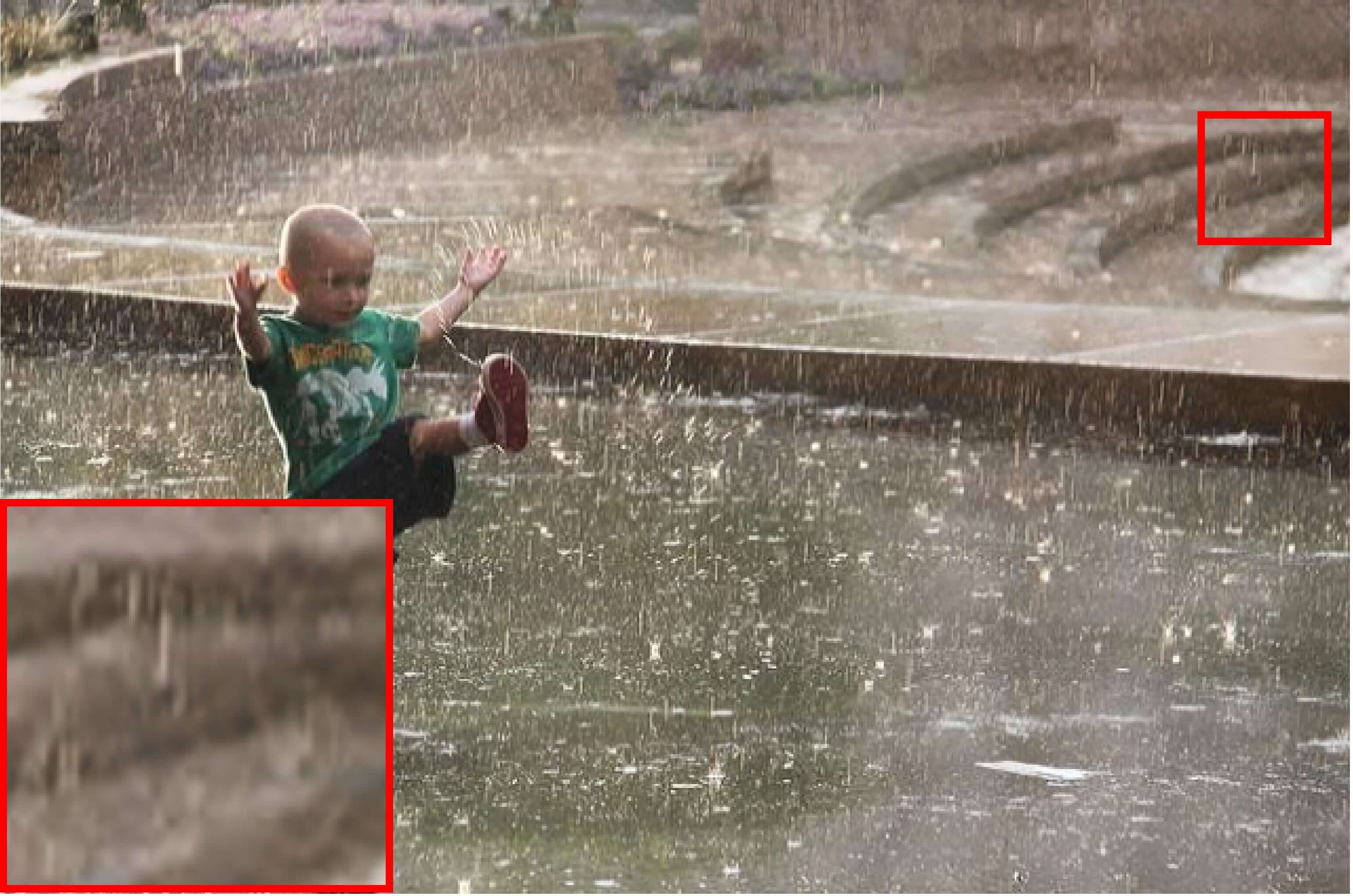}}
    \hfill
    \subfloat[\centering MPRNet~\cite{zamir2021multi}]{\includegraphics[width=\Qwidth]{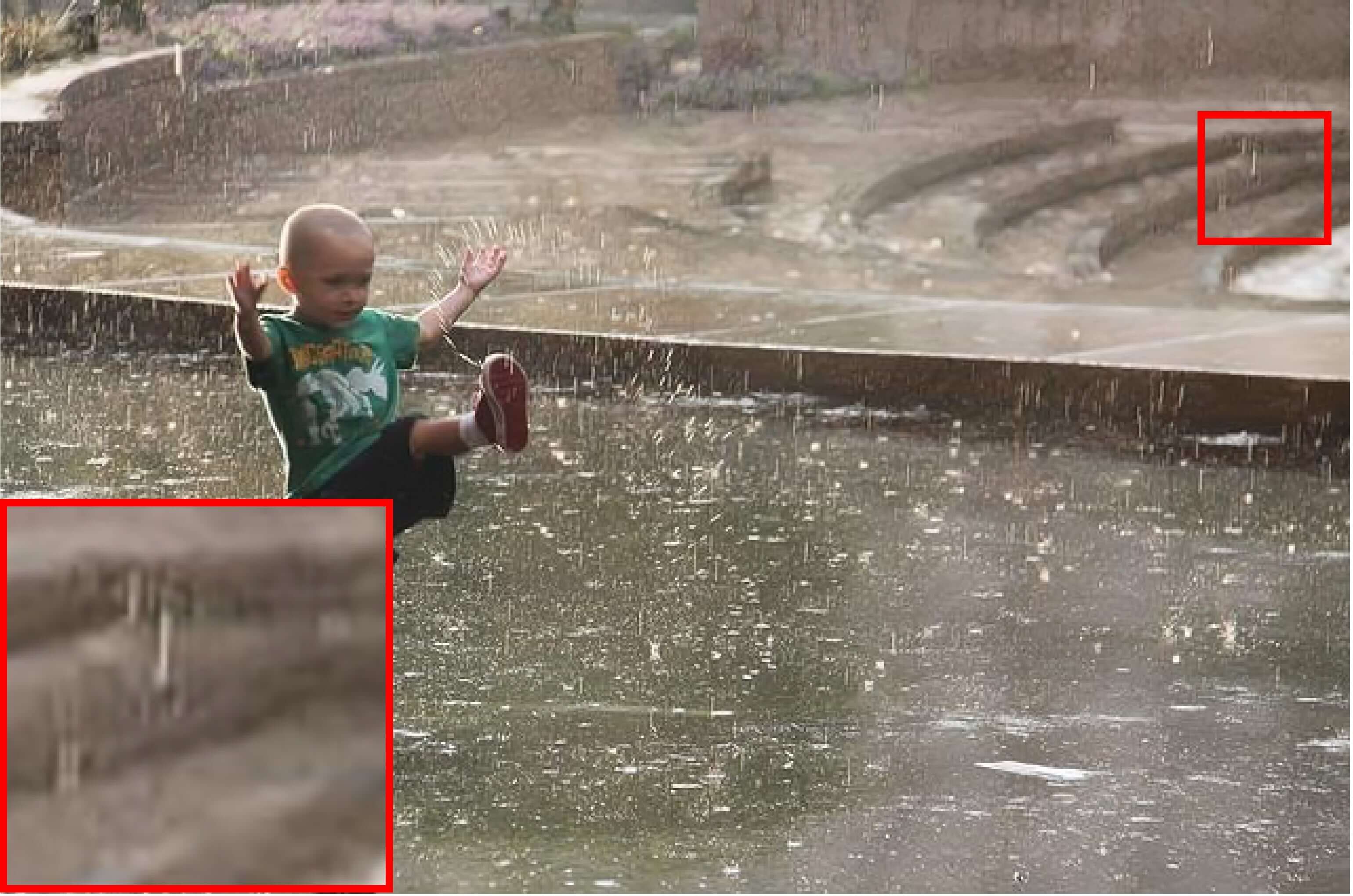}}
    \hfill
    \subfloat[\centering Ours]{\includegraphics[width=\Qwidth]{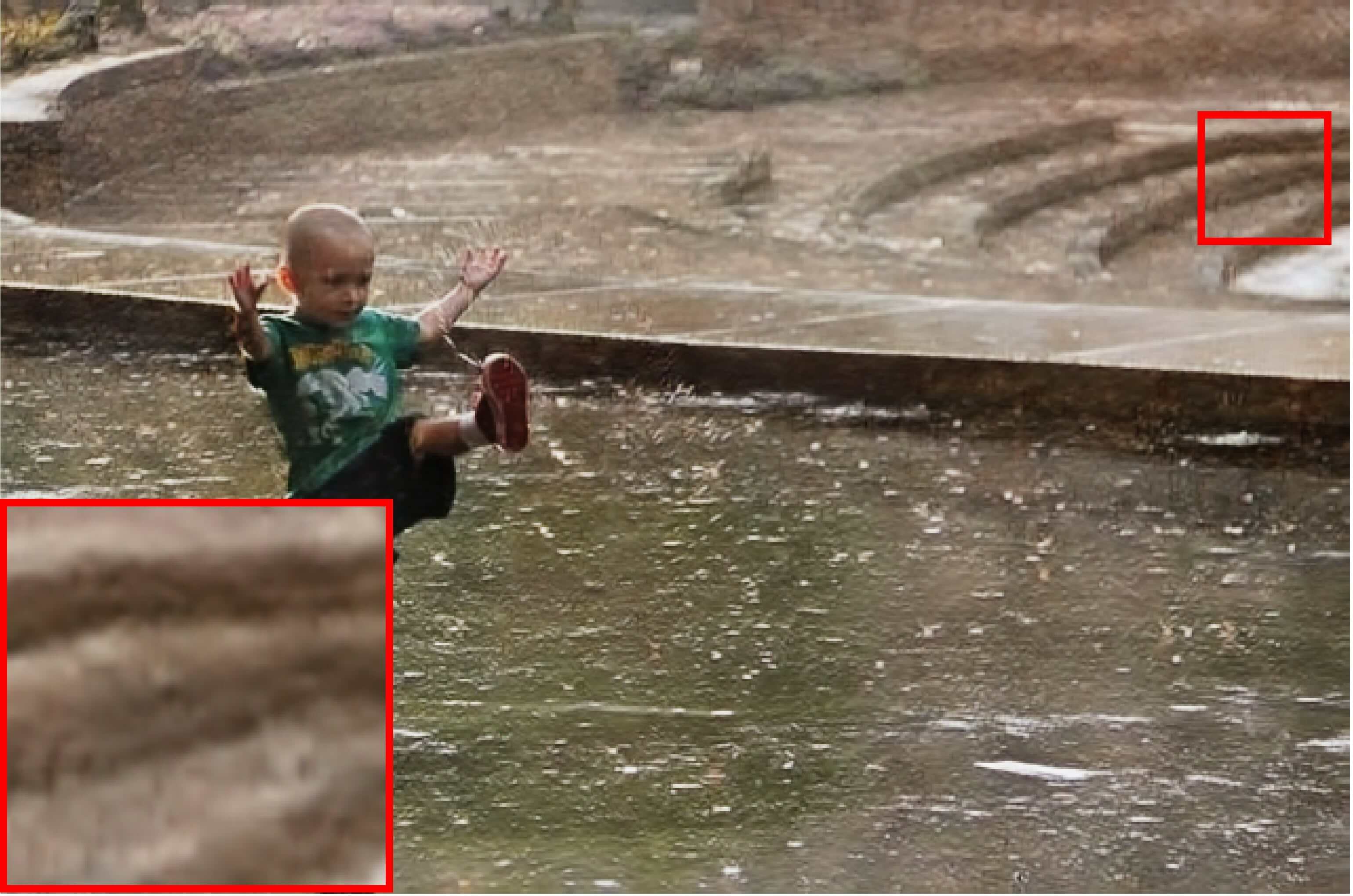}}
    

    
    \par
    
    \subfloat[\centering Rainy Image]{\includegraphics[width=\Qwidth]{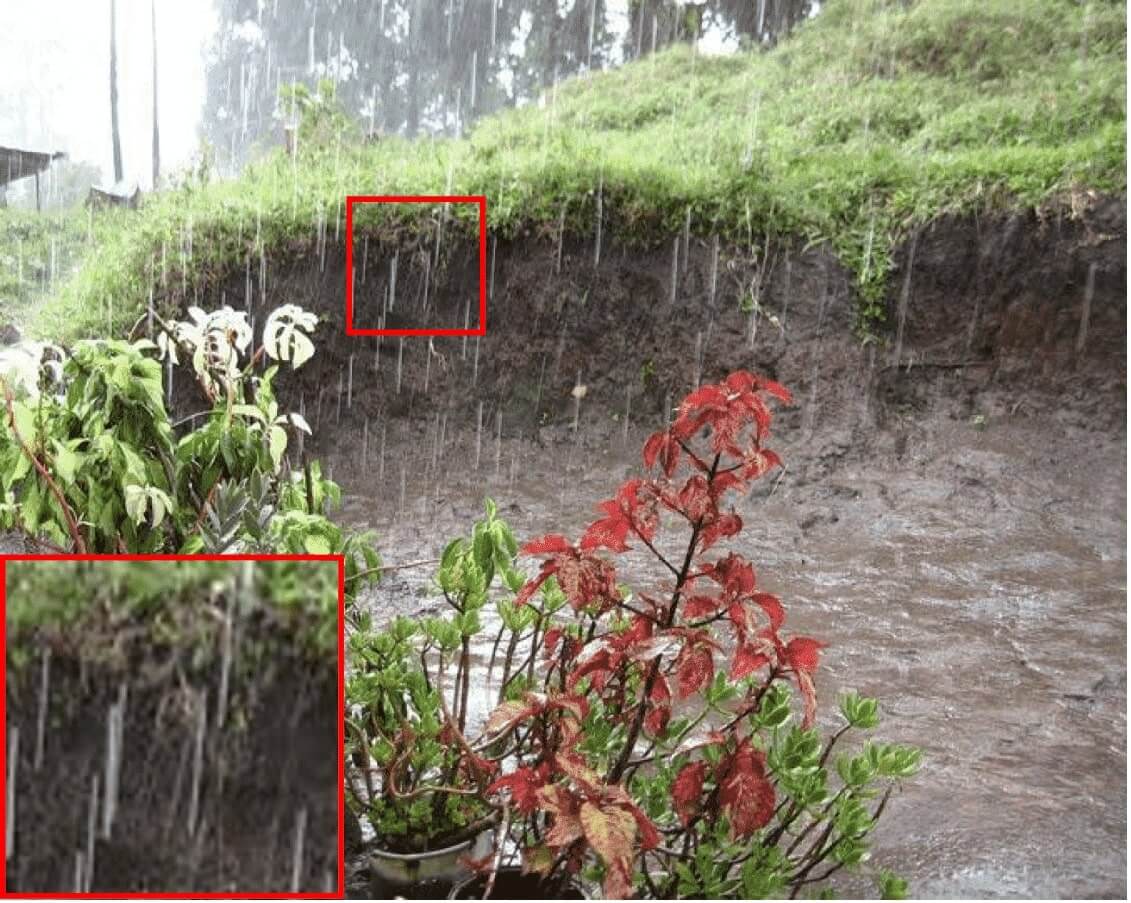}}
    \hfill
    \subfloat[\centering SPANet~\cite{wang2019spatial}]{\includegraphics[width=\Qwidth]{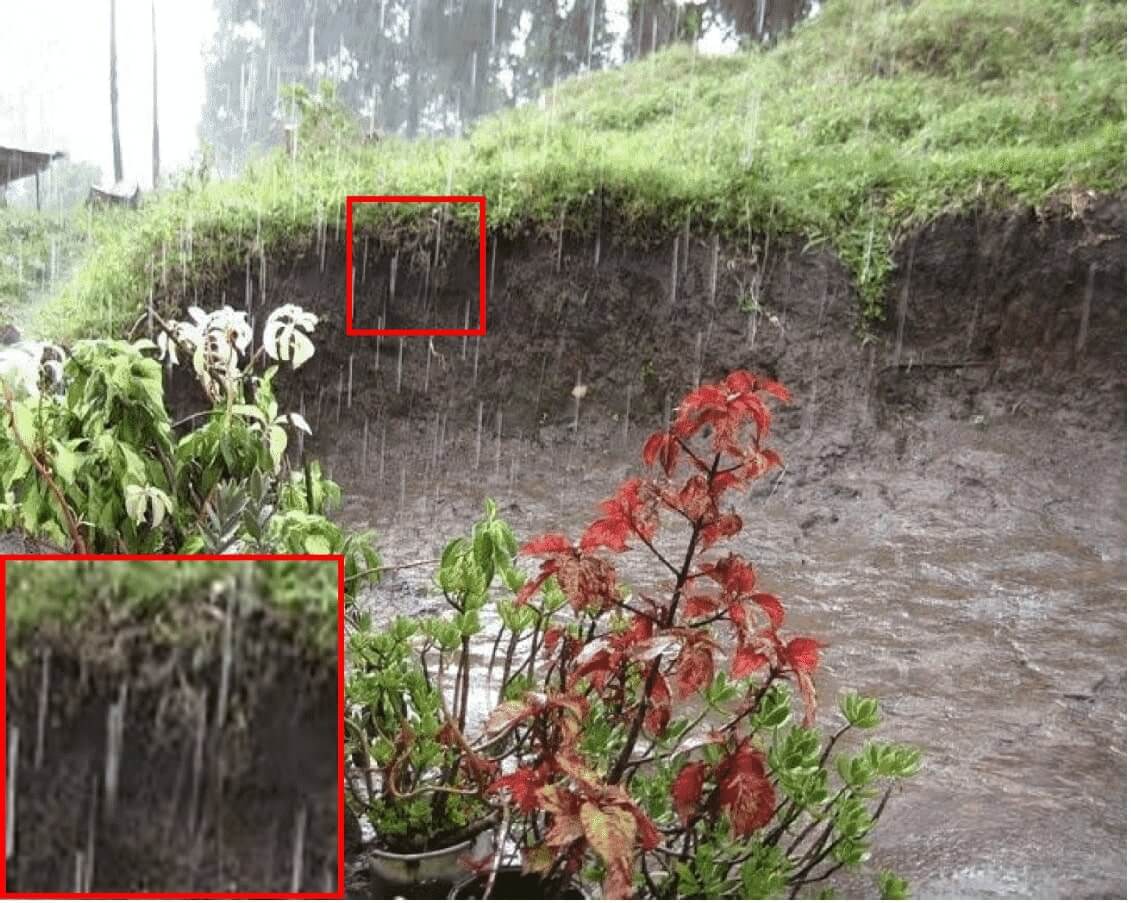}}
    \hfill
    \subfloat[\centering HRR~\cite{li2019heavy}]{\includegraphics[width=\Qwidth]{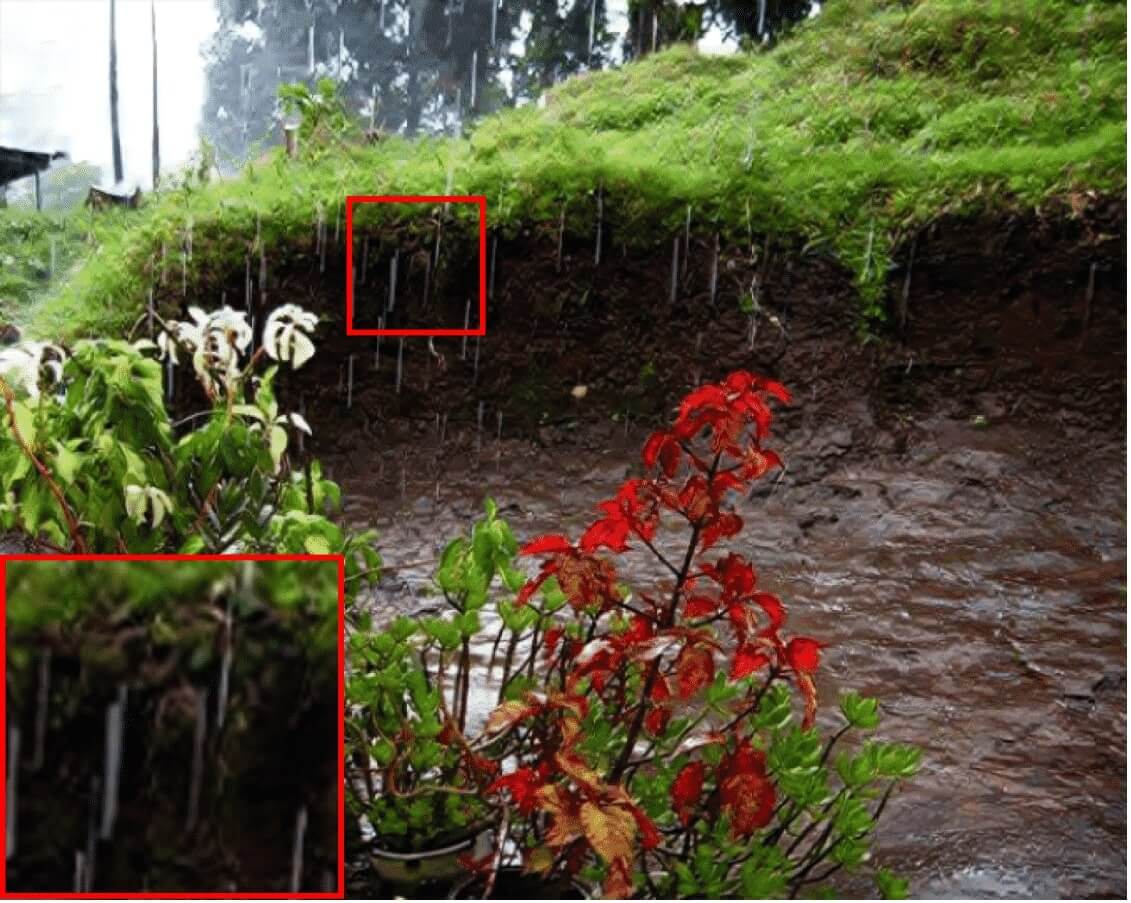}}
    \hfill
    \subfloat[\centering MSPFN~\cite{jiang2020multi}]{\includegraphics[width=\Qwidth]{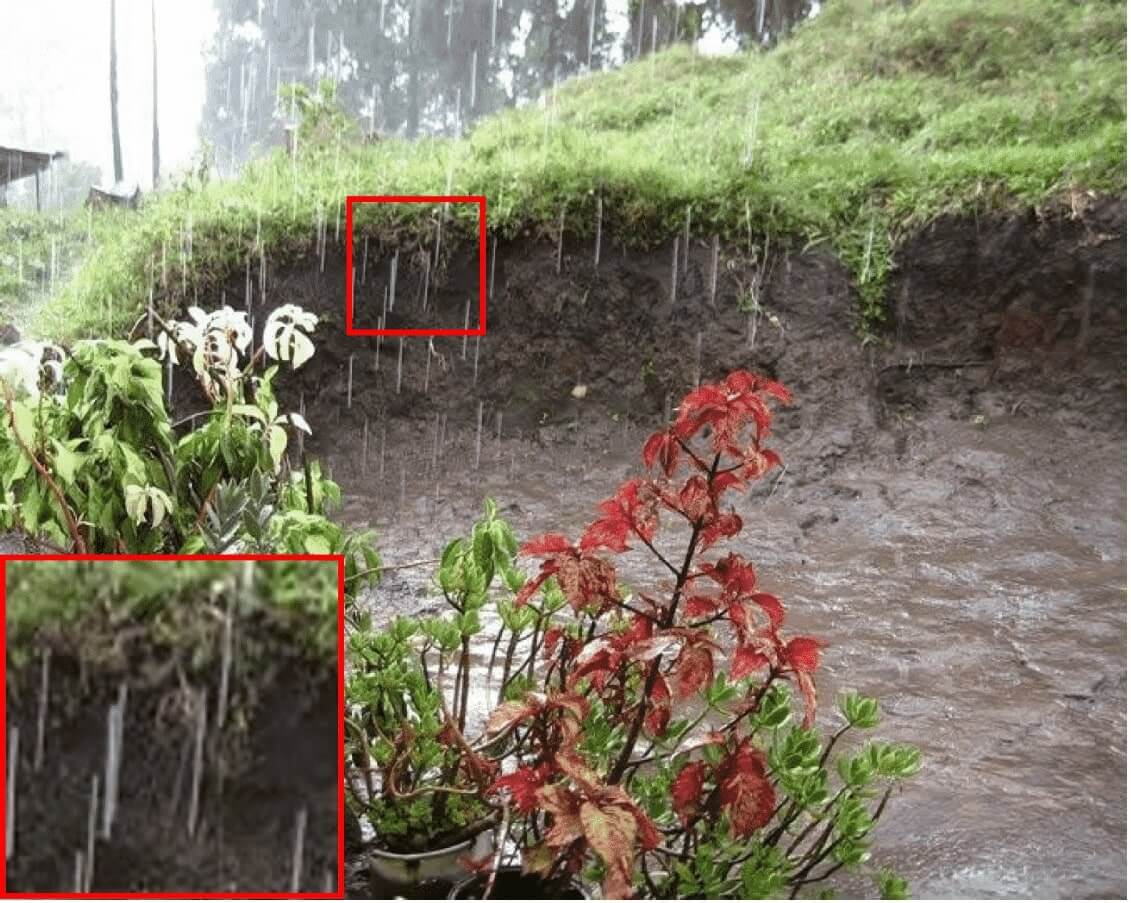}}
    \hfill
    \subfloat[\centering RCDNet~\cite{wang2020a}]{\includegraphics[width=\Qwidth]{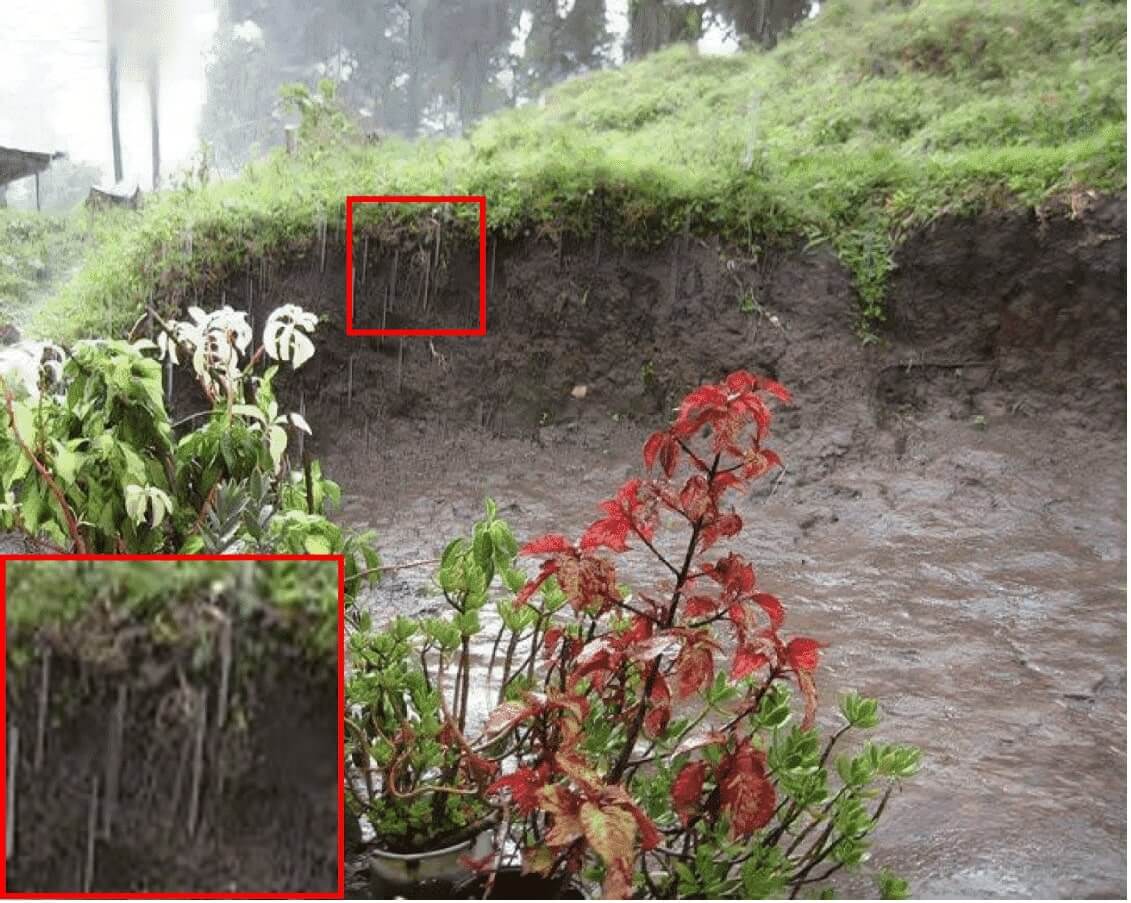}}
    
    \par
    
    
    \subfloat[\centering DGNL-Net~\cite{hu2021single}]{\includegraphics[width=\Qwidth]{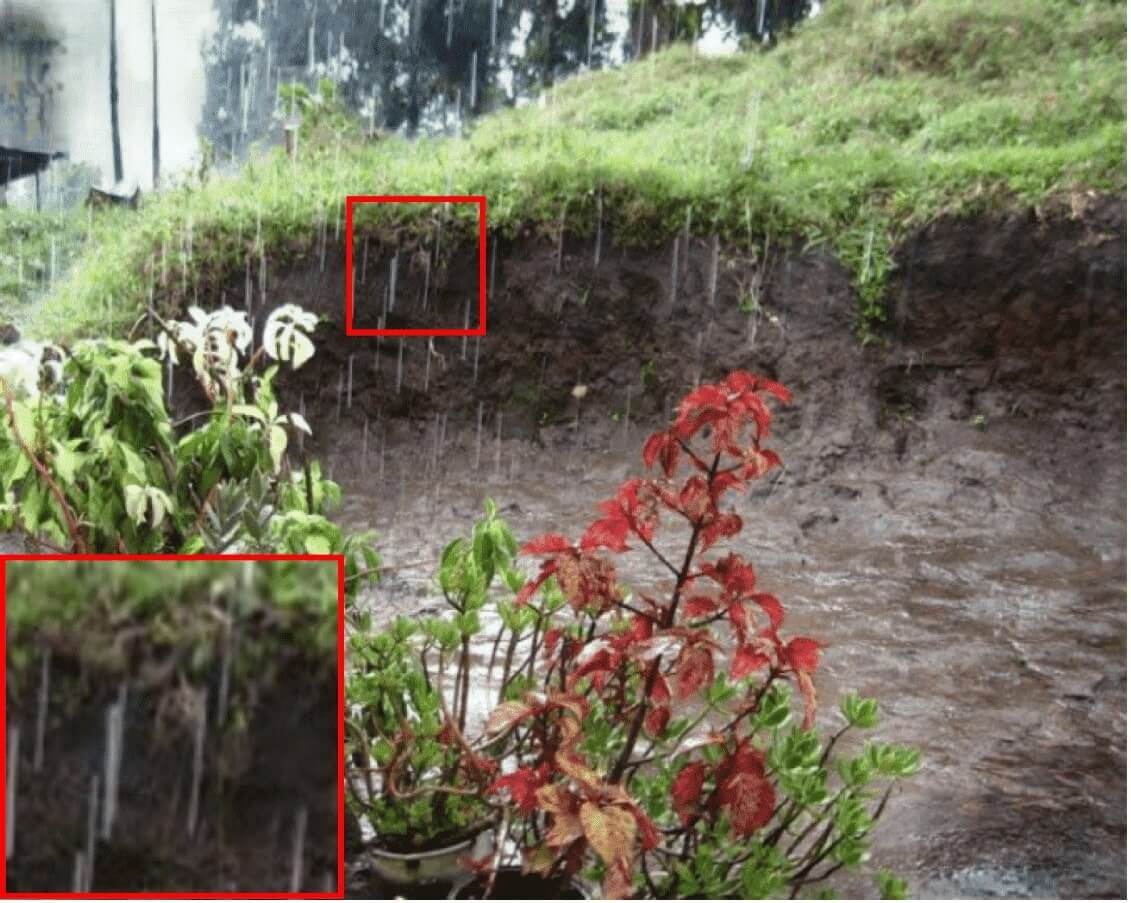}}
    \hfill
    \subfloat[\centering EDR V4 (S)~\cite{guo2021efficientderain}]{\includegraphics[width=\Qwidth]{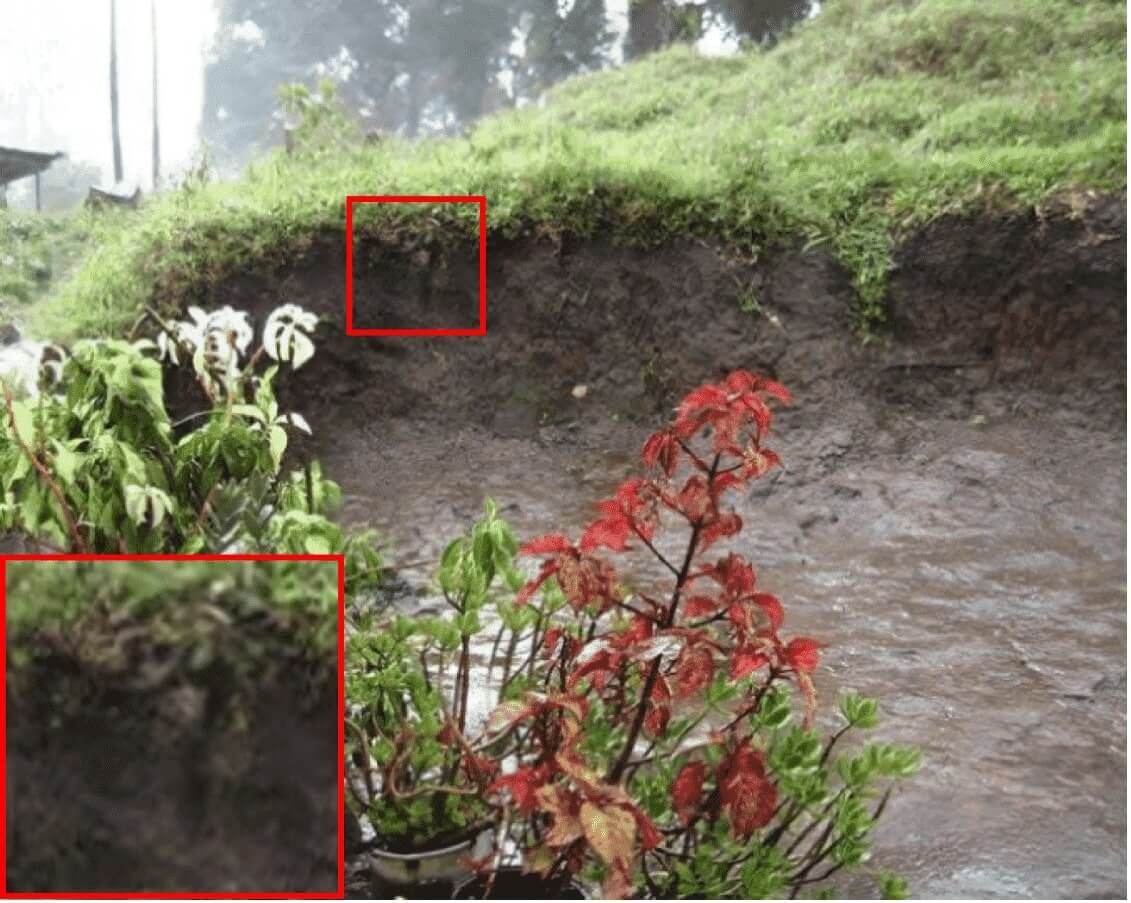}}
    \hfill
    \subfloat[\centering EDR V4 (R)~\cite{guo2021efficientderain}]{\includegraphics[width=\Qwidth]{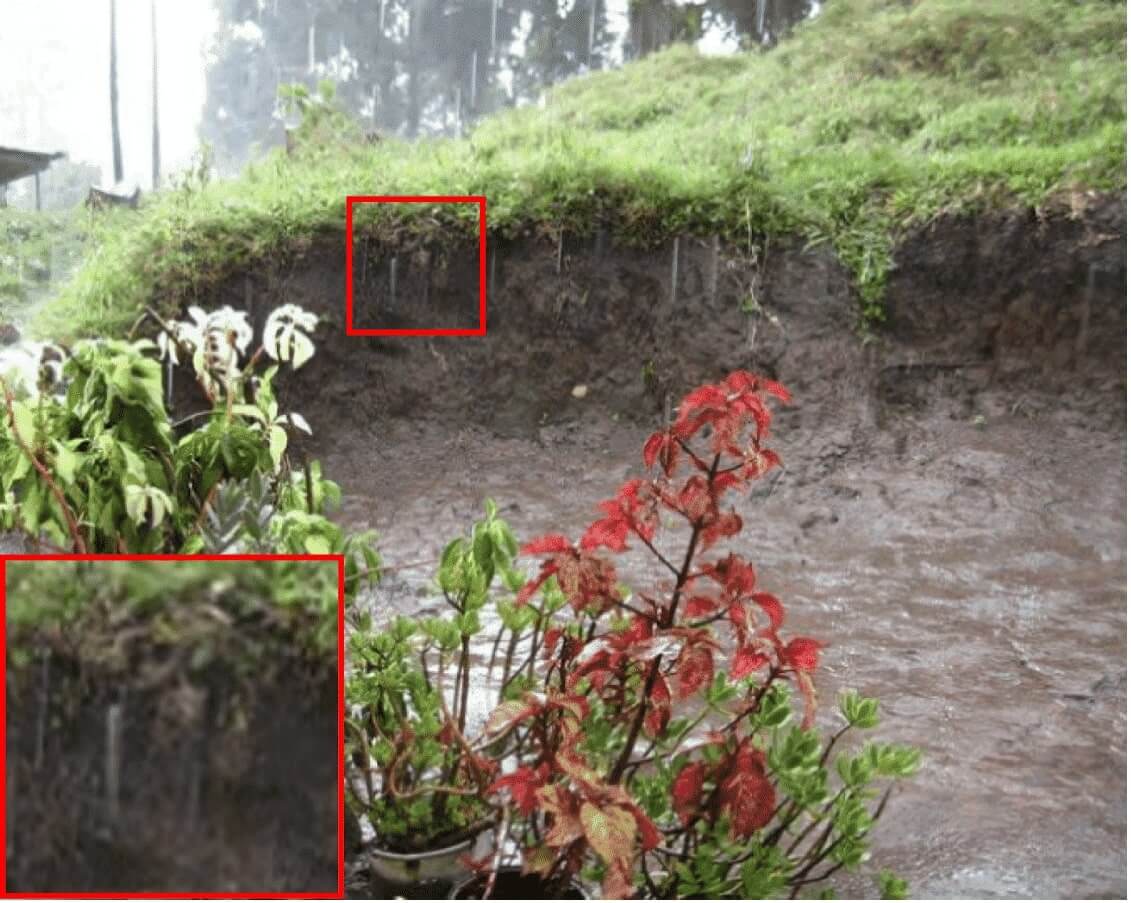}}
    \hfill
    \subfloat[\centering MPRNet~\cite{zamir2021multi}]{\includegraphics[width=\Qwidth]{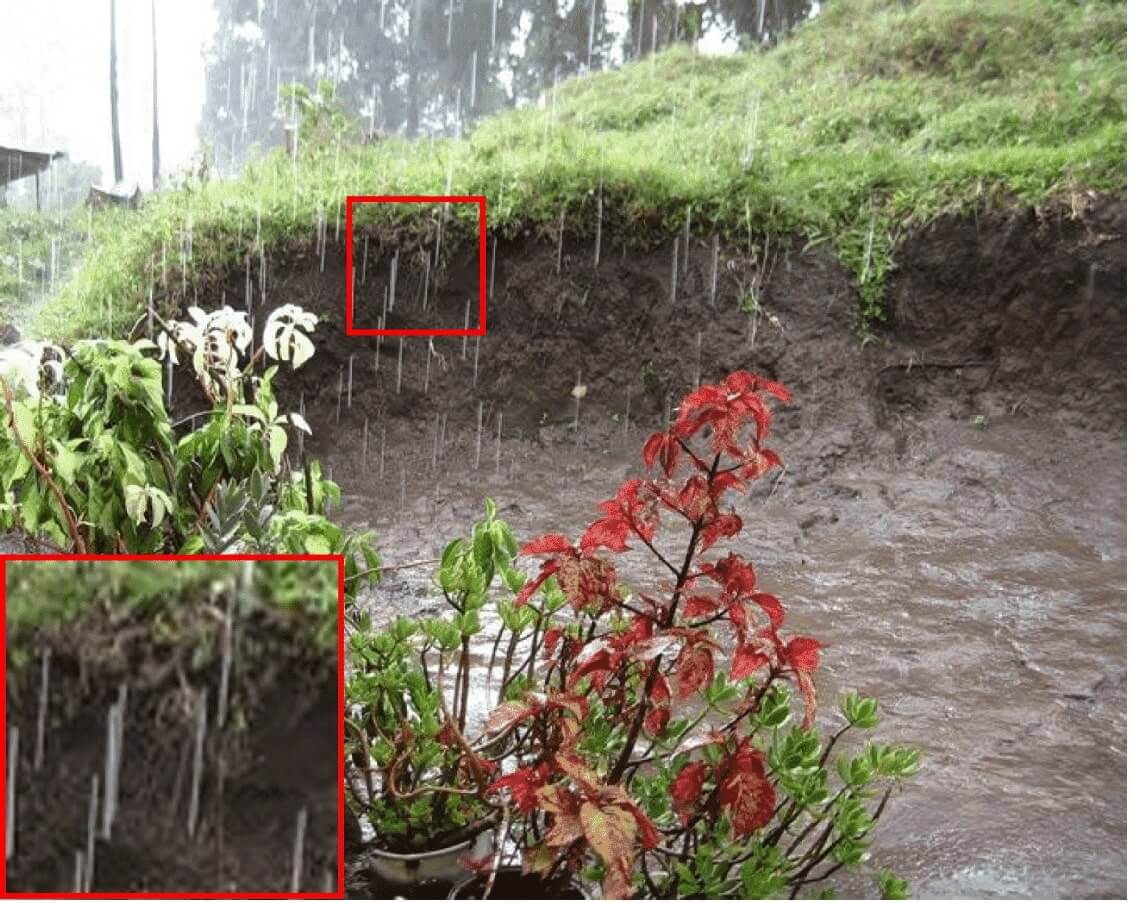}}
    \hfill
    \subfloat[\centering Ours]{\includegraphics[width=\Qwidth]{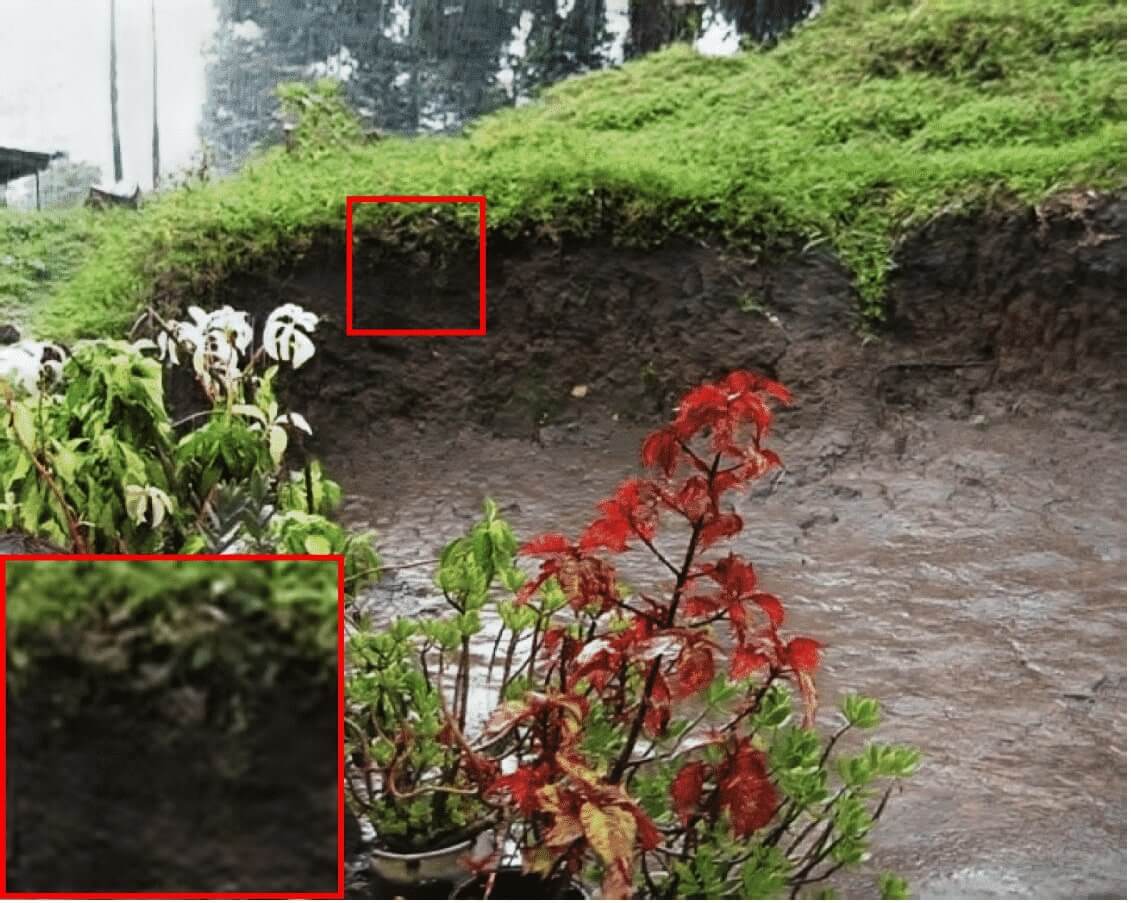}}
	\caption{\textbf{Our model can generalize across real rainy images with robust performance.} We select representative real rainy images with various rain patterns and backgrounds for comparison (zoom for details). EDR V4 (S)~\cite{guo2021efficientderain} denotes EDR trained on SPA-Data~\cite{wang2019spatial}, and EDR V4 (R)~\cite{guo2021efficientderain} denotes EDR trained on Rain14000~\cite{fu2017removing}.}
   \label{fig:other_real_results}
\end{figure*}

\noindent\textbf{Qualitative evaluation on other real images:}
Other than the models described in the above section, we also include EDR V4~\cite{guo2021efficientderain} trained on SPA-Data~\cite{wang2019spatial} for the qualitative comparison, since it shows more robust rain streak removal results as compared the version trained on Rain14000~\cite{fu2017removing}. The derained results on Internet rainy images are illustrated in~\cref{fig:other_real_results}. The model trained on the proposed \dname\ (i.e. ours) deals with large rain streaks of various shapes and sizes as well as the associated rain accumulation effects, while preserving the features present in the scene. In contrast, we observe that models~\cite{hu2021single,li2019heavy} trained on data with synthetic rain accumulation introduce unwanted color shifts and residual rain streaks in their results. Moreover, the state-of-the-art methods~\cite{jiang2020multi,wang2020a,zamir2021multi} are unable to remove the majority of rain streaks in general as highlighted in the red zoom boxes. This demonstrates the gap between top methods on synthetic versus one that can be applied to real data. \\

\begin{table}[t]
    \caption{\textbf{Retraining comparison methods on \dname.} The improvement of these derainers further demonstrates the effectiveness of real paired data.}
  \label{tab:retrain_results} 
  \centering
  \scriptsize
  \resizebox{\columnwidth}{!}{
  \begin{tabular}{ccccccccccccc}
    \toprule
    \makecell{Data Split} & Metrics & \makecell{Rainy \\ Images} & \makecell{RCDNet~\cite{wang2020a} \\ (Original)} & \makecell{RCDNet~\cite{wang2020a} \\ (\dname)} & 
    \makecell{EDR~\cite{guo2021efficientderain} \\ (Original)} & \makecell{EDR~\cite{guo2021efficientderain} \\ (\dname)} & \makecell{MPRNet~\cite{zamir2021multi} \\ (Original)} & 
    \makecell{MPRNet~\cite{zamir2021multi} \\ (\dname)} & 
    \makecell{Ours} \\
    \midrule
    \makecell{Dense Rain \\ Streaks} & \makecell{PSNR$\uparrow$ \\ SSIM$\uparrow$} & \makecell{18.46 \\ 0.6284} & \makecell{19.50 \\ 0.6218} & \makecell{19.60 \\ 0.6492} & \makecell{18.86 \\ 0.6296} & \makecell{19.95 \\ 0.6436} & \makecell{19.12 \\ 0.6375} & \makecell{20.19 \\ 0.6542} & \makecell{\textbf{20.84} \\ \textbf{0.6573}} \\
    \midrule
    \makecell{Dense Rain \\ Accumulation} & \makecell{PSNR$\uparrow$ \\ SSIM$\uparrow$} & \makecell{20.87 \\ 0.7706} & \makecell{21.27 \\ 0.7765} & \makecell{22.74 \\ 0.7891} & \makecell{21.07 \\ 0.7766} & \makecell{23.42 \\ 0.7994} & \makecell{21.38 \\ 0.7808} & \makecell{23.38 \\ 0.8009} & \makecell{\textbf{24.78} \\ \textbf{0.8279}} \\
    \midrule
    \makecell{Overall} & \makecell{PSNR$\uparrow$ \\ SSIM$\uparrow$} & \makecell{19.49 \\ 0.6893} & \makecell{20.26 \\ 0.6881} & \makecell{20.94 \\ 0.7091} & \makecell{19.81 \\ 0.6926} & \makecell{21.44 \\ 0.7104} & \makecell{20.09 \\ 0.6989} & \makecell{21.56 \\ 0.7171} & \makecell{\textbf{22.53} \\ \textbf{0.7304}} \\
    \bottomrule
  \end{tabular}}
\end{table}

\noindent\textbf{Retraining other methods on \dname:}
We additionally train several state-of-the-art derainers~\cite{guo2021efficientderain,wang2020a,zamir2021multi} on the \dname\ training set to demonstrate that our real dataset leads to more robust real-world deraining and benefits all models. We have selected the most recent derainers for this retraining study.\footnote{Both DGNL-Net~\cite{hu2021single} and HRR~\cite{li2019heavy} cannot be retrained on our real dataset, as both require additional supervision, such as transmission maps and depth maps.} All the models are trained from scratch, and the corresponding PSNR and SSIM scores on the \dname\ test set are provided in~\cref{tab:retrain_results}. For all the retrained models, we can observe a PSNR and SSIM gain by using the proposed \dname\ dataset. In addition, with all models trained on the same dataset, our model still outperforms others in all categories. \\

\noindent\textbf{Fine-tuning other methods on \dname:}
To demonstrate of the effectiveness of combining real and synthetic datasets, we also fine-tune several more recent derainers~\cite{guo2021efficientderain,wang2020a,zamir2021multi} that are previously trained on synthetic datasets with the proposed \dname\ dataset. We fine-tune from the official weights as described in the above quantitative evaluation section, and the fine-tuning learning rate is 20\% of the original learning rate for each method. For the proposed method, we pretrain the model on the synthetic dataset used by MSPFN~\cite{jiang2020multi} and MPRNet~\cite{zamir2021multi}. The corresponding PSNR and SSIM scores on the \dname\ test set are listed in~\cref{tab:finetune_results}. In the table, we can observe a further boost as compared with training the models from scratch with just real or synthetic data.

\begin{table}[t]
    \caption{\textbf{Fine-tuning comparison methods on \dname.} (F) denotes the fine-tuned models, and (O) denotes the original models trained on synthetic/real data.}
  \label{tab:finetune_results} 
  \centering
  \scriptsize
  \resizebox{\columnwidth}{!}{
  \begin{tabular}{cccccccccccccc}
    \toprule
    \makecell{Data Split} & Metrics & \makecell{Rainy \\ Images} & \makecell{RCDNet~\cite{wang2020a} \\ (O)} & \makecell{RCDNet~\cite{wang2020a} \\ (F)} & 
    \makecell{EDR~\cite{guo2021efficientderain} \\ (O)} & \makecell{EDR~\cite{guo2021efficientderain} \\ (F)} & \makecell{MPRNet~\cite{zamir2021multi} \\ (O)} & 
    \makecell{MPRNet~\cite{zamir2021multi} \\ (F)} & 
    \makecell{Ours \\ (O)} &
    \makecell{Ours \\ (F)} \\
    \midrule
    \makecell{Dense Rain \\ Streaks} & \makecell{PSNR$\uparrow$ \\ SSIM$\uparrow$} & \makecell{18.46 \\ 0.6284} & \makecell{19.50 \\ 0.6218} & \makecell{19.33 \\ 0.6463} & \makecell{18.86 \\ 0.6296} & \makecell{20.03 \\ 0.6433} & \makecell{19.12 \\ 0.6375} & \makecell{20.65 \\ 0.6561} & \makecell{\textbf{20.84} \\ 0.6573} & \makecell{20.79 \\ \textbf{0.6655}}\\
    \midrule
    \makecell{Dense Rain \\ Accumulation} & \makecell{PSNR$\uparrow$ \\ SSIM$\uparrow$} & \makecell{20.87 \\ 0.7706} & \makecell{21.27 \\ 0.7765} & \makecell{22.50 \\ 0.7893} & \makecell{21.07 \\ 0.7766} & \makecell{23.57 \\ 0.8016} & \makecell{21.38 \\ 0.7808} & \makecell{24.37 \\ 0.8250} & \makecell{24.78 \\ 0.8279} & \makecell{\textbf{25.20} \\ \textbf{0.8318}}\\
    \midrule
    \makecell{Overall} & \makecell{PSNR$\uparrow$ \\ SSIM$\uparrow$} & \makecell{19.49 \\ 0.6893} & \makecell{20.26 \\ 0.6881} & \makecell{20.69 \\ 0.7076} & \makecell{19.81 \\ 0.6926} & \makecell{21.55 \\ 0.7111} & \makecell{20.09 \\ 0.6989} & \makecell{22.24 \\ 0.7285} & \makecell{22.53 \\ 0.7304} & \makecell{\textbf{22.68} \\ \textbf{0.7368}} \\
    \bottomrule
  \end{tabular}}
\end{table}

\begin{table}
  \caption{\textbf{Ablation study.} Our rain-robust loss improves both PSNR and SSIM.}
  \setlength\tabcolsep{18.5pt}
  \scriptsize
  \label{tab:ablation}
  \centering
  \begin{tabular}{cccc}
    \toprule
    Metrics & \makecell{Rainy Images} & \makecell{Ours w/o $\mathcal{L}_{\text{robust}}$} & \makecell{Ours w/ $\mathcal{L}_{\text{robust}}$} \\
    \midrule
    PSNR$\uparrow$ & 19.49 & 21.82 & \textbf{22.53}\\
    SSIM$\uparrow$ & 0.6893 & 0.7148 & \textbf{0.7304}\\
    \bottomrule
  \end{tabular}
\end{table}

\noindent\textbf{Ablation study:}
We validate the effectiveness of the rain-robust loss with two variants of the proposed method: (1) the proposed network with the full objective as describe in~\cref{sec:proposed_method}; and (2) the proposed network with just MS-SSIM loss and $\ell_1$ loss. The rest of the training configurations and hyperparameters remain identical. The quantitative metrics for these two variants on the proposed \dname\ test set are listed in~\cref{tab:ablation}. Our model trained with the proposed rain-robust loss produces a normalized correlation between rainy and clean latent vectors of .95 $\pm$ .03; whereas it is .85 $\pm$ .10 for the one without. These rain-robust features help the model to show improved performance in both PSNR and SSIM. \\

\begin{figure}[t]
    \captionsetup[subfloat]{font=scriptsize,farskip=3pt,captionskip=2pt}
    \newcommand{\failurewidth}{0.195\textwidth}
    \centering
    \subfloat[\centering Rainy]{\includegraphics[width=\failurewidth]{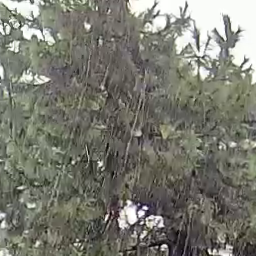}}  \hfill 
    \subfloat[\centering EDR V4 (R)~\cite{guo2021efficientderain}]{\includegraphics[width=\failurewidth]{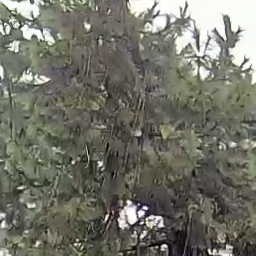}}  \hfill 
    \subfloat[\centering MPRNet~\cite{zamir2021multi}]{\includegraphics[width=\failurewidth]{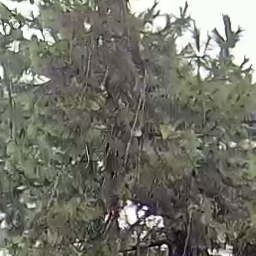}}  \hfill 
    \subfloat[\centering Ours]{\includegraphics[width=\failurewidth]{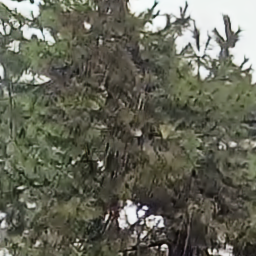}}  \hfill 
    \subfloat[\centering Ground Truth]{\includegraphics[width=\failurewidth]{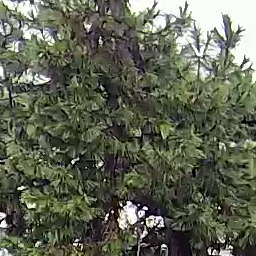}}
    \caption{\textbf{Deraining is still an open problem.} Both the proposed method and the existing work have difficulty in generalizing the performance to some challenging scenes.}
    \label{fig:failure_cases}
\end{figure}

\noindent\textbf{Failure cases:} Apart from the successful cases illustrated in~\cref{fig:our_real_results}, we also provide some of the failure cases in the \dname\ test set in~\cref{fig:failure_cases}. Deraining is still an open problem, and we hope future work can take advantages of both real and synthetic samples to make derainers more robust in diverse environments. 

\section{Conclusions} \label{sec:conclusion}
Many of us in the deraining community probably wish for the existence of parallel universes, where we could capture the exact same scene with and without weather effects at the exact same time. Unfortunately, however, we are stuck with our singular universe, in which we are left with two choices: (1) synthetic data at the same timestamp with simulated weather effects or (2) real data at different timestamps with real weather effects. Though it is up to opinion, it is our belief that the results of our method in~\cref{fig:other_real_results} reduce the visual domain gap more than those trained with synthetic datasets. Additionally, we hope the introduction of a real dataset opens up exciting new pathways for future work, such as the blending of synthetic and real data or setting goalposts to guide the continued development of existing rain simulators~\cite{halder2019physics,ni2021controlling,wang2021rain,ye2021closing,yue2021semi}. \\

\noindent\textbf{Acknowledgements:} The authors thank members of the Visual Machines Group for their feedback and support, as well as Mani Srivastava and Cho-Jui Hsieh for technical discussions. This research was partially supported by ARL W911NF-20-2-0158 under the cooperative A2I2 program. A.K. was also partially supported by an Army Young Investigator Award.

\clearpage
%
%
\bibliographystyle{splncs04}
\bibliography{egbib}
\end{document}


\pagestyle{headings}
\mainmatter
\def\ECCVSubNumber{XXXX}  

\title{Not Just Streaks: Towards Ground Truth for Single Image Deraining \\ (Supplementary Material)} 

\titlerunning{Not Just Streaks: Towards Ground Truth for Single Image Deraining}
%

\author{Yunhao Ba\inst{1}$^{\star}$ \and
Howard Zhang\inst{1}\thanks{Equal contribution.} \and
Ethan Yang\inst{1} \and
Akira Suzuki\inst{1} \and
Arnold Pfahnl\inst{1} \and
Chethan Chinder Chandrappa\inst{1}\index{Chandrappa, Chethan Chinder} \and
Celso M. de Melo\inst{2}\index{de Melo, Celso M.} \and
Suya You\inst{2} \and
Stefano Soatto\inst{1} \and
Alex Wong\inst{3} \and
Achuta Kadambi\inst{1}}

\authorrunning{Y. Ba et al.}
%
\institute{University of California, Los Angeles
\email{\{yhba,hwdz15508,eyang657,asuzuki100,ajpfahnl,chinderc\}@ucla.edu}\\
\email{soatto@cs.ucla.edu}, \email{achuta@ee.ucla.edu}\\ \and 
DEVCOM Army Research Laboratory
\email{\{celso.m.demelo.civ,suya.you.civ\}@army.mil} \\ \and
Yale University \\
\email{alex.wong@yale.edu}}

\maketitle

\section{Visualization of Previous Deraining Datasets}  \label{sec:data_visualization}
\begin{figure}[h]
\centering
    \captionsetup[subfloat]{farskip=3pt,captionskip=2pt}
    \newcommand{\Awidth}{0.32\textwidth}
    \subfloat[]{\includegraphics[width=0.038\textwidth]{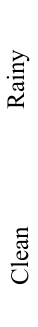}}
    \subfloat[\centering Synthetic Data~\cite{fu2017removing}]{\includegraphics[width=\Awidth]{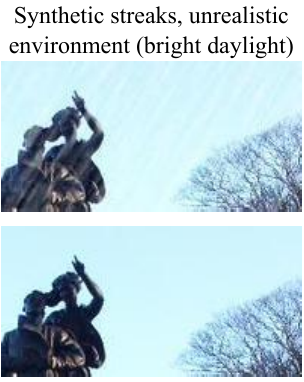}}
    \hfill
    \subfloat[\centering SPA-Data~\cite{wang2019spatial}]{\includegraphics[width=\Awidth]{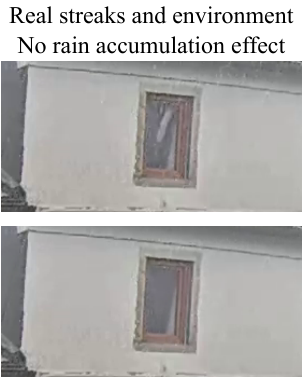}}
    \hfill
    \subfloat[\centering \dname]{\includegraphics[width=\Awidth]{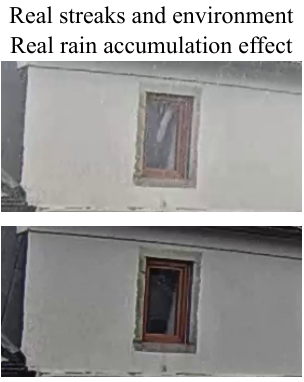}}
    \caption{\textbf{\dname\ contains realistic rain effects (both rain streaks and rain accumulation), while the existing synthetic and semi-real datasets fail to cover the physical complexity and diversity of real-world rain.} The synthetic image pair is from the commonly used Rain14000 dataset~\cite{fu2017removing}, and the pseudo ground-truth image of SPA-Data~\cite{wang2019spatial} in the figure is generated by running the official code from the authors on our collected rainy video.}
    \label{fig:collection_pipeline}
\end{figure}

We illustrate some typical image pairs from various deraining datasets in~\cref{fig:collection_pipeline}. Synthetic datasets in the community are usually generated by adding synthetic rain effects on real images taken under sunny illumination conditions, and the semi-real SPA-Data~\cite{wang2019spatial} only considers rain streaks. As a result, the domain gap between these existing datasets and real rainy scenarios are relatively larger as compared with the proposed \dname\ dataset.


\section{More Results from \dname}  \label{sec:more_images_on_our_dataset}

As an additional supplement to~\fignohref{5} in the main paper, we provide some more quantitative and qualitative results from our test set in~\cref{fig:other_quant_results}. Note that these comparison models are using the weights provided by the authors which are trained on synthetic or semi-real datasets. We see that our proposed model trained on \dname\ continues to outperform other competing models.

 
\newcommand{\figwidth}{.195}
\newcommand{\figheight}{.06cm}
\newcommand{\Qwidth}{\figwidth\textwidth}
\begin{figure*}[!ht]
    \captionsetup[subfloat]{font=scriptsize,farskip=3pt,captionskip=2pt}
    \centering
  
    \subfloat[\centering Rain (17.10/0.5913)]{\includegraphics[width=\Qwidth]{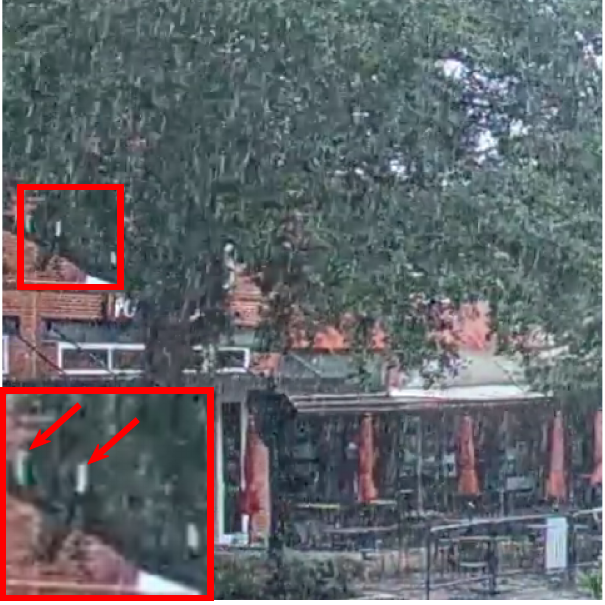}}
    \hfill
    \subfloat[\centering SPANet~\cite{wang2019spatial} (17.39/0.5964)]{\includegraphics[width=\Qwidth]{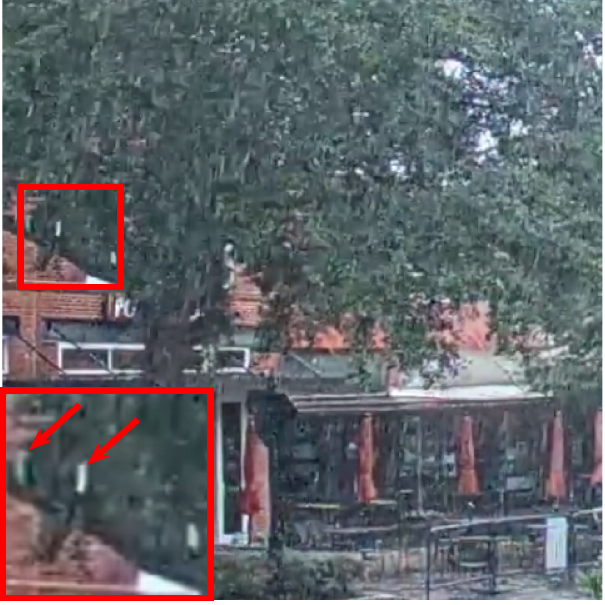}}
    \hfill    
    \subfloat[\centering HRR~\cite{li2019heavy} (19.42/0.6219)]{\includegraphics[width=\Qwidth]{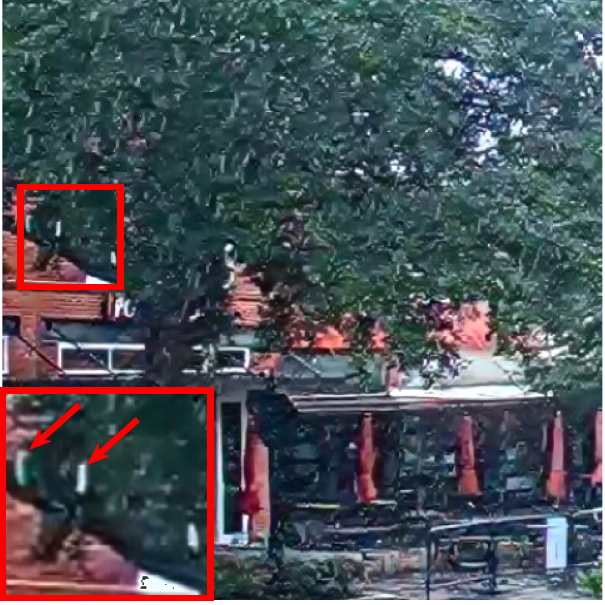}}
    \hfill
    \subfloat[\centering MSPFN~\cite{jiang2020multi} (18.04/0.5907)]{\includegraphics[width=\Qwidth]{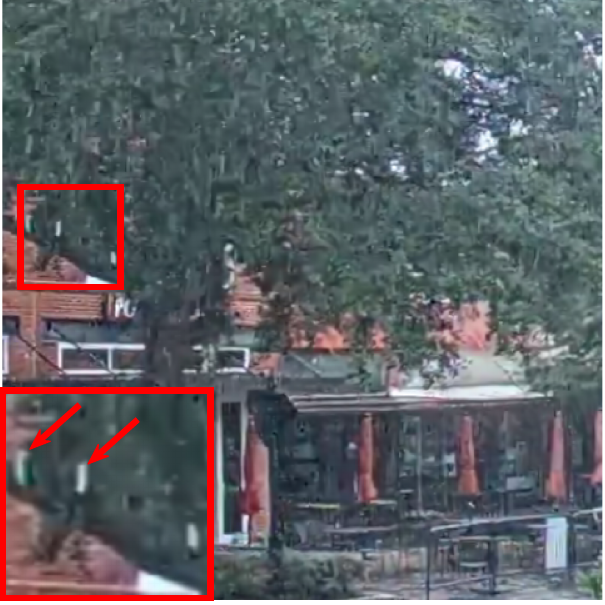}}
    \hfill
    \subfloat[\centering RCDNet~\cite{wang2020a} (17.40/0.5941)]{\includegraphics[width=\Qwidth]{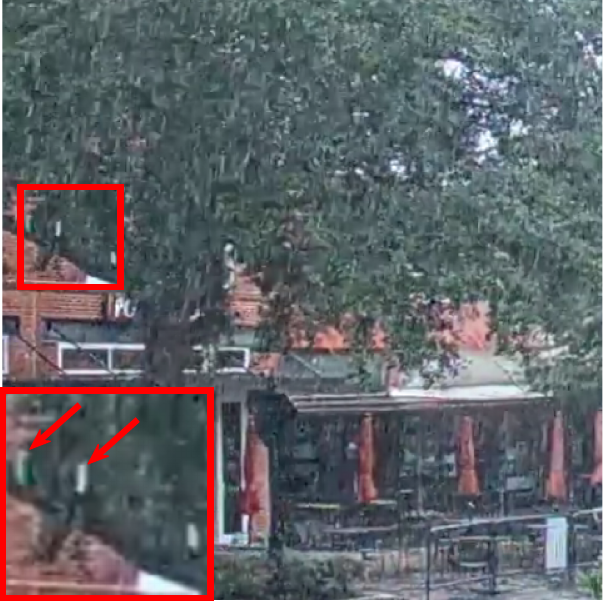}}
  
    \par\vspace{\figheight}

    \subfloat[\centering DGNL-Net~\cite{hu2021single} (19.59/0.6116)]{\includegraphics[width=\Qwidth]{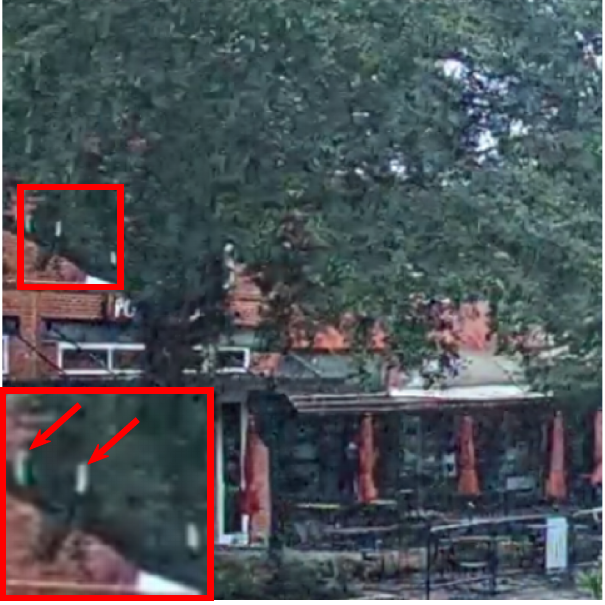}}
    \hfill
    \subfloat[\centering Efficient Derain~\cite{guo2021efficientderain} (17.29/0.5916)]{\includegraphics[width=\Qwidth]{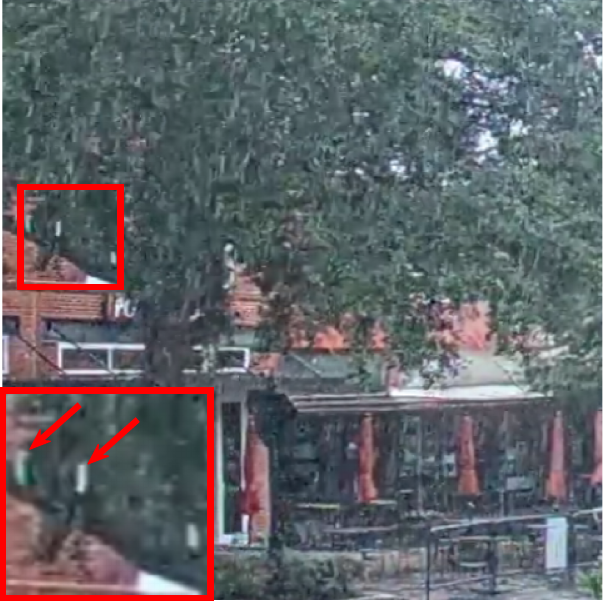}}
    \hfill    
    \subfloat[\centering MPRNet~\cite{zamir2021multi} (17.90/0.5995)]{\includegraphics[width=\Qwidth]{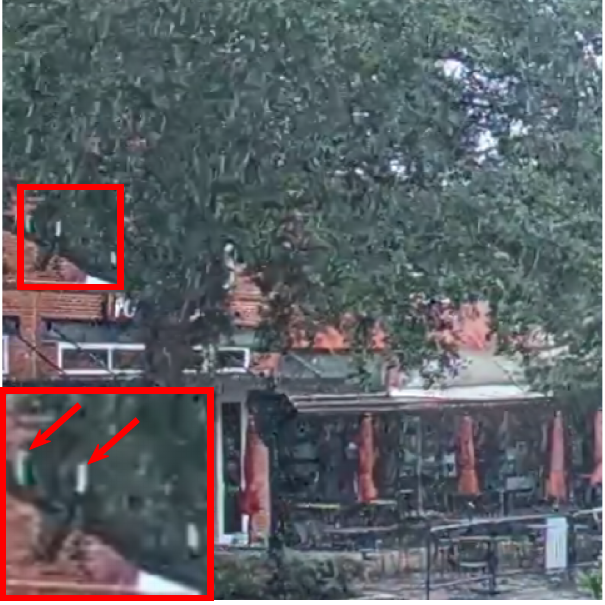}}
    \hfill
    \subfloat[\centering \textbf{Ours (19.84/0.6405)}]{\includegraphics[width=\Qwidth]{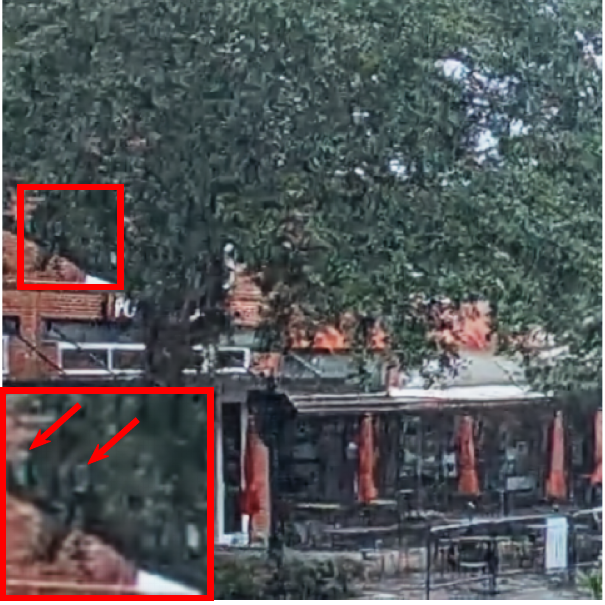}}
    \hfill
    \subfloat[\centering Ground Truth (PSNR/SSIM)]{\includegraphics[width=\Qwidth]{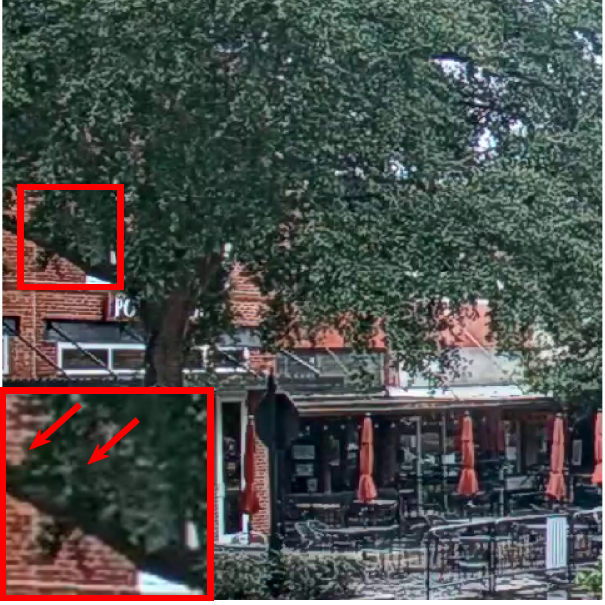}}

  \caption{\textbf{More results on \dname\ test set.} Similarly, the proposed method is capable of removing various rain streaks and rain accumulation effects.}
    \label{fig:other_quant_results}  
  \end{figure*}

\section{More Results on Internet Images} \label{sec:more_qualitative_results}
\begin{figure*}[!ht]
    \newcommand{\Qualwidth}{0.195\textwidth}
    \captionsetup[subfloat]{font=scriptsize,farskip=3pt,captionskip=2pt}
    \centering
  
        


    \subfloat[\centering Rainy Image]{\includegraphics[width=\Qualwidth]{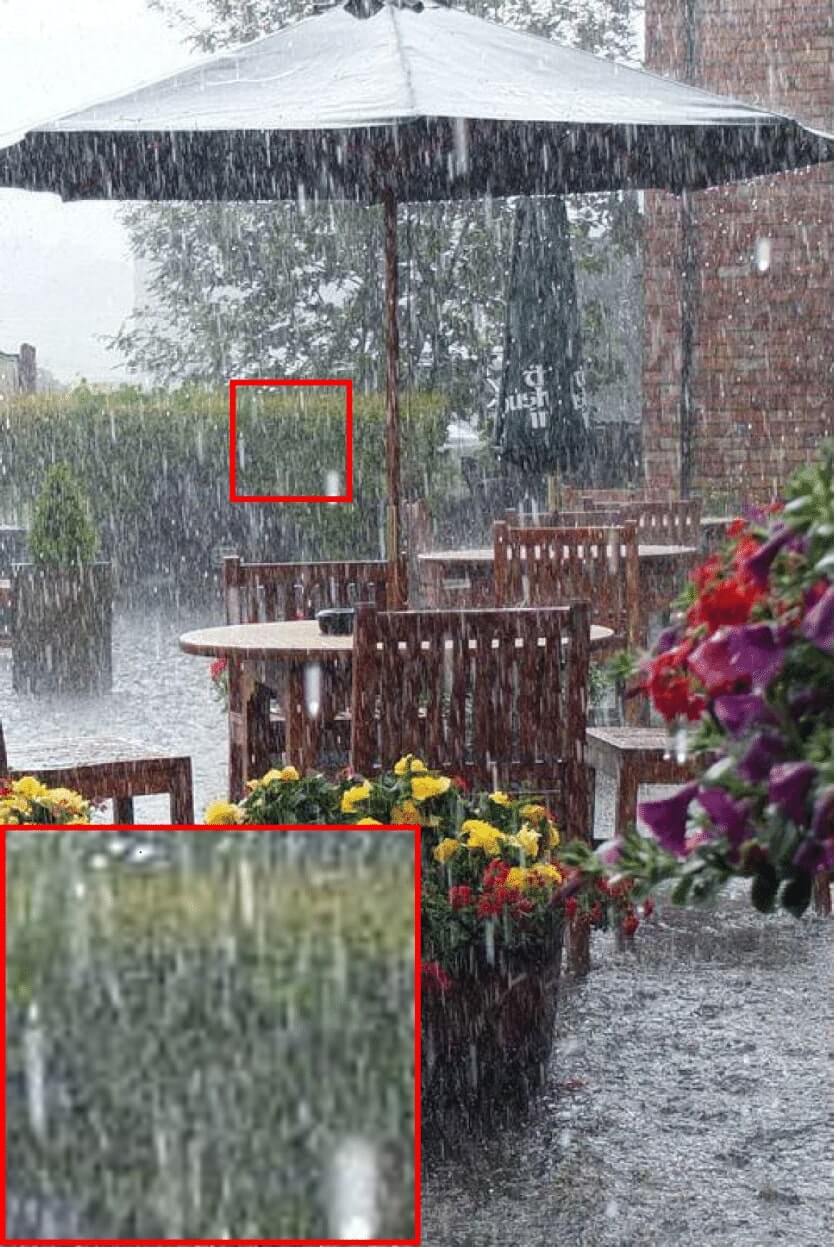}}
    \hfill
    \subfloat[\centering SPANet~\cite{wang2019spatial}]{\includegraphics[width=\Qualwidth]{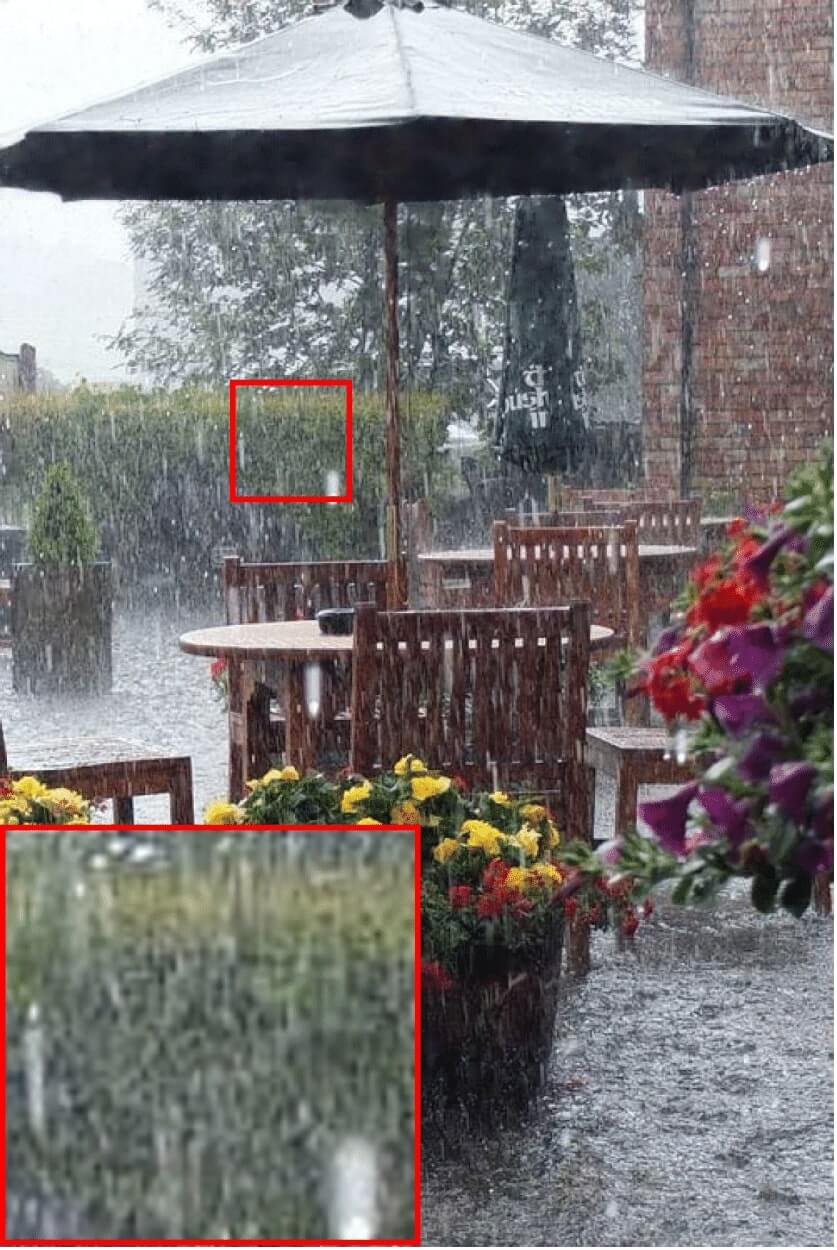}}
    \hfill
    \subfloat[\centering HRR~\cite{li2019heavy}]{\includegraphics[width=\Qualwidth]{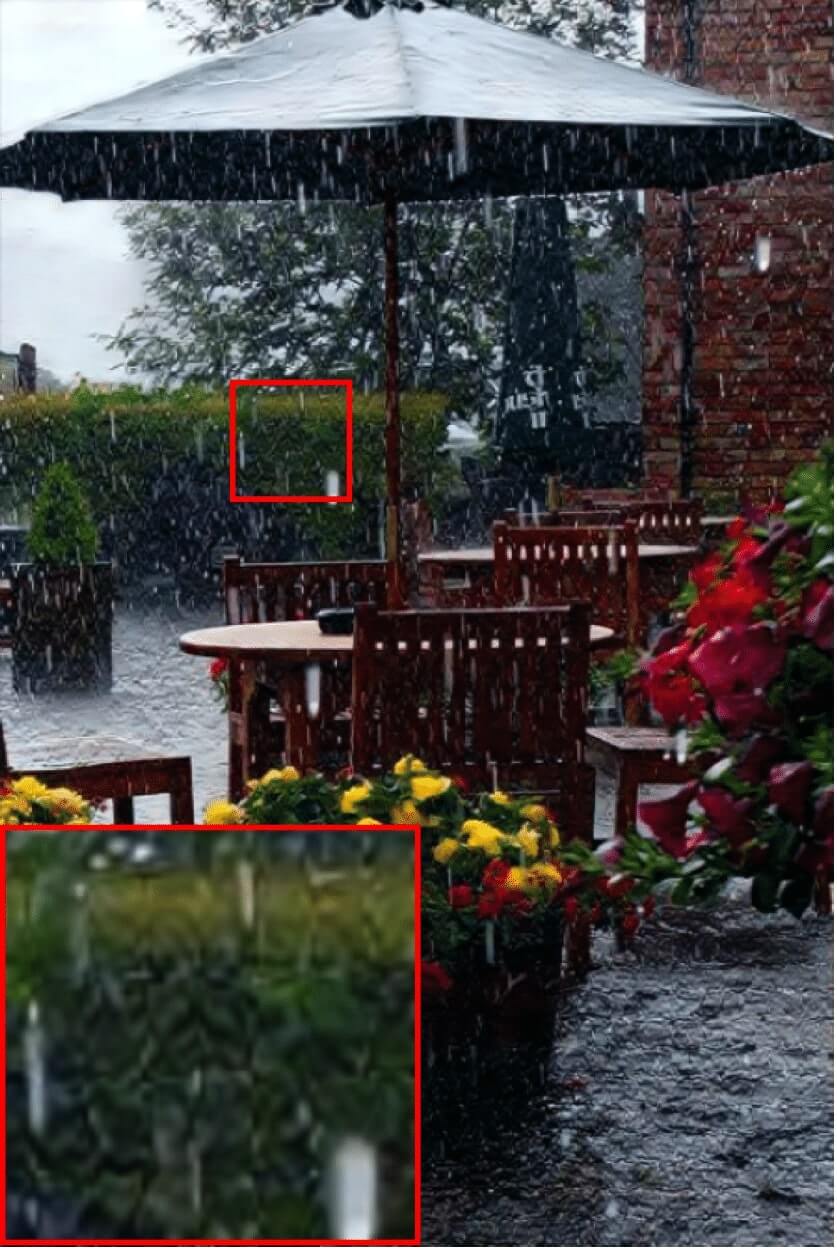}}
    \hfill
    \subfloat[\centering MSPFN~\cite{jiang2020multi} ]{\includegraphics[width=\Qualwidth]{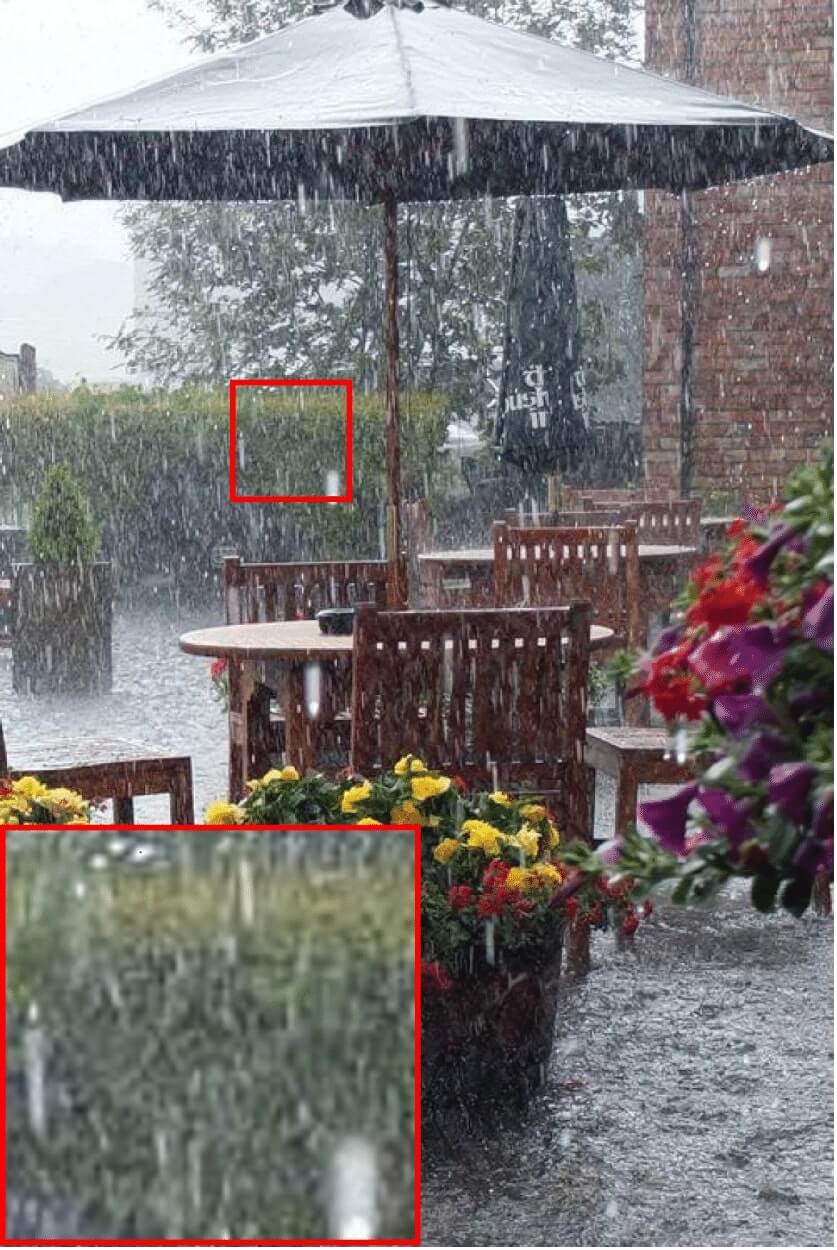}}
    \hfill
    \subfloat[\centering RCDNet~\cite{wang2020a}]{\includegraphics[width=\Qualwidth]{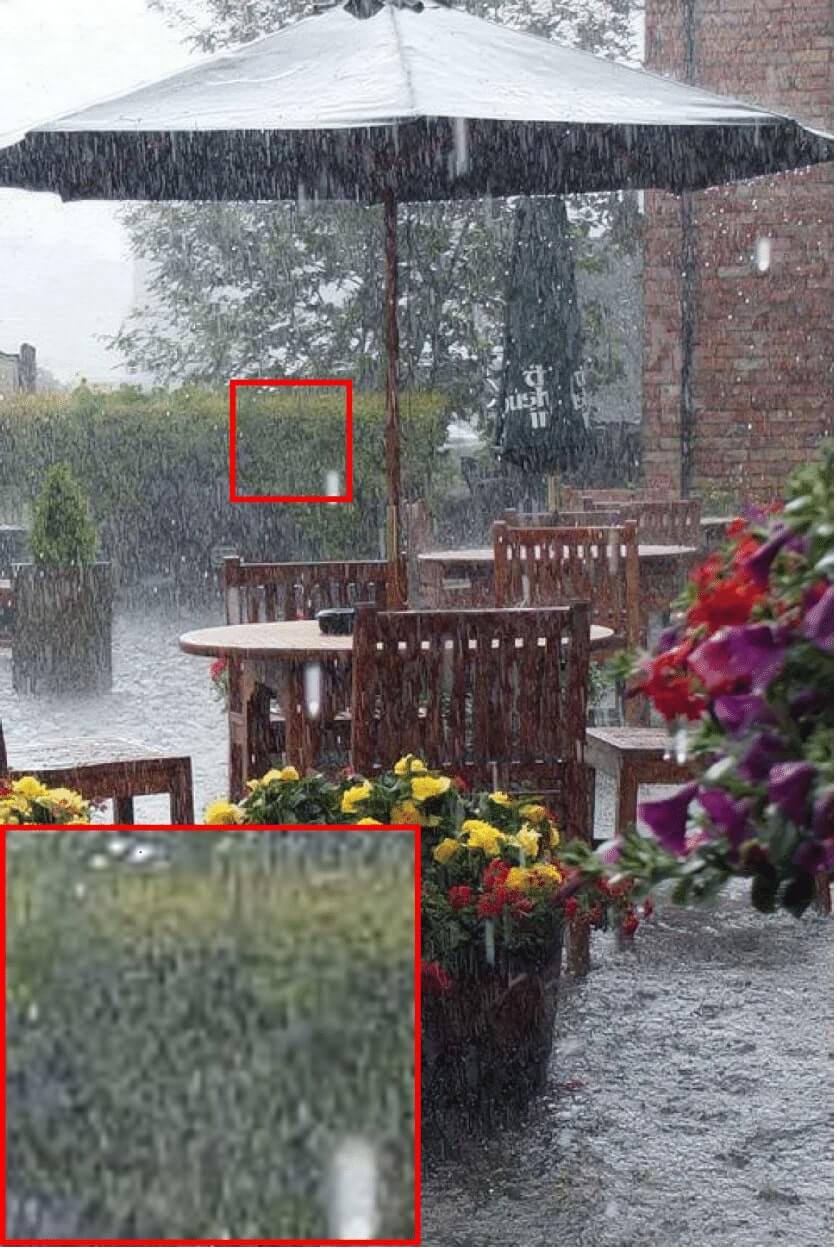}}
        
    \par\vspace{\figheight}

    \subfloat[\centering DGNL-Net~\cite{hu2021single}]{\includegraphics[width=\Qualwidth]{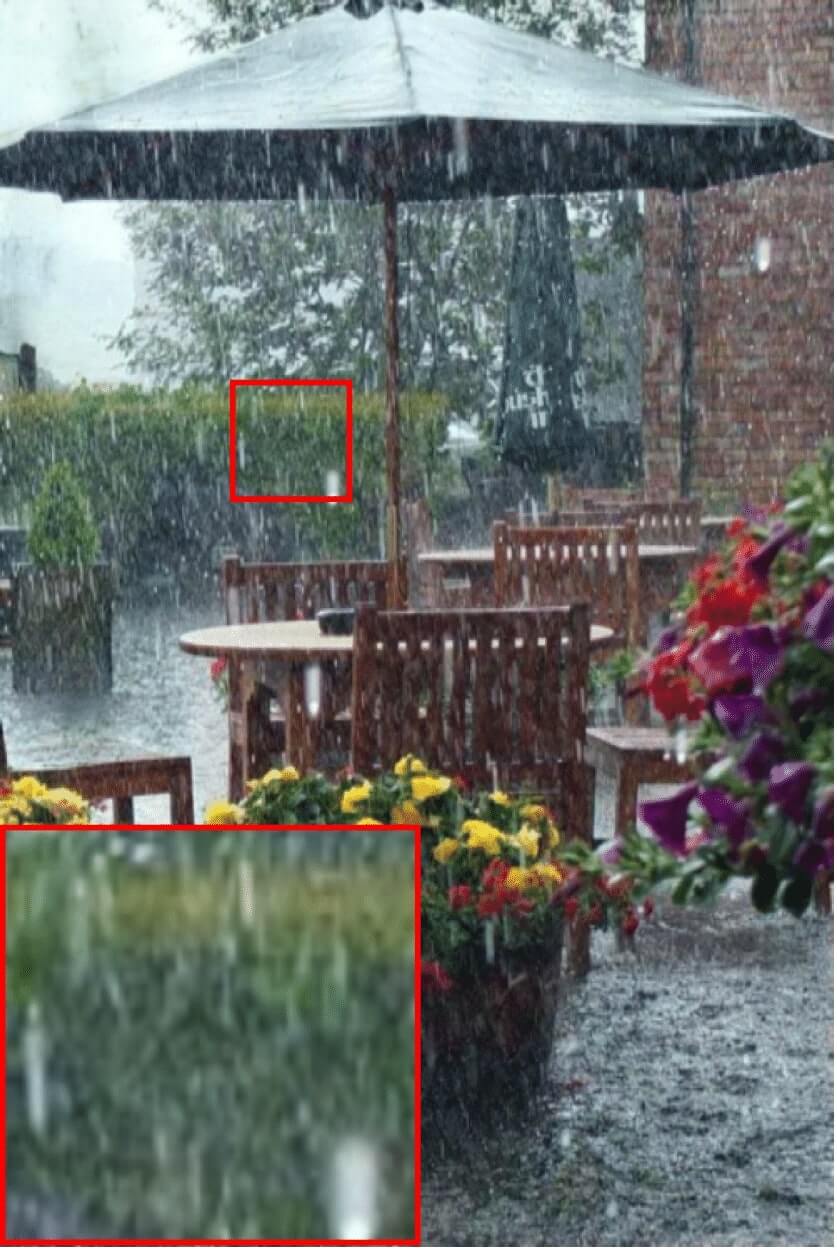}}
    \hfill
    \subfloat[\centering EDR V4 (S)~\cite{guo2021efficientderain}]{\includegraphics[width=\Qualwidth]{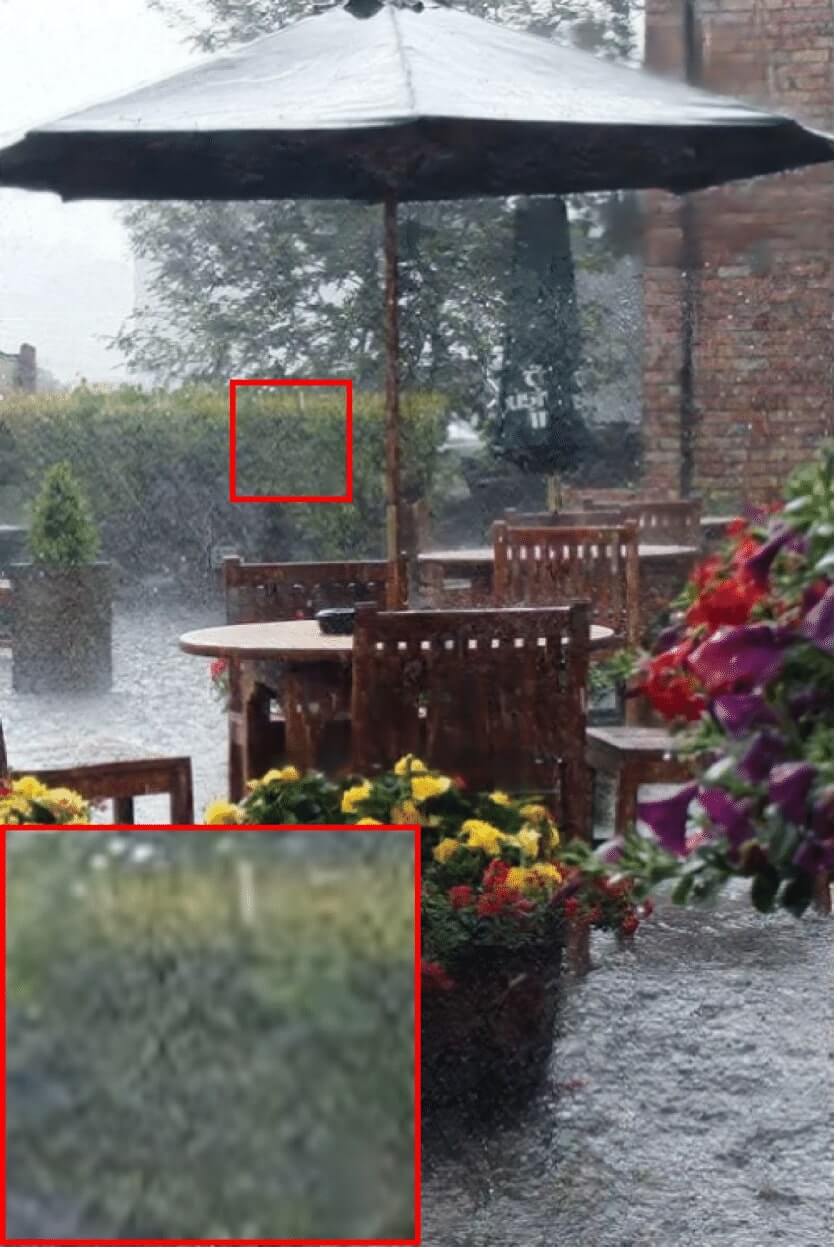}}
    \hfill
    \subfloat[\centering EDR V4 (R)~\cite{guo2021efficientderain}]{\includegraphics[width=\Qualwidth]{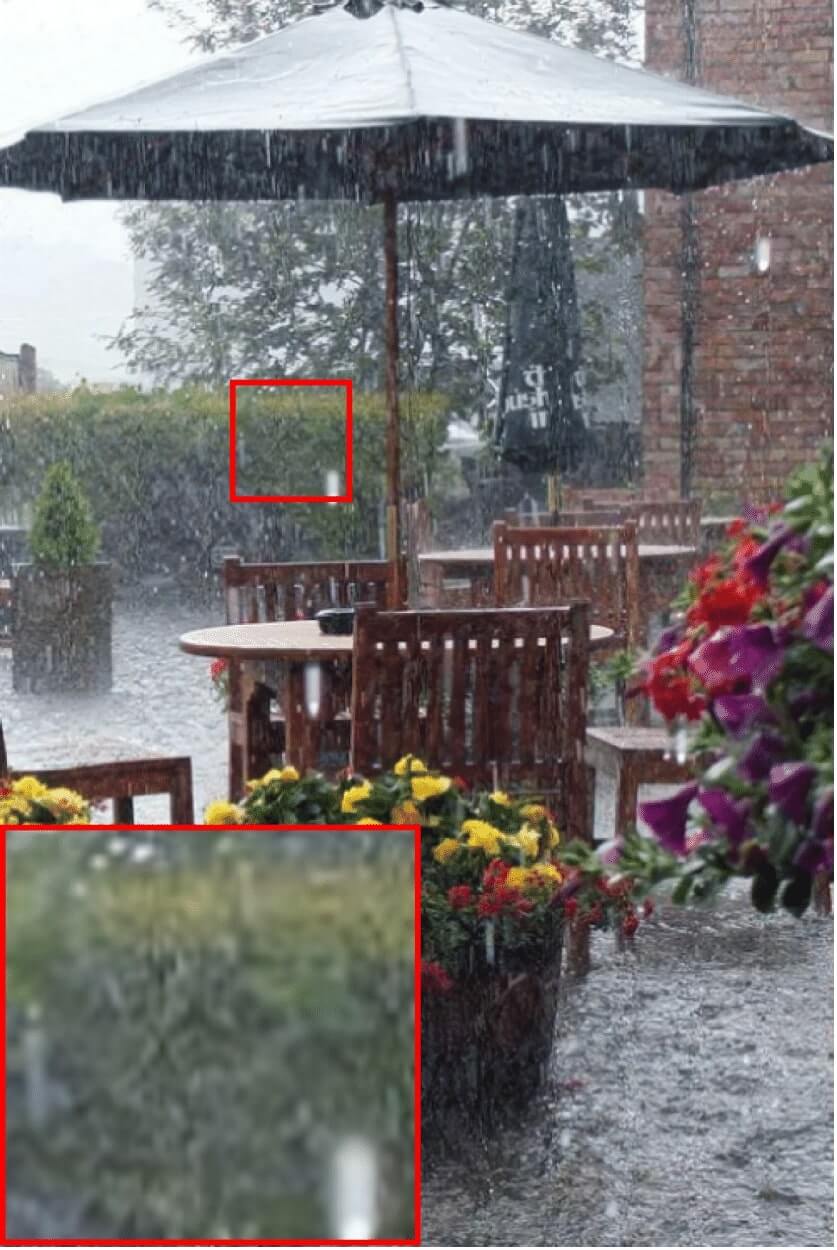}}
    \hfill
    \subfloat[\centering MPRNet~\cite{zamir2021multi} ]{\includegraphics[width=\Qualwidth]{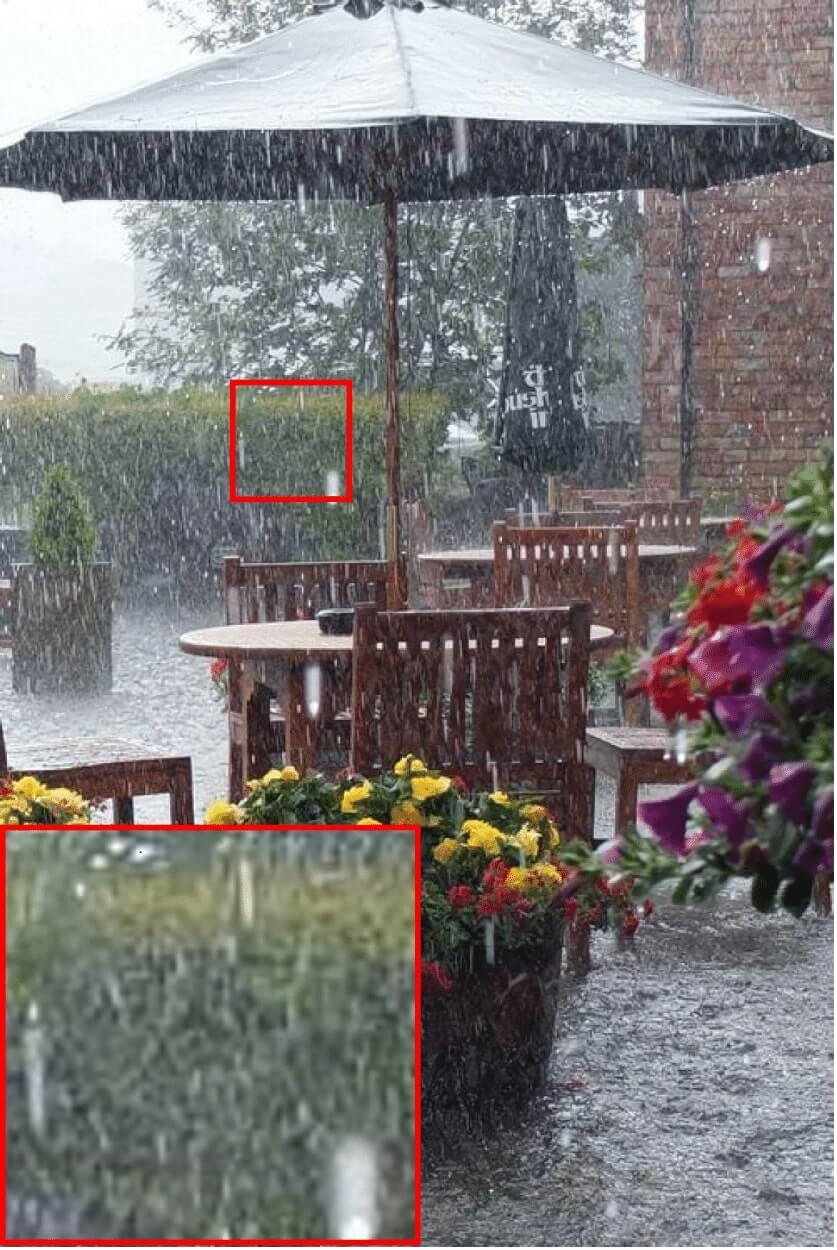}}
    \hfill
    \subfloat[\centering Ours]{\includegraphics[width=\Qualwidth]{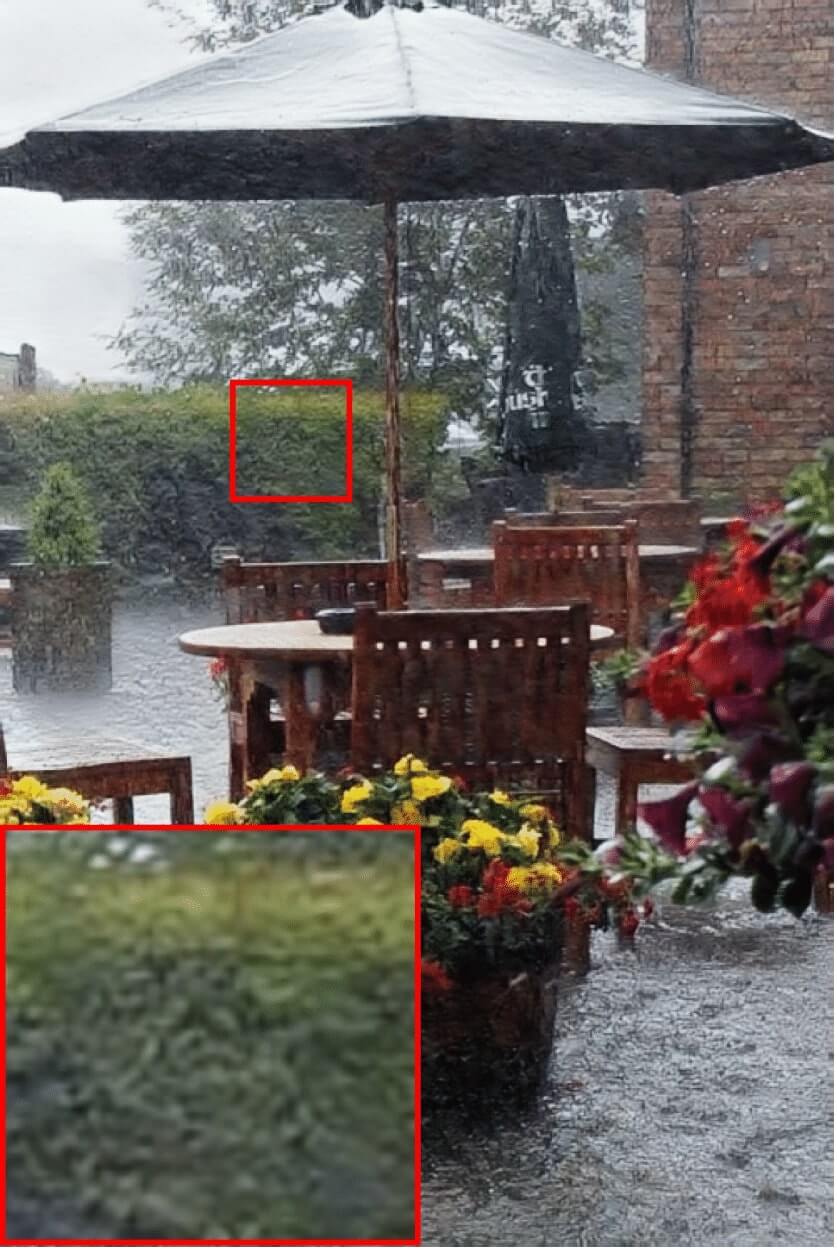}}

    \par\vspace{\figheight}

    \subfloat[\centering Rainy Image]{\includegraphics[width=\Qualwidth]{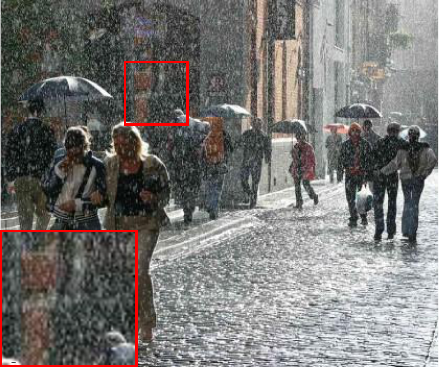}}
    \hfill
    \subfloat[\centering SPANet~\cite{wang2019spatial}]{\includegraphics[width=\Qualwidth]{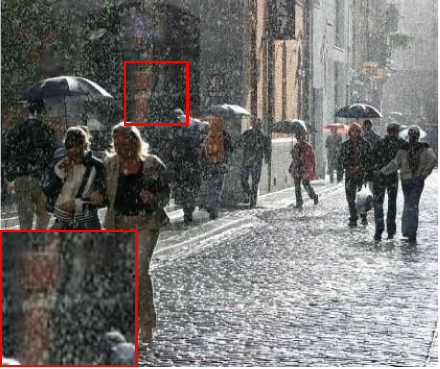}}
    \hfill
    \subfloat[\centering HRR~\cite{li2019heavy}]{\includegraphics[width=\Qualwidth]{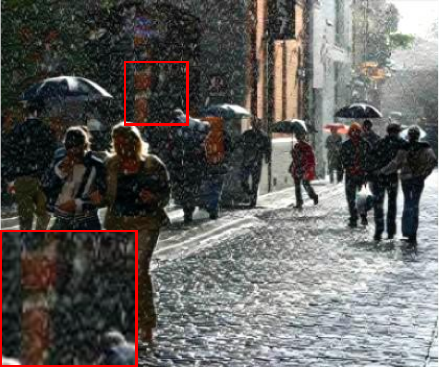}}
    \hfill
    \subfloat[\centering MSPFN~\cite{jiang2020multi}]{\includegraphics[width=\Qualwidth]{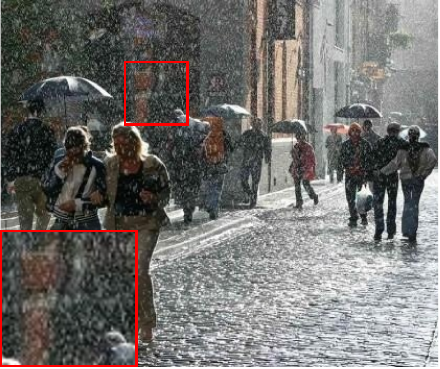}}
    \hfill
    \subfloat[\centering RCDNet~\cite{wang2020a}]{\includegraphics[width=\Qualwidth]{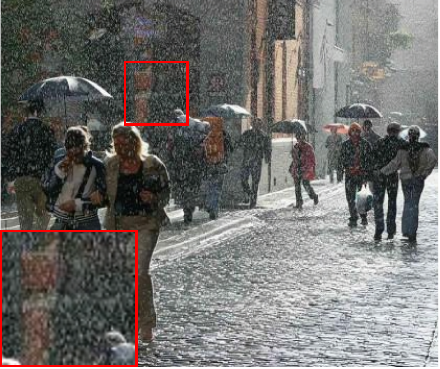}}
    
    \par\vspace{\figheight}
    
    \subfloat[\centering DGNL-Net~\cite{hu2021single}]{\includegraphics[width=\Qualwidth]{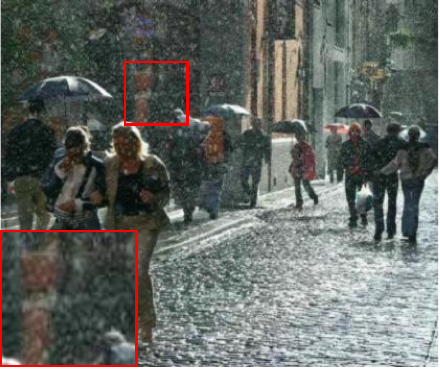}}
    \hfill
    \subfloat[\centering EDR V4 (S)~\cite{guo2021efficientderain}]{\includegraphics[width=\Qualwidth]{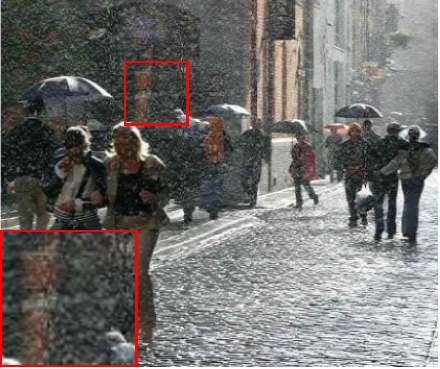}}
    \hfill
    \subfloat[\centering EDR V4 (R)~\cite{guo2021efficientderain}]{\includegraphics[width=\Qualwidth]{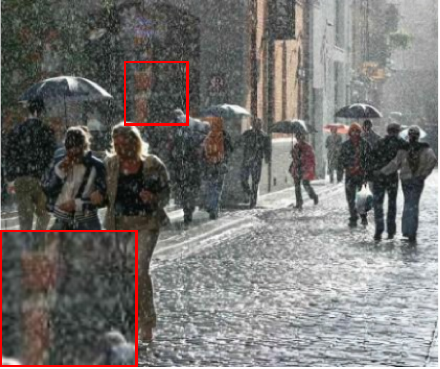}}
    \hfill
    \subfloat[\centering MPRNet~\cite{zamir2021multi}]{\includegraphics[width=\Qualwidth]{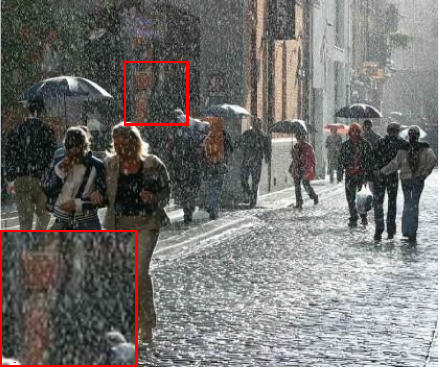}}
    \hfill
    \subfloat[\centering Ours]{\includegraphics[width=\Qualwidth]{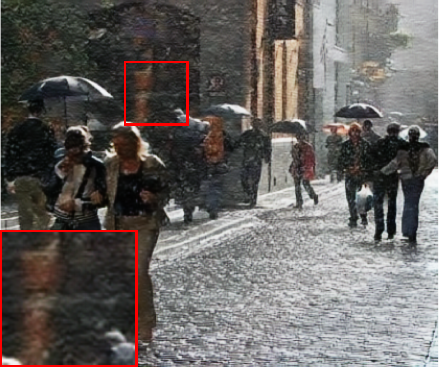}}
    
  \caption{\textbf{More qualitative results on Internet images.} Our model continues to exhibit robust generalization to real rainy images, whereas existing derainers usually fail on removing rain streaks of diverse shapes and sizes. EDR V4 (S)~\cite{guo2021efficientderain} denotes the EDR model trained on SPA-Data~\cite{wang2019spatial}, and EDR V4 (R)~\cite{guo2021efficientderain} denotes the EDR model trained on Rain14000~\cite{fu2017removing}.}
\label{fig:other_qual_results}
\end{figure*}

As a supplement to~\fignohref{6} in the main paper, we provide more qualitative results on real Internet images in~\cref{fig:other_qual_results}. Note that all comparison models are using the weights provided by the author, which are trained on synthetic or semi-real datasets. All images are taken from the dataset of common real rainy images provided by~\cite{wei2019semi}. Our proposed model trained on \dname\ continues to remove rain streaks of varying shapes and sizes as well as rain accumulation without introducing the unwanted color shifts seen in HRR~\cite{li2019heavy} and DGNL-Net~\cite{hu2021single}. 

 
\section{Qualitative Results of Retrained Methods} \label{sec:retrained_qualitative_results}
\begin{figure}
    \newcommand{\Retrainwidth}{0.165\textwidth}
    \captionsetup[subfloat]{font=scriptsize,farskip=3pt,captionskip=2pt}
    \centering
    \subfloat[\centering MPRNet (P) (19.88/0.7551)]{\includegraphics[width=\Retrainwidth]{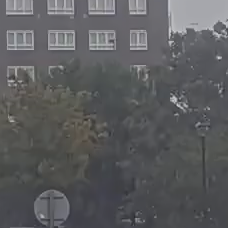}}  \hfill 
    \subfloat[\centering MPRNet (R) (\textbf{23.72}/\textbf{0.7948})]{\includegraphics[width=\Retrainwidth]{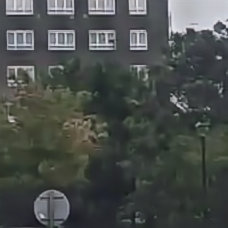}}  \hfill 
    \subfloat[\centering EDR (P) (23.93/0.8539)]{\includegraphics[width=\Retrainwidth]{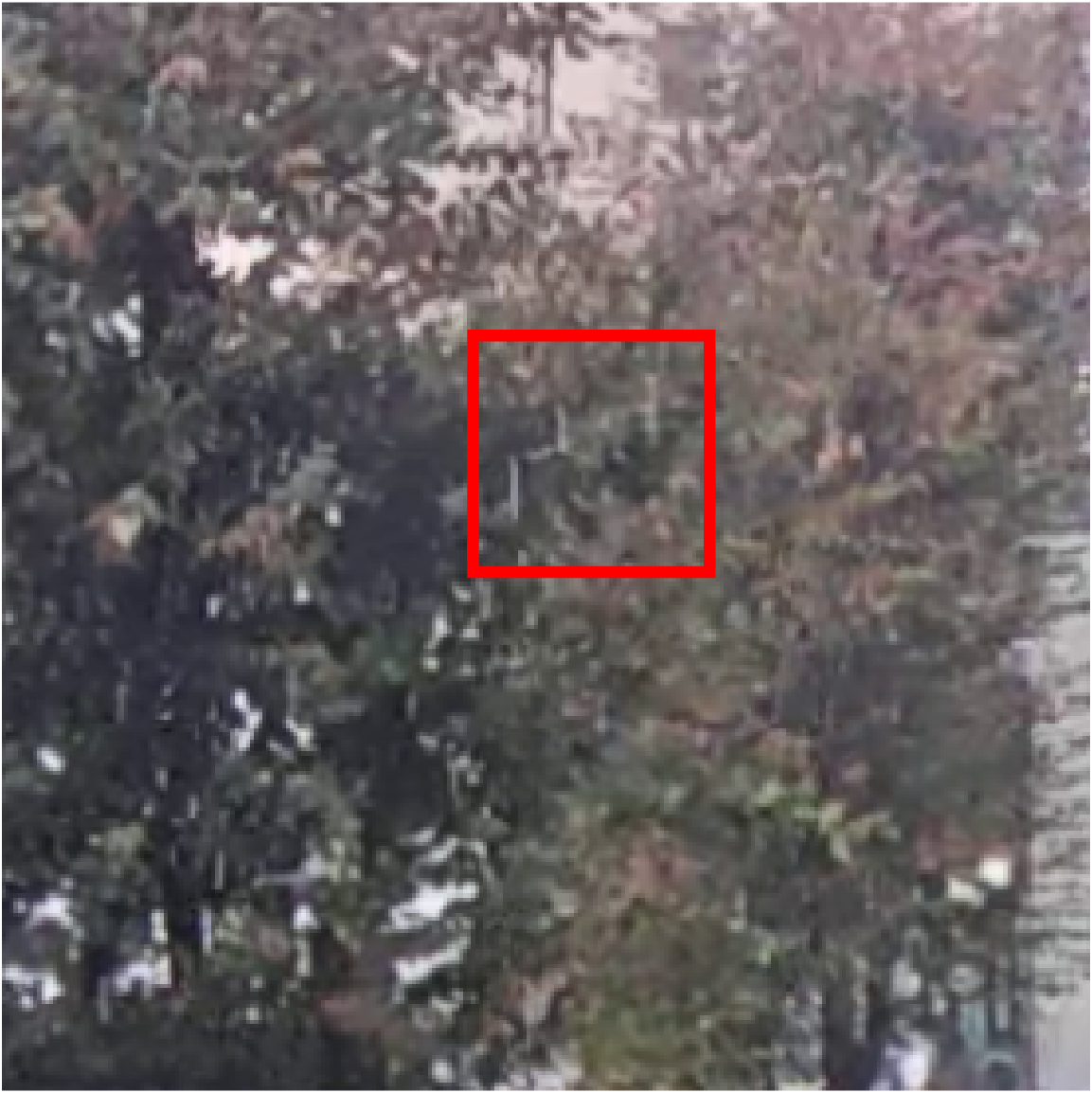}}  \hfill 
    \subfloat[\centering EDR (R) (\textbf{25.19}/\textbf{0.8632})]{\includegraphics[width=\Retrainwidth]{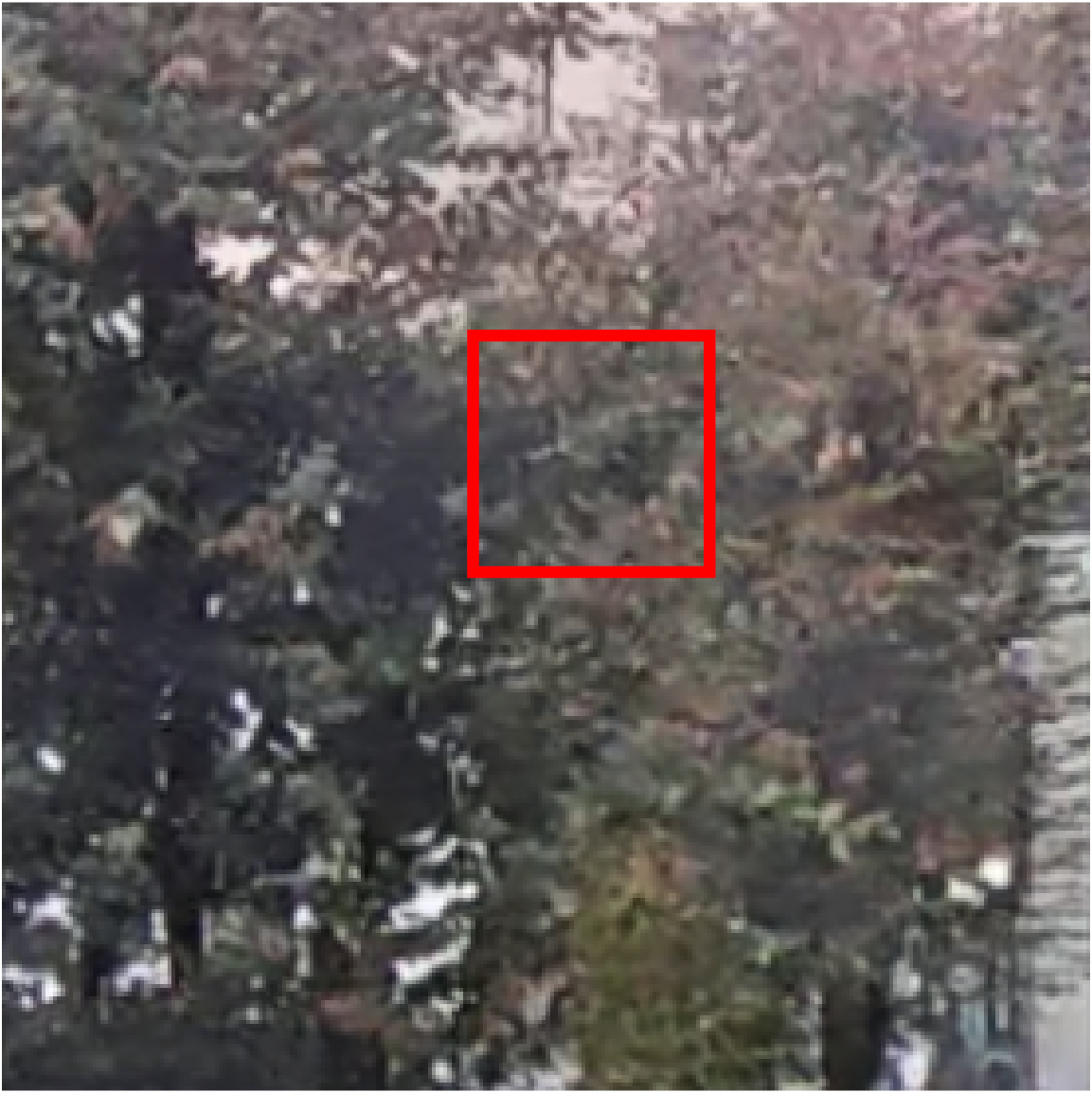}} \hfill 
    \subfloat[\centering RCDNet (P) (16.74/0.6165)]{\includegraphics[width=\Retrainwidth]{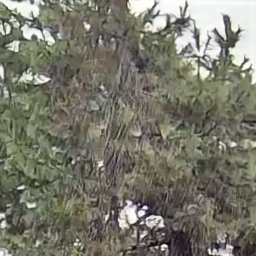}}  \hfill 
    \subfloat[\centering RCDNet (R) (\textbf{17.81}/\textbf{0.6286})]{\includegraphics[width=\Retrainwidth]{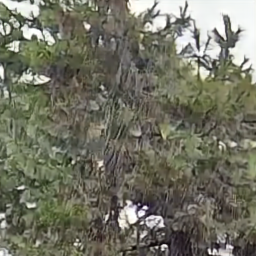}}
    \caption{\textbf{Qualitative results of retrained SOTAs.} (P) denote pretrained models provided by the original authors,  and (R) denotes the retrained models on the proposed \dname\ dataset. The improvements further highlight the effectiveness of the proposed \dname\ dataset.}
    \label{fig:retrain_qual_results}
\end{figure}

As an additional supplement to~\tabnohref{3} in the main paper, we provide some representative samples of the retrained models for some qualitative comparison in~\cref{fig:retrain_qual_results}. The visual improvements of these derainers in rain fog and streak removal further validate the effectiveness of the proposed dataset.

\section{Comparison with Semi-supervised Methods} \label{sec:comparison_semi_supervised_sota}

\begin{figure}
    \newcommand{\Retrainwidth}{0.245\textwidth}
    \centering
    \subfloat[\centering Rainy]{\includegraphics[width=\Retrainwidth]{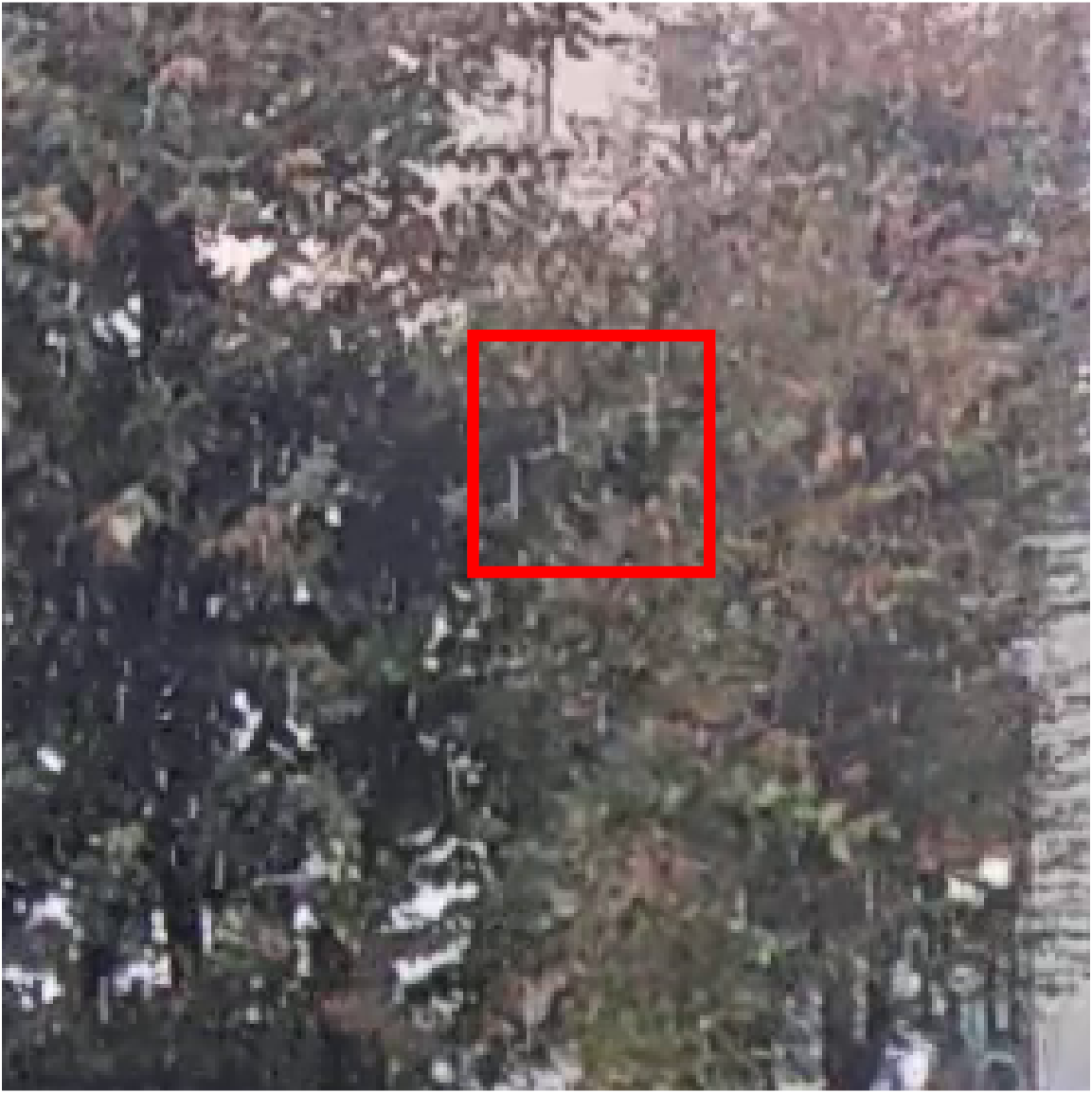}}  \hfill 
    \subfloat[\centering SSIR~\cite{wei2019semi}]{\includegraphics[width=\Retrainwidth]{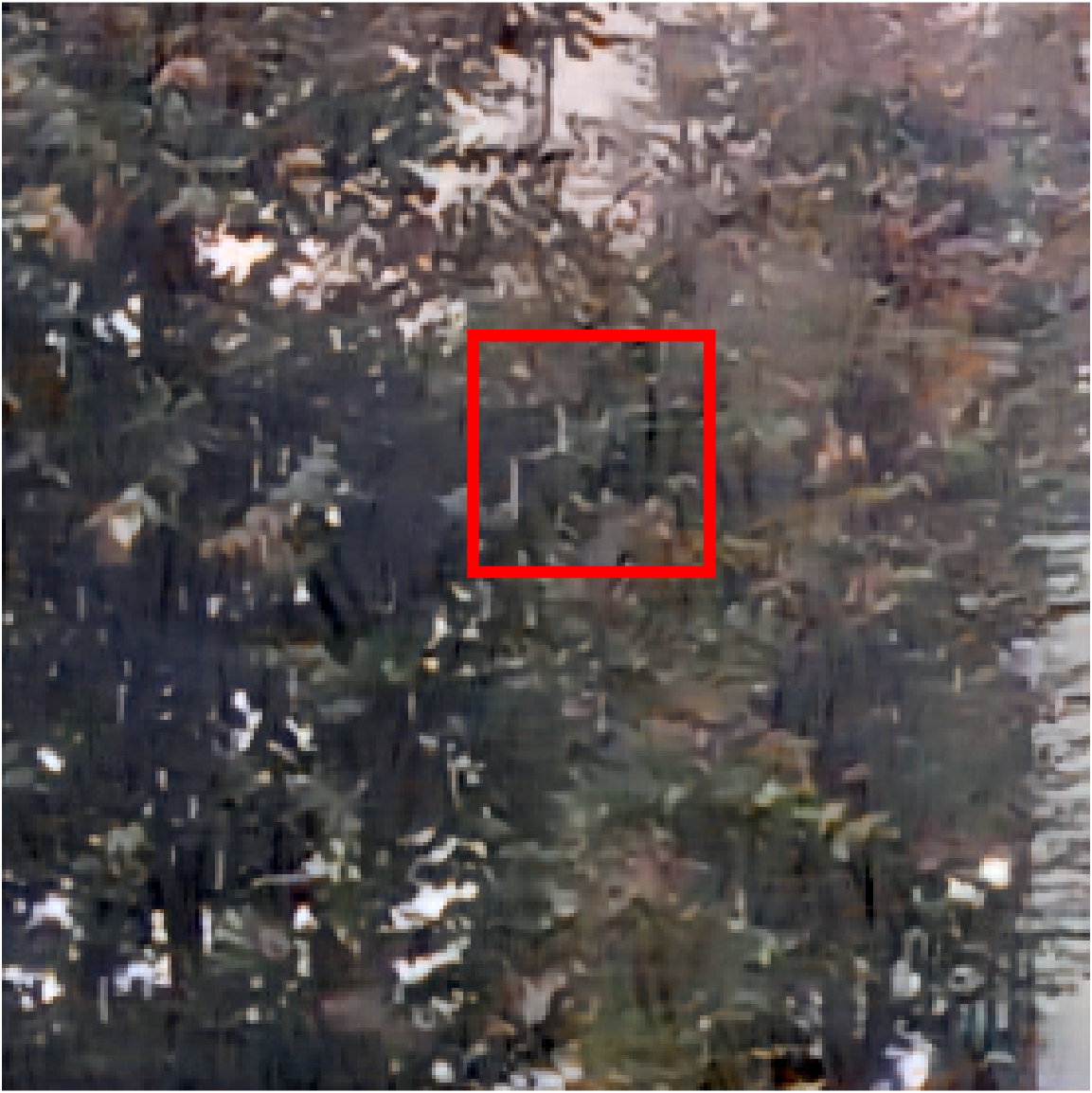}}  \hfill 
    \subfloat[\centering MOSS~\cite{huang2021memory}]{\includegraphics[width=\Retrainwidth]{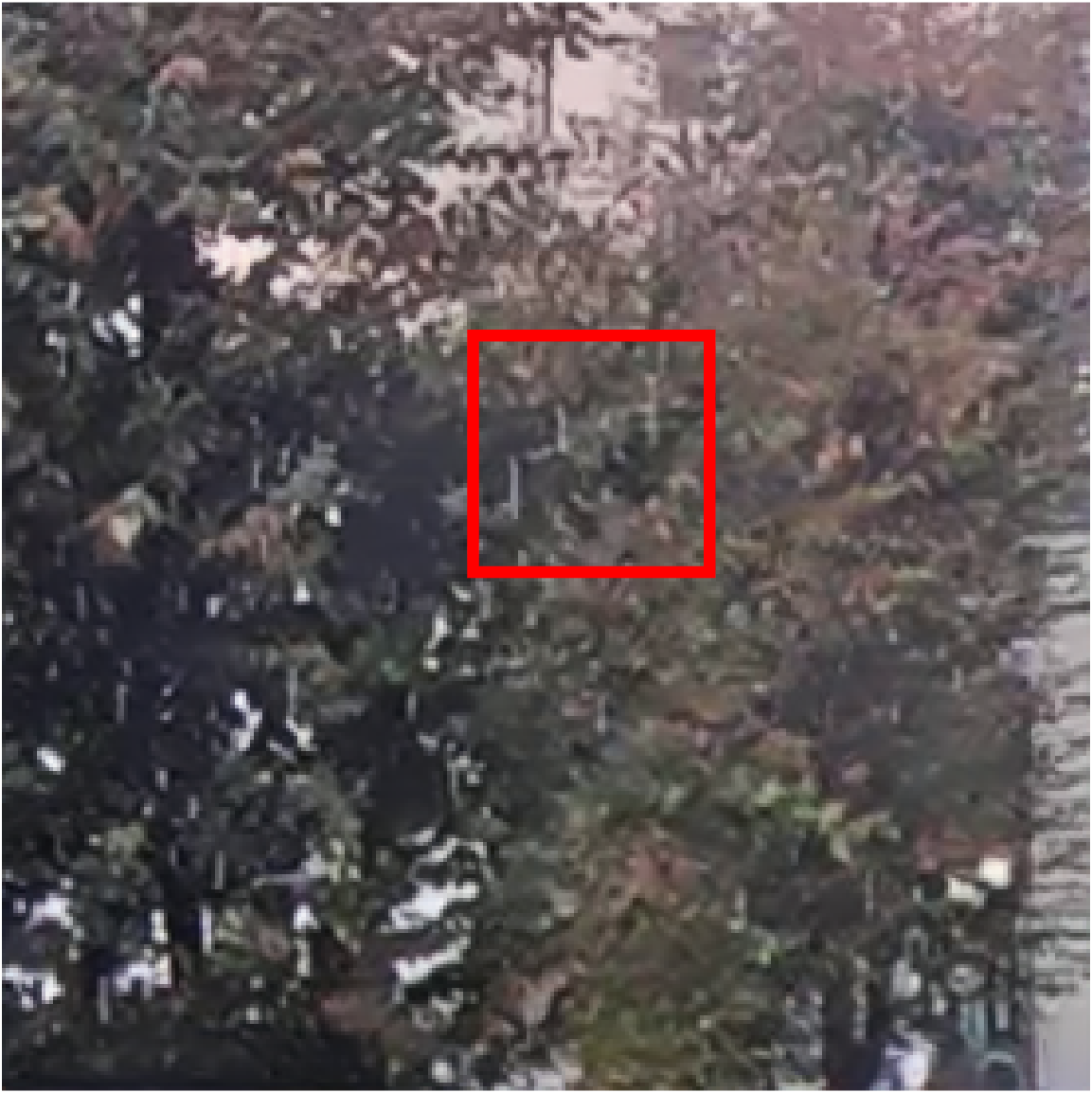}}  \hfill 
    \subfloat[\centering Ours]{\includegraphics[width=\Retrainwidth]{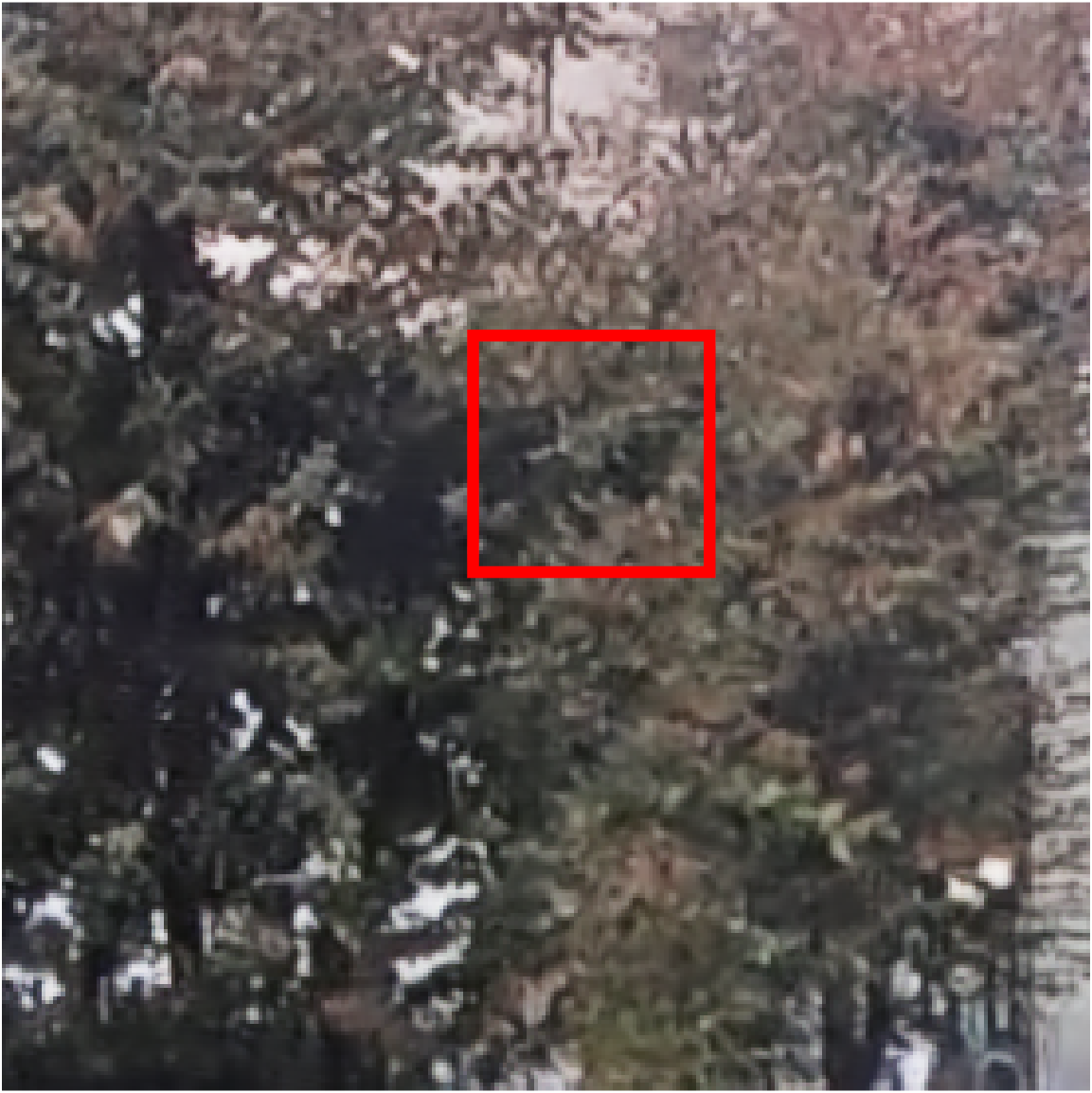}}
    \caption{\textbf{Qualitative comparison with semi-supervised SOTAs.} As compared with semi-supervised models, the proposed method can remove the rain streaks more effectively.}
    \label{fig:semi_supervised_sota_results}
\end{figure}

In addition to the models trained on synthetic and semi-real datasets, we also compare the proposed method with some recent semi-supervised methods, including SIRR~\cite{wei2019semi} and MOSS~\cite{huang2021memory}, that are trained on real images as a complement to~\tabnohref{2} in the main paper. The corresponding PSNR/SSIM scores on the entire \dname\ test set for these two semi-supervised methods are listed as follows: SIRR~\cite{wei2019semi} (20.57/0.6448), and MOSS~\cite{huang2021memory} (21.42/0.7073), where ours are (22.53/0.7304). Some qualitative results can be found in~\cref{fig:semi_supervised_sota_results}. 

\section{Alignment of Small Motions} \label{sec:alignment}
\begin{figure*}[!ht]

    \subfloat[]{\includegraphics[width=0.04\textwidth]{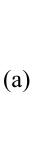}}
    \subfloat[\centering No correction needed]{\includegraphics[width=0.235\textwidth]{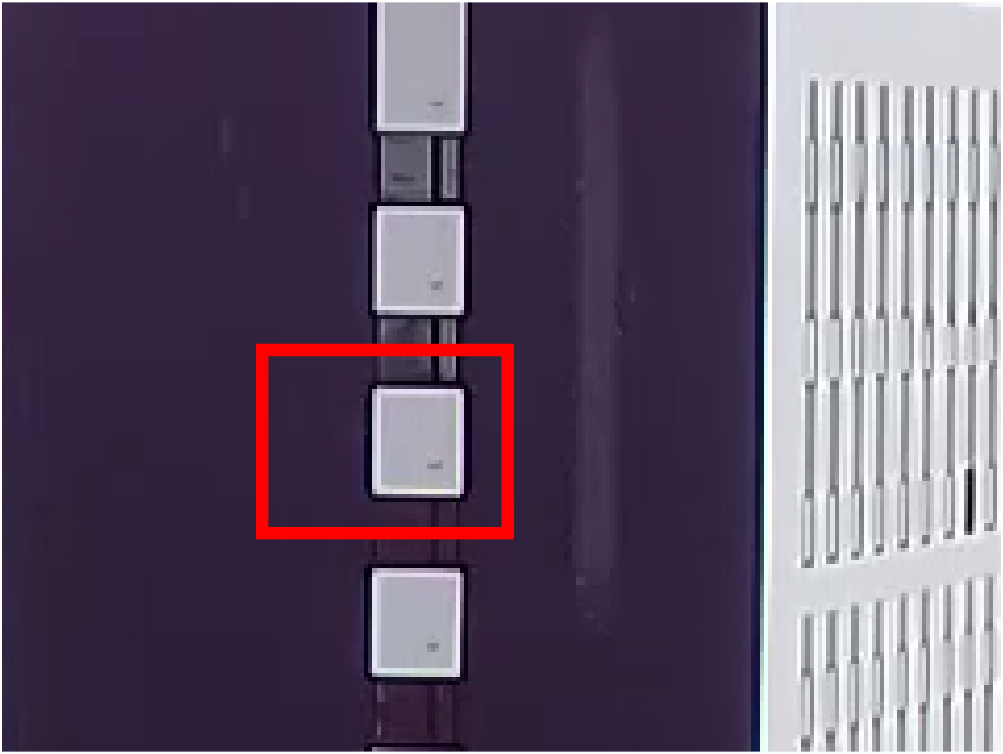}
    \hspace{-6.5pt}
    \includegraphics[width=0.235\textwidth]{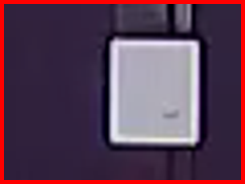}}
    \hfill
    \subfloat[\centering No correction needed]{\includegraphics[width=0.235\textwidth]{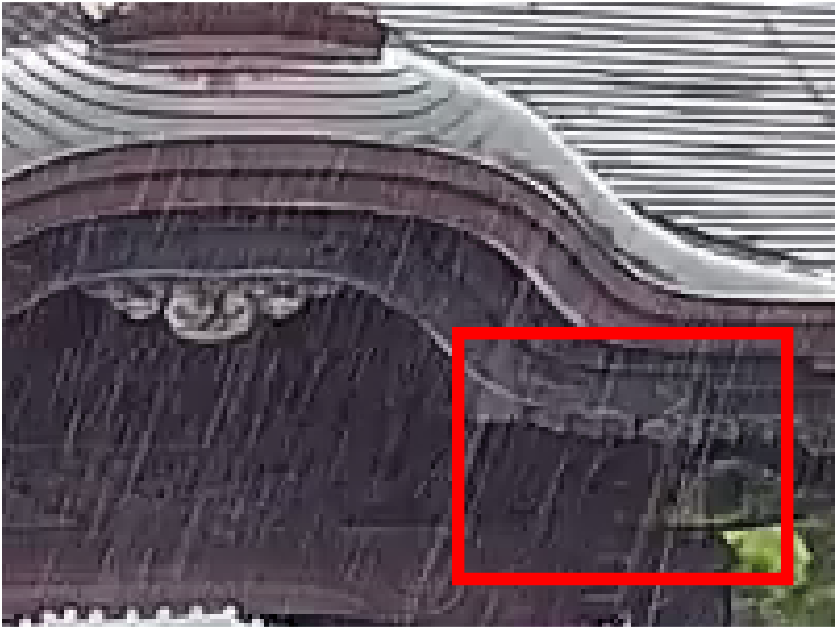}
    \hspace{-6.5pt}
    \includegraphics[width=0.235\textwidth]{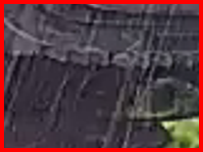}}
    
    \par\vspace{-10pt}
    
    \subfloat[]{\includegraphics[width=0.04\textwidth]{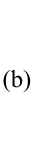}}
    \subfloat[\centering Before correction]{\includegraphics[width=0.235\textwidth]{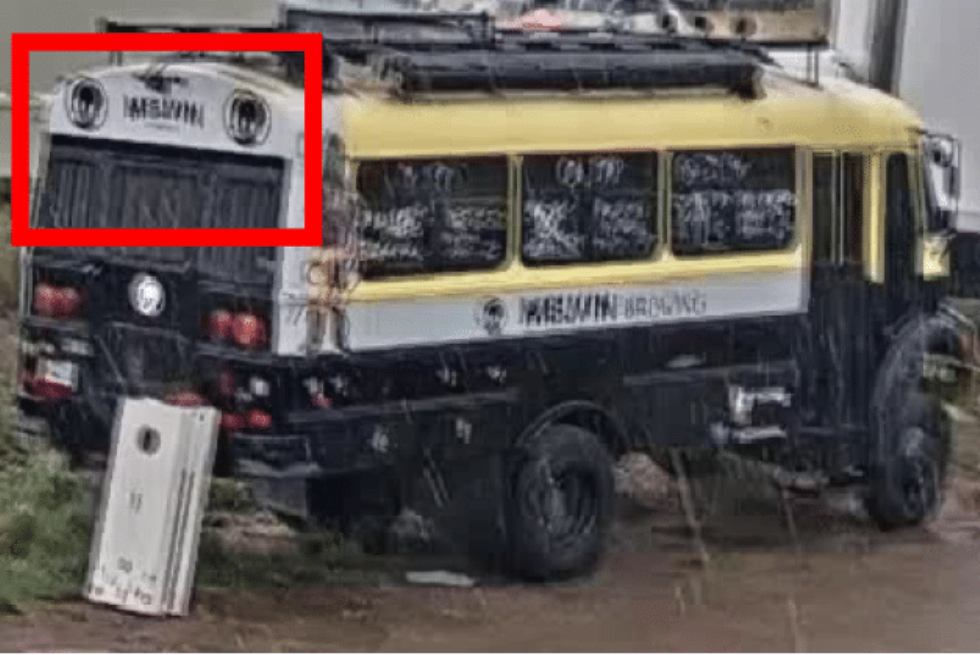}
    \hspace{-6.5pt}
    \includegraphics[width=0.235\textwidth]{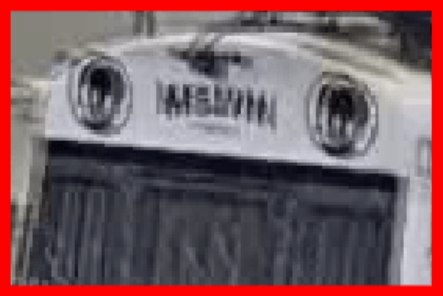}}
    \hfill
    \subfloat[\centering After correction]{\includegraphics[width=0.235\textwidth]{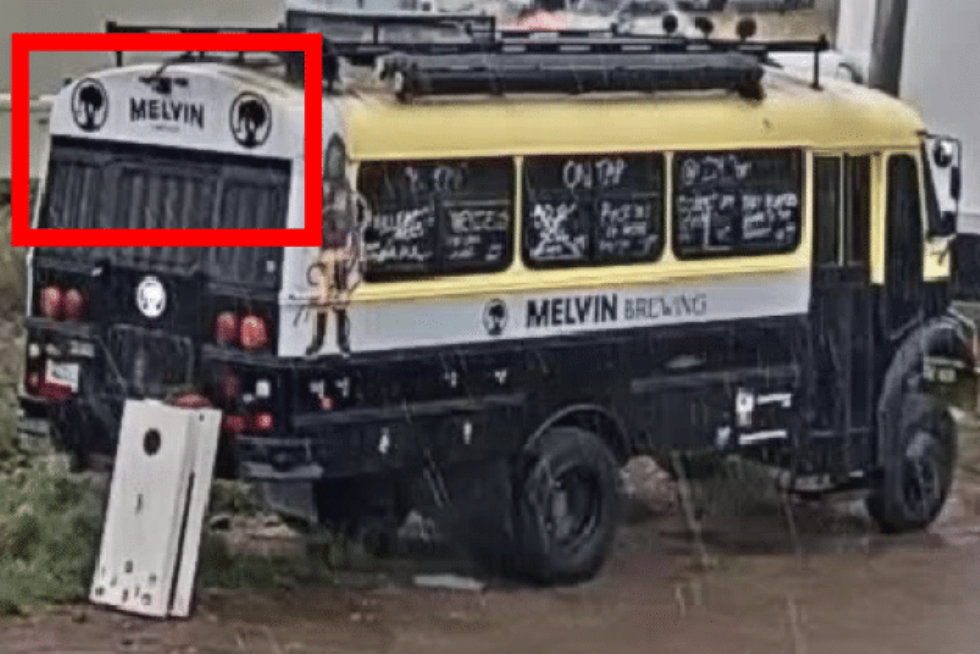}
    \hspace{-6.5pt}
    \includegraphics[width=0.235\textwidth]{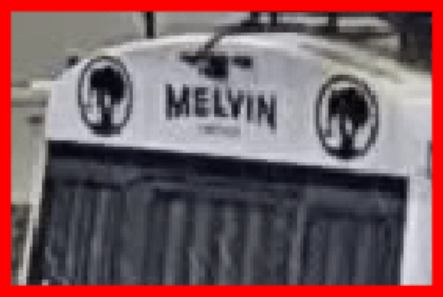}}
    \caption{\textbf{The proposed method can correct for small motions under rain.} We illustrate two types of scenes by overlaying the rainy images on top of their clean ground truths: (a) two scenes that do not need additional image processing for motion alignment; and (b) a scene with motion before and after running the correction algorithms. It should be noted that both types of scenes are aligned properly in \dname.}
\label{fig:alignment_results}
\end{figure*}

As a complement to~\secnohref{3} of the main paper, we first show, in~\cref{fig:alignment_results}-(a), a ground-truth image overlayed on top of a rainy image to demonstrate representative samples that passed our data collection appearance criteria and also motion criterion, where we do not need to perform motion correction. We note that this is the case for the majority of our dataset. Additionally, we show an overlayed image pair that passed our appearance criteria, but failed the motion criterion. \cref{fig:alignment_results}-(b) shows the image pair before and after the motion correction. It should be noted that only a small portion of the data requires such correction, and our correction pipeline is designed to be robust to rain artifacts. It is because even though rain can influence local descriptors, the combinatorial matching stage is designed to be robust to a preponderance of outliers. For most cases, the percentage of outliers affects the time it takes to converge, but not the quality. All samples that require our correction procedure were manually inspected after the alignment -- any failure cases of the procedure, typically due to extreme weather conditions, were manually removed. 

\section{Runtime Comparison} \label{sec:runtime}

We list the total number of parameters with the associated runtime for other state-of-the-art methods and our proposed model in~\cref{tab:runtime}. The comparison is conducted on a single NVIDIA P100 GPU, and each derainer is asked to restore a colored rainy image of size $256 \times 256$. We note that the top three methods (DGNL-Net~\cite{hu2021single}, EDR~\cite{guo2021efficientderain}, and our proposed method) all operate at real-time deraining speeds. However, our method outperforms them by 3.73 dB and 2.72 dB PSNR respectively.

\begin{table}
  \caption{\textbf{Runtime comparison.} The average inference time is calculated on $256 \times 256$ color images.}
  \label{tab:runtime} 
  \centering
  \small
  \resizebox{\columnwidth}{!}{
  \begin{tabular}{ccccccccccc}
    \toprule
    \makecell{Model} & \makecell{SPANet~\cite{wang2019spatial} \\ (CVPR'19)} & \makecell{HRR~\cite{li2019heavy} \\ (CVPR'19)} & \makecell{MSPFN~\cite{jiang2020multi} \\ (CVPR'20)} & \makecell{RCDNet~\cite{wang2020a} \\ (CVPR'20)} & \makecell{DGNL-Net~\cite{hu2021single} \\ (IEEE TIP'21)} & \makecell{EDR~\cite{guo2021efficientderain} \\ (AAAI'21)} & \makecell{MPRNet~\cite{zamir2021multi} \\ (CVPR'21)} & \makecell{Ours} \\
    \midrule
    \makecell{Number of \\ Parameters} & \makecell{284k} & \makecell{40.6M} & \makecell{15.8M} & \makecell{3.16M} & \makecell{4.03M} & \makecell{27.3M} & \makecell{3.63M} & \makecell{12.9M} \\
    \midrule
    \makecell{Inference \\ Time (ms)} & \makecell{86.65} & \makecell{35.35} & \makecell{145.5} & \makecell{189.6} & \makecell{4.230} & \makecell{4.617} & \makecell{36.91} & \makecell{12.79} \\
    \bottomrule
  \end{tabular}}
\end{table}


\section{Limitations} \label{sec:limitations}

Although we achieve the state of the art for deraining real images, our method is not perfect. Our PSNR and SSIM scores on \dname\ are 22.53 dB and 0.7304. This suggests that indeed, we still have ample room for improvement. For example, we leave a slight rain accumulation in the tree in~\cref{fig:other_quant_results}. While the recovered image is sharper and contains less rain artifacts than competing methods, boundaries in highly textured areas (e.g. leaves, bricks, and foliage) are blurred. In~\cref{fig:other_qual_results}, we observe a similar trend. However, this is a challenge that plagues all methods. We hope that further extensions of our approach and \dname\ will help mitigate these artifacts. We also do not consider occlusions from raindrops on the camera lens because the raindrops will likewise be present on the lens after the rain stops. Moreover, we do not consider specular reflections from water surfaces. This is because these reflections are nearly impossible to reconstruct as the water ripples in the puddles will destroy the visual patterns during raining. We hope that future works can address these limitations. While we have describe image restoration as the main task of deraining, we conjecture that our results may also be applicable towards the re-use of pretrained models on clean data for downstream tasks like: depth completion \cite{hu2021penet,liu2022monitored,merrill2021robust,park2020non,wong2021learning,wong2021adaptive,wong2020unsupervised,wong2021unsupervised,yang2019dense}, stereo \cite{berger2022stereoscopic,chang2018pyramid,duggal2019deeppruner,wong2021stereopagnosia,xu2020aanet}, optical flow \cite{aleotti2020learning,lao2017minimum,lao2018extending,lao2019minimum,sun2018pwc,teed2020raft}, object detection~\cite{glenn_jocher_2022_6222936,kalra2021towards,liu2016ssd,redmon2016you}, and monocular depth prediction \cite{fei2019geo,godard2019digging,poggi2020uncertainty,poggi2022real,ranftl2021vision,watson2019self,wong2020targeted,wong2019bilateral}.



\section{Comparison Code Links} \label{sec:comparison_code}
The code links for all the comparison methods in the main paper are listed in~\cref{tab:comparison_code}. 
\begin{table*}[h]
  \centering
  \small
  \caption{\textbf{Code links for the comparison methods.}}
  \begin{tabular}{ll}
    \toprule
    Methods & Links \\
    \midrule
    SPANet~\cite{wang2019spatial} (CVPR'19) & \url{https://github.com/stevewongv/SPANet} \\
    HRR~\cite{li2019heavy} (CVPR'19) & \url{https://github.com/liruoteng/HeavyRainRemoval} \\
    MSPFN~\cite{jiang2020multi} (CVPR'20) & \url{https://github.com/kuijiang0802/MSPFN}\\ 
    RCDNet~\cite{wang2020a} (CVPR'20) & \url{https://github.com/hongwang01/RCDNet}\\
    DGNL-Net~\cite{hu2021single} (IEEE TIP'21) & \url{https://github.com/xw-hu/DGNL-Net}\\
    Efficient Derain~\cite{guo2021efficientderain} (AAAI'21) & \url{https://github.com/tsingqguo/efficientderain} \\
    MPRNet~\cite{zamir2021multi} (CVPR'21) & \url{https://github.com/swz30/MPRNet}\\
    \bottomrule
  \end{tabular}

  \label{tab:comparison_code} 
\end{table*}

\section{Network Architecture \& Implementation} \label{sec:architecture}

As an additional supplement of the network architecture \& implementation section in the main paper, we provide more implementation details here. In our model, the input convolutional block contains two convolutional layers with kernel sizes of $7 \times 7$ and $3 \times 3$ respectively. The downsampling blocks are instantiated by $3 \times 3$ convolutional layers with a stride of 2, and each upsampling block consists of a bilinear interpolation layer and a $3 \times 3$ convolutional layer. Please refer to ~\cref{tab:network_architecutre} for a more detailed illustration of the network architecture. We use batch normalization~\cite{ioffe2015batch} and choose leaky ReLUs~\cite{maas2013rectifier} with a negative slope of 0.1 as the activation function. Our model is implemented in PyTorch~\cite{NEURIPS2019_9015}. The MS-SSIM loss is implemented based on the PyTorch Image Quality (PIQ) library~\cite{piq}. Experiments are conducted on an NVIDIA Tesla P100 GPU.

\begin{table*}[h]
    \centering
    \caption{\textbf{Illustration of our network architecture.}}
\fontsize{8}{11}\selectfont
    \resizebox{0.735\columnwidth}{!}{
    \begin{tabular}{lcccccccc}
        \toprule 
        \multirow{2}{*}[-0.1cm]{\textbf{Network}} & \multicolumn{2}{c}{\centering Kernel} & \multicolumn{2}{c}{\centering Channels} & \multicolumn{2}{c}{\centering Resolution} & \multirow{2}{*}[-0.1cm]{Parameters} & \multirow{2}{*}[-0.1cm]{Input} \\ 
        \cmidrule(lr){2-3} \cmidrule(lr){4-5} \cmidrule(lr){6-7}
         & Size & Stride & In & Out & In & Out & &  \\ 
        \midrule
        \textbf{\textit{Encoder}} & \multicolumn{1}{l}{} & \multicolumn{1}{l}{} & \multicolumn{1}{l}{} & \multicolumn{1}{l}{} & \multicolumn{1}{l}{} & \multicolumn{1}{l}{} & \multicolumn{1}{l}{} \\
        \midrule
        InputConv1          & 7     & 1     & 3     & 64    & 1     & 1     & $\approx$ 9.5k        & Rainy Image       \\ \midrule
        InputConv2          & 3     & 1     & 64    & 64    & 1     & 1     & $\approx$ 37.0k       & InputConv1        \\ \midrule
        DownConv1           & 3     & 2     & 64    & 128   & 1     & 1/2   & $\approx$ 74.0k       & InputConv2        \\ \midrule
        DownConv2           & 3     & 2     & 128   & 256   & 1/2   & 1/4   & $\approx$ 295.4k      & DownConv1         \\ \midrule
        DeformResBlock1 \\
        \cmidrule(lr){1-1}
        DeformConv11        & 3     & 1     & 256   & 256   & 1/4   & 1/4   & $\approx$ 652.6k        & DownConv2               \\
        DeformConv12        & 3     & 1     & 256   & 256   & 1/4   & 1/4   & $\approx$ 652.6k        & DeformConv11            \\ 
        Sum1                & -     & -     & 256   & 256   & 1/4   & 1/4   & \multicolumn{2}{c}{DownConv2 + DeformConv12}      \\ 
        \midrule
        DeformResBlock2 \\
        \cmidrule(lr){1-1}
        DeformConv21        & 3     & 1     & 256   & 256   & 1/4   & 1/4   & $\approx$ 652.6k        & Sum1            \\
        DeformConv22        & 3     & 1     & 256   & 256   & 1/4   & 1/4   & $\approx$ 652.6k        & DeformConv21    \\ 
        Sum2                & -     & -     & 256   & 256   & 1/4   & 1/4   & \multicolumn{2}{c}{Sum1 + DeformConv21}   \\ 
        \midrule
        \multicolumn{1}{c}{\vdots}  \\ \midrule
        DeformResBlock9 \\
        \cmidrule(lr){1-1}
        DeformConv91        & 3     & 1     & 256   & 256   & 1/4   & 1/4   & $\approx$ 652.6k        & Sum8                \\
        DeformConv92        & 3     & 1     & 256   & 256   & 1/4   & 1/4   & $\approx$ 652.6k        & DeformConv91        \\
        Sum9                & -     & -     & 256   & 256   & 1/4   & 1/4   & \multicolumn{2}{c}{Sum8 + DeformConv92}       \\ \midrule
        \textbf{\textit{Decoder}} & \multicolumn{1}{l}{} & \multicolumn{1}{l}{} & \multicolumn{1}{l}{} & \multicolumn{1}{l}{} & \multicolumn{1}{l}{} & \multicolumn{1}{l}{} & \multicolumn{1}{l}{} \\
        \midrule
        UpConvBlock1 \\
        \cmidrule(lr){1-1}
        Bilinear1       & -     & -     & 256       & 256   & 1/4   & 1/2   & -                     & Sum9                      \\
        Conv11          & 3     & 1     & 256       & 128   & 1/2   & 1/2   & $\approx$ 295.2k      & Bilinear2                 \\
        Concat1         & -     & -     & 128 + 128 & 256   & 1/2   & 1/2   & \multicolumn{2}{c}{DownConv1, Conv11}             \\
        Conv12          & 3     & 1     & 256       & 128   & 1/2   & 1/2   & $\approx$ 295.2k        & Concat1                 \\ 
        \midrule
        UpConvBlock2 \\
        \cmidrule(lr){1-1}
        Bilinear2       & -     & -     & 128       & 128   & 1/2   & 1     & -                     & Conv12                \\
        Conv21          & 3     & 1     & 128       & 64    & 1     & 1   & $\approx$ 73.9k         & Bilinear2             \\
        Concat2         & -     & -     & 64 + 64   & 128   & 1     & 1   & \multicolumn{2}{c}{InputConv2, Conv21}          \\
        Conv22          & 3     & 1     & 128       & 64    & 1     & 1   & $\approx$ 73.9k         & Concat2               \\ 
        \midrule 
        OutputConv      & 3     & 1     & 64        & 3     & 1     & 1     & $\approx$ 1.7k        & Conv22        \\ \midrule
        \midrule
        Total Parameters & $\approx$ 12.9M \\
        \bottomrule
    \end{tabular}}

    \label{tab:network_architecutre}
\end{table*} 




\clearpage
%
%

 \bibliographystyle{splncs04}
 \bibliography{egbib}